\documentclass{article} %
\usepackage[dvipsnames]{xcolor}
\usepackage{iclr2023_conference,times}

\usepackage{amsmath,amsfonts,bm}

\def\eqref#1{equation~\ref{#1}}

\def\1{\bm{1}}

\def\vh{{\bm{h}}}

\def\vm{{\bm{m}}}

\def\vs{{\bm{s}}}

\def\vy{{\bm{y}}}
\def\vz{{\bm{z}}}

\DeclareMathAlphabet{\mathsfit}{\encodingdefault}{\sfdefault}{m}{sl}
\SetMathAlphabet{\mathsfit}{bold}{\encodingdefault}{\sfdefault}{bx}{n}

\DeclareMathOperator*{\argmax}{arg\,max}

\let\originalleft\left
\let\originalright\right
\renewcommand{\left}{\mathopen{}\mathclose\bgroup\originalleft}
\renewcommand{\right}{\aftergroup\egroup\originalright}

\newcommand{\Fig}[1]{\hyperref[{#1}]{Figure~\ref*{#1}}}  %
\newcommand{\fig}[1]{\hyperref[{#1}]{Fig.~\ref*{#1}}}    %
\newcommand{\figs}[1]{\hyperref[{#1}]{Figs.~\ref*{#1}}}  %
\newcommand{\Tab}[1]{\hyperref[{#1}]{Table~\ref*{#1}}}
\newcommand{\tab}[1]{\hyperref[{#1}]{Table~\ref*{#1}}}
\newcommand{\Eqn}[1]{\hyperref[{#1}]{Equation~\ref*{#1}}}
\newcommand{\eqn}[1]{\hyperref[{#1}]{Eq.~\ref*{#1}}} %
\renewcommand{\sec}[1]{\hyperref[{#1}]{Sec.~\ref*{#1}}} %
\newcommand{\supp}[1]{\hyperref[{#1}]{Suppl.~\ref*{#1}}}
\newcommand{\app}[1]{\hyperref[{#1}]{App.~\ref*{#1}}}
\newcommand{\App}[1]{\hyperref[{#1}]{Appendix~\ref*{#1}}}

\usepackage{xspace}
\makeatletter
\DeclareRobustCommand\onedot{\futurelet\@let@token\@onedot}
\def\@onedot{\ifx\@let@token.\else.\null\fi\xspace}
\def\eg{e.g\onedot}
\def\ie{i.e\onedot}
\def\cf{cf\onedot}

\def\vs{vs\onedot}
\makeatother
\usepackage{minitoc}
\usepackage{microtype}
\usepackage{booktabs}       %
\usepackage{nicefrac}       %
\usepackage{csquotes}
\usepackage{numprint}
\npdecimalsign{.}
\usepackage{wrapfig}
\usepackage{cancel}
\usepackage{caption}
\usepackage{subcaption}
\usepackage[export]{adjustbox}
\usepackage{makecell}
\usepackage{relsize}
\usepackage{scalerel}
\usepackage{enumitem}
\usepackage{placeins}

\usepackage{hyperref}
\usepackage{url}

\usepackage{fontawesome}

\usepackage{comment}
\usepackage{tikz}
\usetikzlibrary{fit}
\usepackage{xfp}

\usepackage{multirow}

\usepackage{hyperref}
\hypersetup{colorlinks,
            linkcolor=NavyBlue,
            urlcolor=NavyBlue,
            citecolor=NavyBlue}

\title{Bridging the Gap to Real-World Object-Centric Learning}

\author{Maximilian Seitzer\textsuperscript{1,$\dagger$}\hfill
    Max Horn\textsuperscript{2}\hfill
    Andrii Zadaianchuk\textsuperscript{1,3,$\dagger$}\hfill
    Dominik Zietlow\textsuperscript{2}\\
    \textbf{Tianjun Xiao\textsuperscript{2}}\hfill
    \textbf{Carl-Johann Simon-Gabriel\textsuperscript{2}}\hfill
    \textbf{Tong He\textsuperscript{2}}\hfill
    \textbf{Zheng Zhang\textsuperscript{2}} \\
    \textbf{Bernhard Schölkopf\textsuperscript{2}}\hfill
    \textbf{Thomas Brox\textsuperscript{2}}\hfill
    \textbf{Francesco Locatello\textsuperscript{2}}\\
    \textsuperscript{1}Max-Planck Institute for Intelligent Systems, Tübingen, Germany \\
    \textsuperscript{2}Amazon Web Services\\
    \textsuperscript{3}Department of Computer Science, ETH Zürich\\
    \hspace{5in}
}

\makeatletter
\def\blfootnote{\xdef\@thefnmark{}\@footnotetext}
\makeatother

\DeclareMathOperator*{\mysoftmax}{softmax}

\newcommand{\mBOclass}{mBO$^c$\xspace}
\newcommand{\mBOinst}{mBO$^i$\xspace}
\newcommand{\dinosaur}{\texttt{DINOSAUR}\xspace}

\newcommand{\meanandstd}[3][s]{%
    \ifx s#1{\nprounddigits{1}\numprint{#2}{\relsize{-1.2}\,\,$\pm$\numprint{#3}}}\else  %
    \ifx m#1{\nprounddigits{1}\numprint{#2}}\else                          %
    \mathrm{Illegal~option}%
    \fi\fi
}
\newcommand{\meanandstdnoround}[3][s]{%
    \ifx s#1{{1}#2{\relsize{-1.2} $\pm$\!#3}}\else  %
    \ifx m#1{#2}\else                          %
    \mathrm{Illegal~option}%
    \fi\fi
}

\usepackage{todonotes}

\iclrfinalcopy %

\begin{document}

\doparttoc %
\faketableofcontents %
\part[Main Paper]{} %

\colorlet{mywhite}{white}
\colorlet{mygray}{white!60!black}
\colorlet{myblue}{blue!30!white!90!black}
\colorlet{myorange}{red!40!yellow!90!black}
\colorlet{mygreen}{green!60!black!60!white}
\colorlet{myred}{red!80!black}
\colorlet{mypurple}{myred!40!myblue}

\vspace{-2em}  %
\maketitle

\begin{abstract}
Humans naturally decompose their environment into entities at the appropriate level of abstraction to act in the world.
Allowing machine learning algorithms to derive this decomposition in an unsupervised way has become an important line of research.
However, current methods are restricted to simulated data or require additional information in the form of motion or depth in order to successfully discover objects.
In this work, we overcome this limitation by showing that reconstructing features from models trained in a self-supervised manner is a sufficient training signal for object-centric representations to arise in a fully unsupervised way.
Our approach, \dinosaur, significantly out-performs existing image-based object-centric learning models on simulated data and is the first unsupervised object-centric model that scales to real-world datasets such as COCO and PASCAL VOC.
\dinosaur is conceptually simple and shows competitive performance compared to more involved pipelines from the computer vision literature.
\end{abstract}

\section{Introduction}

\blfootnote{\hspace{-1em}$\dagger$: Work done during an internship at Amazon Web Services.}
\blfootnote{\hspace{-1em}Correspondence to: \texttt{hornmax@amazon.de}, \texttt{maximilian.seitzer@tuebingen.mpg.de}}
\looseness-1Object-centric representation learning has the potential to greatly improve generalization of computer vision models, as it aligns with causal mechanisms that govern our physical world~\citep{scholkopf2021toward,dittadi2021generalization}. Due to the compositional nature of scenes~\citep{greff2020binding}, object-centric representations can be more robust towards out-of-distribution data~\citep{dittadi2021generalization} and support more complex tasks like reasoning~\citep{assouel2022object,yang2020object} and control~\citep{zadaianchuk2020self,mambelli2022compositional,biza2022binding}.
They are in line with studies on the characterization of human perception and reasoning~\citep{kahneman1992reviewing, spelke2007core}.
Inspired by the seemingly unlimited availability of unlabeled image data, this work focuses on \emph{unsupervised} object-centric representation learning.

Most unsupervised object-centric learning approaches rely on a reconstruction objective, which struggles with the variation in real-world data.
Existing approaches typically implement \enquote{slot}-structured bottlenecks which transform the input into a set of object representations and a corresponding decoding scheme which reconstructs the input data.
The emergence of object representations is primed by the set bottleneck of models like Slot Attention~\citep{Locatello2020SlotAttention} that groups together independently repeating visual patterns across a fixed data set.
While this approach was successful on simple synthetic datasets, where low-level features like color help to indicate the assignment of pixels to objects, those methods have failed to scale to complex synthetic or real-world data~\citep{eslami2016attend, greff2019multi, burgess2019monet, Locatello2020SlotAttention,engelcke2021genesis2}.

To overcome these limitations, previous work has used additional information sources, \eg motion or depth~\citep{kipf2022conditional, elsayed2022savi++}.
Like color, motion and depth act as grouping signals when objects move or stand-out in 3D-space.
Unfortunately, this precludes training on most real-world image datasets, which do not include depth annotations or motion cues. %
Following deep learning's mantra of scale, another appealing approach could be to increase the capacity of the Slot Attention architecture.
However, our experiments (\sec{subsec:analysis}) suggest that scale alone is \emph{not} sufficient to close the gap between synthetic and real-world datasets.
We thus conjecture that the image reconstruction objective on its own does not provide sufficient inductive bias to give rise to object groupings when objects have complex appearance.
But instead of relying on auxiliary external signals, we introduce an additional inductive bias by reconstructing features that have a high level of homogeneity within objects.
Such features can easily be obtained via recent self-supervised learning techniques like DINO~\citep{Caron2021DINO}.
We show that combining such a feature reconstruction loss with existing grouping modules such as Slot Attention leads to models that significantly out-perform other image-based object-centric methods and \emph{bridge the gap to real-world object-centric representation learning}.
The proposed architecture \dinosaur~(\textbf{DINO} and \textbf{S}lot \textbf{A}ttention \textbf{U}sing \textbf{R}eal-world data) is conceptually simple and highly competitive with existing unsupervised segmentation and object discovery methods in computer vision.

\section{Related Work}

Our research follows a body of work studying the emergence of \emph{object-centric representations} in neural networks trained end-to-end with certain architectural biases~\citep{eslami2016attend,burgess2019monet,greff2019multi,Lin2020SPACE,Engelcke2020GENESIS,  Locatello2020SlotAttention,Singh2022SLATE}.
These approaches implicitly define objects as repeating patterns across a closed-world dataset that can be discovered \eg via semantic discrete- or set-valued bottlenecks.
As the grouping of low-level features into object entities is often somewhat arbitrary (it depends for example on the scale and level of detail considered), recent work has explored additional information sources such as video~\citep{Kosiorek2018SQAIR, Jiang2020SCALOR,weis2021benchmarking,Singh2022STEVE,traub2022learning}, optical flow~\citep{kipf2022conditional, elsayed2022savi++,Bao2022ObjectsThatMove}, text descriptions of the scene~\citep{Xu2022GroupViT} or some form of object-location information (\eg with bounding boxes)~\citep{kipf2022conditional}.
In contrast, we completely avoid additional supervision by leveraging the implicit inductive bias contained in the self-supervised features we reconstruct, which present a high level of homogeneity within objects~\citep{Caron2021DINO}.
This circumvents the scalability challenges of previous works that rely on pixel similarity as opposed to perceptual similarity~\citep{dosovitskiy2016generating} and enables object discovery on real-world data without changing the existing grouping modules.
Our approach can be considered similar to SLATE~\citep{Singh2022SLATE}, but with the crucial difference of reconstructing \emph{global} features from a Vision Transformer~\citep{Dosovitskiy2021ViT} instead of \emph{local} features from a VQ-VAE~\citep{Oord2017VQVAE}.

\looseness-1
Challenging object-centric methods by \emph{scaling dataset complexity} has been of recent interest:
\citet{Karazija2021clevrtex} propose ClevrTex, a textured variant of the popular CLEVR dataset, and show that previous object-centric models perform mostly poorly on it.
\citet{Greff2021Kubric} introduce the MOVi datasets with rendered videos of highly realistic objects with complex shape and appearance.
Arguably the most advanced synthetic datasets to date, we find that current state-of-the-art models struggle with them in the unsupervised setting.
Finally, \citet{yang2022promising} show that existing image-based object-centric methods catastrophically fail on real-world datasets such as COCO, likely because they can not cope with the diversity of shapes and appearances presented by natural data.
In contrast, we demonstrate that our approach works well on both complex synthetic and real-world datasets.

In the computer vision literature, structuring natural scenes without any human annotations has also enjoyed popularity, with tasks such as \emph{unsupervised semantic segmentation} and \emph{object localization}.
Those tasks are interesting for us because they constitute established real-world benchmarks related to unsupervised object discovery, and we show that our method is also competitive on them.
We refer to \app{app:expanded-related-work} for a detailed discussion of prior research in these areas.

\section{Method}
\label{sec:method}

\begin{figure}[t]
    \centering%
    \adjustbox{width=\textwidth,trim=6em 0 1em 0}{
    \input{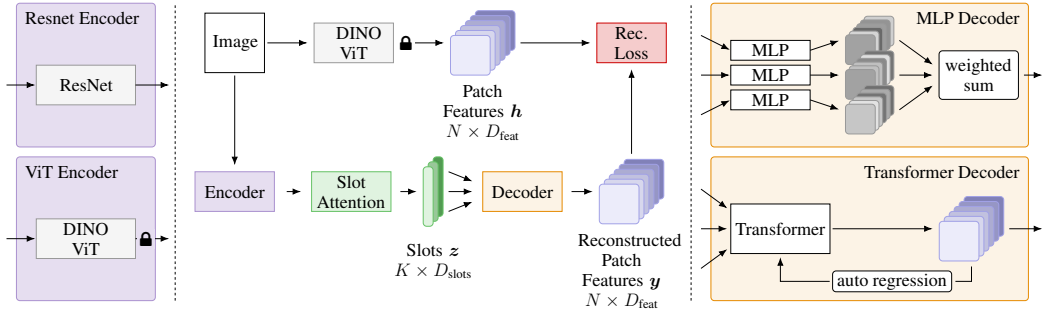}
    }
    \caption{\looseness-1
    Overview of the proposed architecture \dinosaur.
    The image is processed into a set of patch features $\vh$ by a frozen DINO ViT model (pre-trained using the self-supervised DINO method) and encoded via either a ResNet or the DINO ViT.
    Slot attention groups the encoded features into a set of slots.
    The model is trained by reconstructing the DINO features from the slots, either independently per-slot (MLP decoder) or jointly via auto regression (Transformer decoder).
    }
    \label{fig:method}
\end{figure}

Our approach essentially follows the usual autoencoder-like design of object-centric models and is summarized in \Fig{fig:method}:
a first module extracts features from the input data (the encoder), a second module groups them into a set of latent vectors called \emph{slots}, and a final one (the decoder) tries to reconstruct some target signal from the latents.
However, our method crucially differs from other approaches in that instead of reconstructing the original inputs, the decoder is tasked to reconstruct \emph{features from self-supervised pre-training}.
We start with the discussion of this training signal in \sec{subsec:training-signal} and describe further architectural choices in \sec{subsec:architecture}.

\subsection{Feature Reconstruction as a Training Signal}%
\label{subsec:training-signal}

Why are models based on image reconstruction like Slot Attention not successful beyond simpler synthetic datasets?
We hypothesize that reconstruction on the pixel level produces too weak of a signal for object-centricness to emerge; the task focuses (at least initially) strongly on low-level image features such as color statistics.
This quickly decreases the reconstruction error, but the resulting model does not discover objects beyond datasets where objects are mostly determined by distinct object colors.
Instead, if we had an (unsupervised) signal that required higher-level semantic information to reconstruct, there would be pressure on the slots to efficiently encode this information as well.
Luckily, such signals can nowadays be easily obtained with self-supervised learning algorithms, which have been successful in learning powerful representations for vision tasks such as classification and object detection purely from images~\citep{Chen2020SimCLR,Grill2020BYOL,He2022MAE}.
Thus, given $K$ slots $\vz \in \mathbb{R}^{K \times D_\text{slots}}$, the model is trained to reconstruct self-supervised features $\vh \in \mathbb{R}^{N \times D_\text{feat}}$, by minimizing the following loss:
\begin{equation}
    \mathcal{L}_\text{rec} = \lVert \vy - \vh\rVert^2, \quad\quad\quad\vy=\mathrm{Decoder}(\vz).
\end{equation}
This loss can be viewed as a form of student-teacher knowledge distillation~\citep{Hinton2015DistillingTK}, where the student has a particular form of bottleneck that condenses the high-dimensional, unstructured information contained in the teacher features into a lower-dimensional, structured form.
We can also draw parallels between this loss and perceptual similarity losses for image generation~\citep{dosovitskiy2016generating}, that is, the optimization takes place in a space more semantic than pixel space.

For pre-training, we utilize the ImageNet dataset~\citep{Deng2009ImageNet}.
From the student-teacher perspective, this means that the teacher additionally transports knowledge gained from a larger image collection to the (smaller) datasets at hand.
It is well-known that using large datasets for pre-training can significantly improve performance, but to our knowledge, we are the first to exploit such transfer learning for object-centric learning.
In general, studying the role additional \emph{data} can play for object-centric learning is an interesting topic, but we leave that for future investigations.

\emph{Which self-supervised algorithm should we use?}
In our analysis (\sec{subsec:analysis}), we investigate several recent ones (DINO~\citep{Caron2021DINO}, MoCo-v3~\citep{Chen2021MoCov3}, MSN~\citep{Assran2022MSN}, MAE~\citep{He2022MAE}).
Interestingly, we find that they all work reasonably well for the emergence of real-world object grouping.
In the following, we mainly apply the DINO method~\citep{Caron2021DINO}, because of its good performance and accessibility in open source libraries~\citep{Wightman2019timm}.
We experiment with features from ResNets~\citep{He2015ResNet} and Vision Transformers (ViTs)~\citep{Dosovitskiy2021ViT}, and find that the latter yield significantly better results.

\subsection{An Architecture for Real-World Object-Centric Learning}
\label{subsec:architecture}

\paragraph{Encoder}
Previous work has shown that powerful feature extractors help in scaling object-centric methods to more complex data~\citep{kipf2022conditional}.
To this end, we experiment with two choices: a ResNet-34 encoder with increased spatial resolution used by~\cite{kipf2022conditional}, and Vision Transformers.
Unfortunately, we were not able to optimize randomly initialized ViTs with our model, as training collapsed.
Instead, we found it sufficient to initialize the ViT using weights from self-supervised pre-training, and keeping them fixed throughout training\footnote{Another option would be further finetuning the pre-trained ViT, but we found that this leads to slots that do not focus on objects. Combining ViT training with Slot Attention might require very careful training recipes.}.
In terms of results, we find that the ResNet and the pre-trained ViT encoder perform similarly.
However, the model converges faster with the pre-trained ViT, and it is also computationally more efficient: we can directly use the ViT outputs as the target features $\vh$.
Consequently, we mainly use the ViT encoder in the following.

\paragraph{Slot Attention Grouping}
The grouping stage of our model uses Slot Attention~\citep{Locatello2020SlotAttention} to turn the set of encoder features into a set of $K$ slot vectors $\vz \in \mathbb{R}^{K \times D_\text{slots}}$.
This follows an iterative process where slots compete for input features using an attention mechanism, starting from randomly sampled initial slots.
We largely use the original Slot Attention formulation (including GRU~\citep{Cho2014GRU} and residual MLP modules), with one difference when using ViT features: we do not add positional encodings on the ViT features before Slot Attention, as we found the ViT's initial position encodings to be sufficient to support spatial grouping of the features.
Additionally, we add a small one-hidden-layer MLP that transforms each encoder feature before Slot Attention.

\paragraph{Feature Decoding}
\looseness=-1As we apply feature instead of image reconstruction as the training objective, we need a decoder architecture suitable for this purpose.
To this end, we consider two different designs: a MLP decoder that is applied independently to each slot, and a Transformer decoder~\citep{Vaswani2017Attention} that autoregressively reconstructs the set of features.
We describe both options in turn.

The \emph{MLP decoder} follows a similar design as the commonly used spatial broadcast decoder~\citep{Watters2019SpatialBD}.
Each slot is first broadcasted to the number of patches, resulting in a set of $N$ tokens for each slot.
To make the spatial positions of the tokens identifiable, a learned positional encoding is added to each token.
The tokens for each slot are then processed token-wise by the same MLP, producing the reconstruction $\hat{\bm{y}}_{k}$ for slot $k$, plus an alpha map $\bm{\alpha}_{k}$ that signifies where the slot is active.
The final reconstruction $\vy \in \mathbb{R}^{N \times D_\text{feat}}$ is formed by taking a weighted sum across the slots:
\begin{equation}
\vy = \sum_{k=1}^K \hat{\bm{y}}_{k} \odot \vm_{k}, \quad\quad\quad\vm_k = \mysoftmax_k \bm{\alpha}_k
\end{equation}
The advantage of this simple design is its computational efficiency: as the MLP is shared across slots and positions, decoding is heavily parallelizable.

\looseness-1 The \emph{Transformer decoder}~\citep{Vaswani2017Attention} reconstructs features $\vy$ jointly for all slots in an autoregressive manner.
In particular, the feature at position $n$ is generated while conditioning on the set of previously generated features $\vy_{<n}$ \emph{and} the set of slots $\vz$: $\vy_n = \mathrm{Decoder}(\vy_{<n}; \vz)$.
This decoder design is more powerful than the MLP decoder as it can maintain global consistency across the reconstruction, which might be needed on more complex data.
However, we found several drawbacks of the Transformer decoder: it does not work with training ResNet encoders from scratch, higher resolution target features (see \app{app:subsec:exp-transformer-decoder}), and requires more effort to tune (see \app{app:subsec:exp-decoder-analysis}).
Thus, we recommend using the MLP decoder as the first choice when applying \dinosaur to a new dataset.
We note that Transformer decoders have also been previously explored by SLATE~\citep{Singh2022SLATE} and STEVE~\citep{Singh2022STEVE}, but to reconstruct the discrete token map of a VQ-VAE~\citep{Oord2017VQVAE}.

\paragraph{Evaluation} \looseness=-1Object-centric methods are commonly evaluated by inspecting masks associated with each slots.
Previous approaches reconstructing to image-level typically use the decoder's alpha mask for this purpose; for the MLP decoder, we also make use of this option.
The Transformer decoder does not produce an alpha mask.
Instead, we have two options: the attention masks of Slot Attention (used by SLATE), or the decoder's attention mask over the slots.
We found that the latter performed better (see \sec{app:subsec:exp-mask-type}), and we use it throughout.
As the masks from feature reconstruction are of low resolution, we bilinearly resize them to image resolution before comparing them to ground truth masks.

\section{Experiments}
\label{sec:experiments}

Broadly, we pursue two goals with our experiments: 1) demonstrating that our approach significantly extends the capabilities of object-centric models towards real-world applicability (\sec{subsec:unsup-object-discovery}), and 2) showing that our approach is competitive with more complex methods from the computer vision literature (\sec{subsec:comp-cv-methods}).
Additionally, we ablate key model components to find what is driving the success of our method (\sec{subsec:analysis}).
The main task we consider in this work is \emph{object discovery}, that is, finding pixel masks for all object instances in an image.

\vspace{-0.5em}
\paragraph{Datasets}\looseness-1
We consider two synthetic and two real-world image datasets.
As synthetic datasets, we use the MOVi datasets~\citep{Greff2021Kubric}, recently introduced as challenging testbeds for object-centric methods.
In particular, we use the variants MOVi-C and MOVi-E, which contain around \numprint{1000} realistic 3D-scanned objects on HD backgrounds.
For our purposes, the main difference is that MOVi-C contains 3--10, and MOVi-E 11--23 objects per scene.
Note that we treat the video-based MOVi datasets as image datasets by randomly sampling frames.
As real-world datasets, we use PASCAL VOC 2012~\citep{Everingham2012PASCALVOC} and MS COCO 2017~\citep{Lin2014COCO}, commonly used for object detection and segmentation.
Whereas PASCAL VOC contains many images with only a single large object, COCO consists of images with at least two and often dozens of objects.
Both datasets represent a significant step-up in complexity to what object-centric models have been tested on so far.
In \app{app:results-kitti}, we also report preliminary results on the KITTI driving dataset.

\vspace{-0.5em}
\paragraph{Training Details}
We train \dinosaur using the Adam optimizer~\citep{kingma2014adam} with a learning rate of $4 \cdot 10^{-4}$, linear learning rate warm-up of \numprint{10000} optimization steps and an exponentially decaying learning rate schedule.
Further, we clip the gradient norm at $1$ in order to stabilize training and train for 500k steps for the MOVI and COCO datasets and 250k steps for PASCAL VOC.
The models were trained on 8 NVIDIA V100 GPUs with a local batch size of $8$, with 16-bit mixed precision.
For the experiments on synthetic data, we use a ViT with patch size 8 and the MLP decoder.
For the experiments on real-world data, we use a ViT with patch size 16 and the Transformer decoder.
We analyze the impact of different decoders in \sec{subsec:analysis}.
The main results are averaged over 5 random seeds; other experiments use 3 seeds.
Further implementation details can be found in \app{app:subsec-detail-architecture}.

\subsection{Comparison to Object-Centric Learning Methods}
\label{subsec:unsup-object-discovery}

Our goal in this section is two-fold: 1) demonstrating that previous object-centric methods fail to produce meaningful results on real-world datasets and struggle even on synthetic datasets, and 2) showcase how our approach of incorporating strong pre-trained models results in a large step forward for object-centric models on both kinds of datasets.

\begin{figure}[tb]
    \centering
    {
\centering
\newcommand{\myig}[1]{\includegraphics[width=0.13\textwidth,valign=c]{#1}}
\newcommand{\mycaption}[1]{{\rotatebox[origin=c]{90}{\footnotesize#1}}}
\renewcommand{\arraystretch}{3.9}
\setlength{\tabcolsep}{1pt}
\begin{tabular}{@{}rccccccc@{}}
    & \multicolumn{3}{c}{\textbf{MOVi-C}} & \phantom{xxx} & \multicolumn{3}{c}{\textbf{MOVi-E}} \\[-1em]
    \mycaption{Image} & \myig{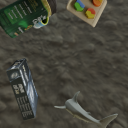} & \myig{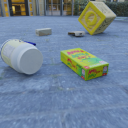} & \myig{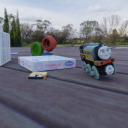} & \phantom & \myig{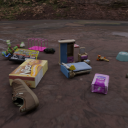} & \myig{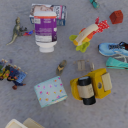} & \myig{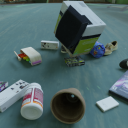} \\
    \mycaption{\dinosaur} & \myig{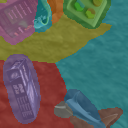} & \myig{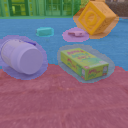} & \myig{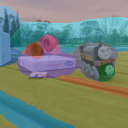} & \phantom & \myig{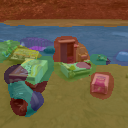} & \myig{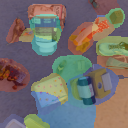} & \myig{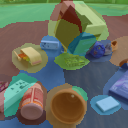} \\
    \mycaption{SLATE} & \myig{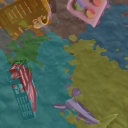} & \myig{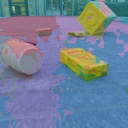} & \myig{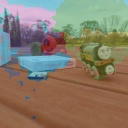} & \phantom & \myig{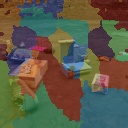} & \myig{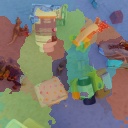} & \myig{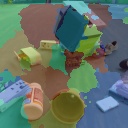} \\
    \mycaption{Slot Attention} & \myig{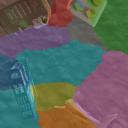} & \myig{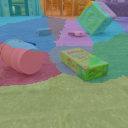} & \myig{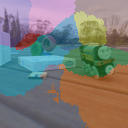} & \phantom & \myig{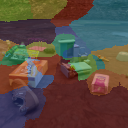} & \myig{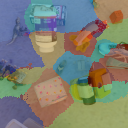} & \myig{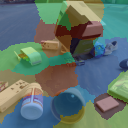} \\
\end{tabular}
}
    \caption{Example results on the synthetic MOVi-C and MOVi-E datasets~\citep{Greff2021Kubric}.}
    \label{fig:results-object-discovery-synthetic}
\end{figure}

\begin{figure}[tb]
    \centering
    \includegraphics[width=\textwidth]{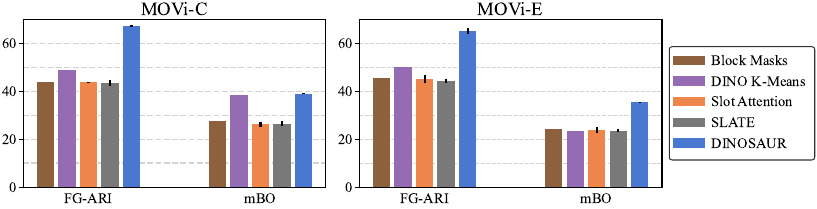}
    \vspace{-1.5em}
    \caption{Object Discovery on synthetic datasets (mean $\pm$ standard dev., 5 seeds) with 11 (MOVi-C) and 24 slots (MOVi-E). We report foreground adjusted rand index (FG-ARI) and mean best overlap (mBO).
    \dinosaur uses a ViT-B/8 encoder with the MLP decoder.}
    \label{fig:results-object-discovery-synthetic-quant}
\end{figure}

\textbf{Tasks\quad}\looseness-1
We evaluate on the task object-centric models are most frequently tested on: object discovery~\citep{burgess2019monet}, that is, producing a set of masks that cover the independent objects appearing on an image.
We also present preliminary results testing the quality of the learned representations on the COCO dataset in \app{app:subsec:coco-property-prediction}, though this is not the main focus of our work.

\textbf{Metrics\quad}
As common in the object-centric literature, we evaluate this task using foreground adjusted rand index (FG-ARI), a metric measuring cluster similarity.
Additionally, we compute a metric based on intersection-over-union (IoU), the mean best overlap (mBO)~\citep{Arbelaez2014MCG}.
mBO is computed by assigning each ground truth mask the predicted mask with the largest overlap, and then averaging the IoUs of the assigned mask pairs.
In contrast to ARI, mBO takes background pixels into account, thus also measuring how close masks fit to objects.
On datasets where objects have a semantic label attached (\eg on COCO), we can evaluate this metric with instance-level (\ie object) masks, and semantic-level (\ie class) masks.
This allows us to find model preferences towards instance- or semantic-level groupings.

\textbf{Baselines\quad}
We compare our approach to a more powerful version of Slot Attention~\citep{Locatello2020SlotAttention} based on a ResNet encoder that has been shown to scale to more complex data~\citep{elsayed2022savi++}.
Further, we compare with SLATE~\citep{Singh2022SLATE}, a recent object-centric model that trains a discrete VQ-VAE~\citep{Oord2017VQVAE} as the feature extractor and a Transformer as the decoder.
We refer to \app{app:subsec:details-baselines} for details about baseline configurations.

\looseness-1
As it can be hard to gauge how well object-centric methods perform on new datasets solely from metrics, we add one trivial baseline:
dividing the image into a set of regular blocks.
These \emph{block masks} (see \fig{app:fig:block_patterns}) thus show the performance of a method that only follows a geometric strategy to group the data, completely ignoring the semantic aspects of the image.
Familiar to practitioners, this is a common failure mode of object-centric methods, particularly of Slot Attention.
Last, we apply the \emph{K-Means algorithm} on the DINO features and use the resulting clustering to generate spatial masks.
This baseline shows to which extent objects are already trivially extractable from self-supervised features.

\textbf{Results on Synthetic Datasets (\fig{fig:results-object-discovery-synthetic} and \fig{fig:results-object-discovery-synthetic-quant})\quad}
Both Slot Attention and SLATE struggle on the challenging MOVi datasets, performing similar to the naive block masks and worse than the K-Means baselines.
Our model achieves good performance on both MOVI-C and MOVi-E.
In \app{app:subsec:comparison-video-methods}, we also find that our method compares favorably to video methods that can use temporal information and/or weak supervision~\citep{elsayed2022savi++,Singh2022STEVE}

\begin{figure}[tb]
    \centering
    {
\centering
\newcommand{\myig}[1]{\includegraphics[width=0.125\textwidth,valign=c]{#1}}
\newcommand{\mycaption}[1]{{\rotatebox[origin=c]{90}{\footnotesize#1}}}
\renewcommand{\arraystretch}{3.8}
\setlength{\tabcolsep}{1pt}
\begin{tabular}{@{}rccccccc@{}}
    \mycaption{Ground Truth} & \myig{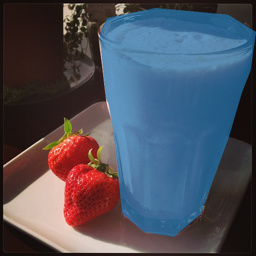} & \myig{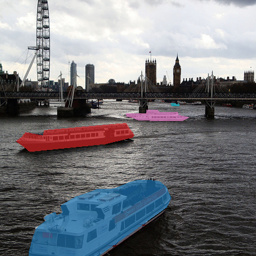} & \myig{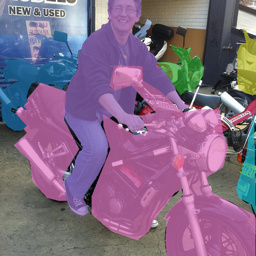} & \myig{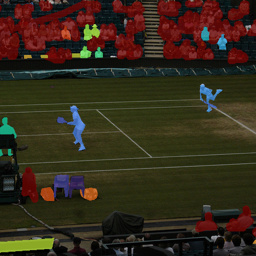} & \myig{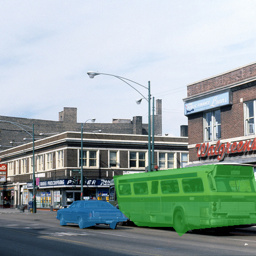} & \myig{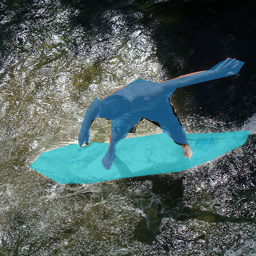} & \myig{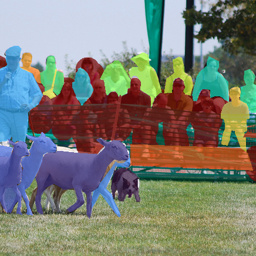} \\
    \mycaption{\dinosaur} & \myig{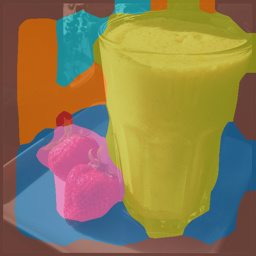} & \myig{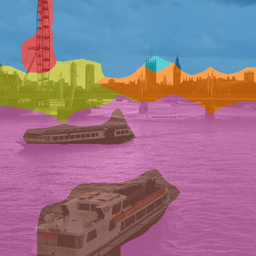} & \myig{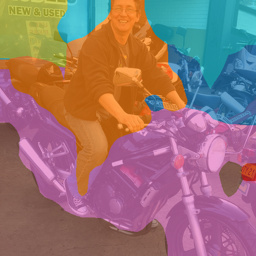} & \myig{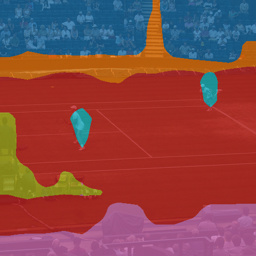} & \myig{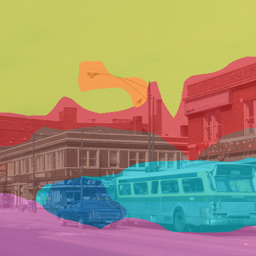} & \myig{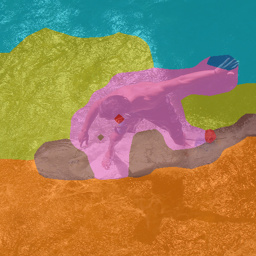} & \myig{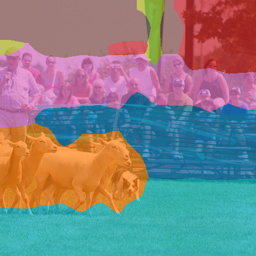} \\
    \mycaption{SLATE} & \myig{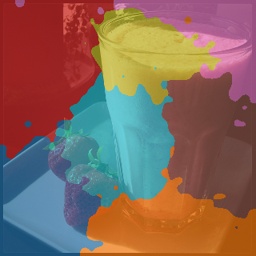} & \myig{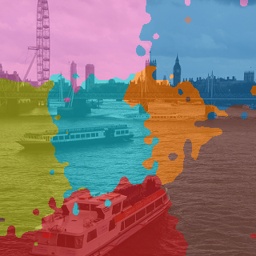} & \myig{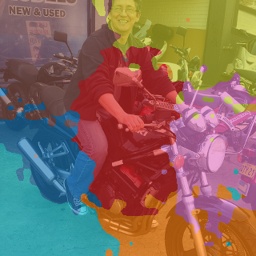} & \myig{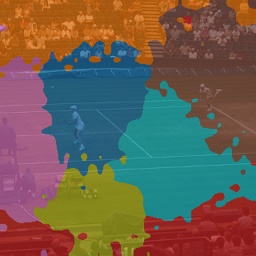} & \myig{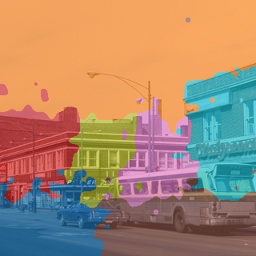} & \myig{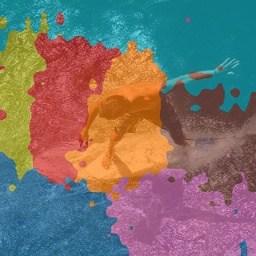} & \myig{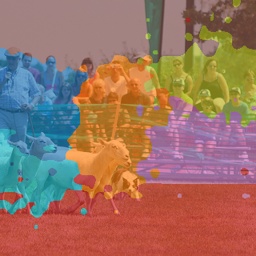} \\
    \mycaption{Slot Attention} & \myig{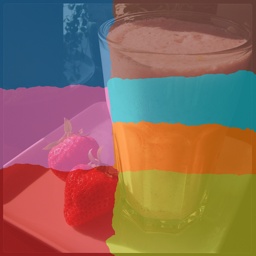} & \myig{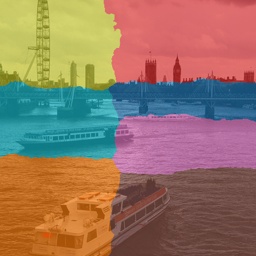} & \myig{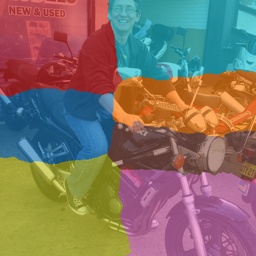} & \myig{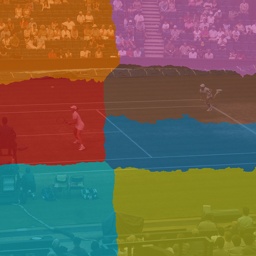} & \myig{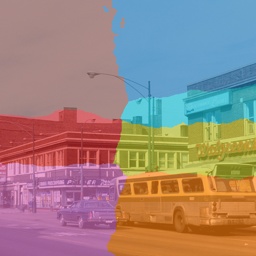} & \myig{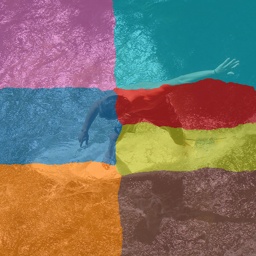} & \myig{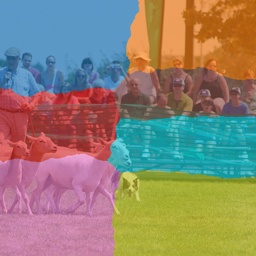} \\
\end{tabular}
}
    \caption{Example reults on COCO 2017, using 7 slots.
    Additional examples are provided in \app{app:sec:additional-examples}.
    }
    \label{fig:results-object-discovery-real}
\end{figure}

\begin{figure}[tb]
    \centering
    \includegraphics[width=\textwidth]{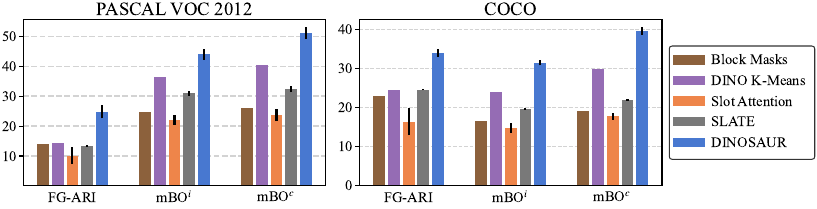}
    \vspace{-1.5em}
    \caption{Object Discovery on real-world datasets (mean $\pm$ standard dev., 5 seeds) with 6 (PASCAL) and 7 slots (COCO). We report foreground adjusted rand index (FG-ARI) and instance/class mean best overlap (\mBOinst/\mBOclass). \dinosaur uses a ViT-B/16 encoder with the Transformer decoder.}
    \label{fig:results-object-discovery-real-quant}
\end{figure}

\textbf{Results on Real-World Datasets (\fig{fig:results-object-discovery-real} and \fig{fig:results-object-discovery-real-quant})\quad}%
As expected, Slot Attention can not handle the increased complexity of real-world data and degrades to non-semantic grouping patterns.
For SLATE, semantic grouping begins to emerge (\eg of backgrounds), but not consistently; it still performs worse than the K-Means baseline.
Note that it was necessary to early-stop SLATE's training as performance would degrade to Slot Attention's level with more training.
In contrast, \dinosaur captures a variety of objects of different size, appearance and shape.
To the best of our knowledge, we are the first to show a successful version of an object-centric model on unconstrained real-world images in the fully unsupervised setting.
Our result represents a significant step-up in complexity of what object-centric methods can handle.
Note that the examples in \fig{fig:results-object-discovery-real} show mostly semantic rather than instance grouping emerging: this is a by-product of using the Transformer decoder.
In contrast, the MLP decoder is biased towards instance grouping, an effect which we analyze in \sec{subsec:analysis}.

\subsection{Comparison to Computer Vision Methods}
\label{subsec:comp-cv-methods}

In this section, our goal is to show that our method fares well on two tasks closely related to object discovery from the computer vision literature: unsupervised object localization and segmentation.
Being competitive on these benchmarks is difficult, as there has been a stream of methods with quickly improving results recently~\citep{Wang2022TokenCut, Hamilton2022Stego,Zadaianchuk2022COMUS}.
Due to space issues, we defer most of the discussion to \app{app:comparison-cv-methods}.

\looseness-1
\textbf{Tasks, Metrics and Baselines}\quad
We briefly introduce the two tasks: in \emph{object localization}\footnote{{This task is often called ``object discovery'' in the literature as well, but we term it ``object localization'' in this work in order to avoid confusion with the task evaluated in the object-centric literature.}}, the goal is to find object location and size by predicting bounding boxes, and in \emph{unsupervised semantic segmentation} the goal is to separate the image into semantically consistent labeled regions.
For the latter, we consider two variations: object segmentation, where only foreground objects should get segmented and labeled, and scene decomposition, where each pixel of the image has to be labeled with a semantic class.
We evaluate object localization in terms the fraction of
images on which at least one object was correctly localized (CorLoc)~\citep{Vo2020OSD}, and semantic segmentation in terms of mean intersection-over-union over classes (mIoU).
For semantic segmentation, we obtain class labels by running K-Means clustering on features associated with each slot after training the model, then
assigning clusters to ground truth classes by maximizing IoU using Hungarian matching, similar to~\citet{VanGansbeke2021MaskContrast} (see \app{app:comparison-cv-methods} for details).
On each task, we compare with the current state-of-the-art~\citep{Wang2022TokenCut, Hamilton2022Stego,Zadaianchuk2022COMUS}, and a recent, but competitive method~\citep{VanGansbeke2021MaskContrast,Melaskyriazi2022DeepSpectral,Wen2022SlotCon}.

\textbf{Results (\tab{tab:results-cv-methods})\quad}
For object localization, our method reaches comparable results to what has been previously reported.
For object segmentation, our method falls behind the state-of-the-art, though it is still competitive with other recent work.
Note that the best methods on this task employ additional steps of training segmentation networks which improves results and allows them to run at the original image resolution.
In contrast, the masks we evaluate are only of size $14 \times 14$; we leave it to future work to improve the resolution of the produced masks.
For the task of scene decomposition, DINOSAUR comes close to the current state-of-the-art.
All in all, our method is competitive with often more involved methods on these benchmarks, demonstrating a further step towards real-world usefulness of object-centric methods.

\begin{table}[tb]
    \caption{Representative comparisons on three tasks from the computer vision literature.
    We refer to \app{app:comparison-cv-methods} for a detailed discussion including more datasets, baselines, and metrics.
    Here, we compare with (a) DeepSpectral \citep{Melaskyriazi2022DeepSpectral} and TokenCut \citep{Wang2022TokenCut}, (b) MaskContrast \citep{VanGansbeke2021MaskContrast} and COMUS \citep{Zadaianchuk2022COMUS}, and (c) SlotCon~\citep{Wen2022SlotCon} and STEGO~\citep{Hamilton2022Stego}.
    DINOSAUR uses a ViT-B/16 encoder with the Transformer decoder (mean $\pm$ standard dev., 5 seeds).
    }
    \label{tab:results-cv-methods}

    \begin{subtable}[t]{.33\linewidth}
        \centering
        \caption{Unsup.~Object Localization.}
        \label{tab:results-object-loc-short}
        \vspace{-0.3em}
        \begin{tabular}{@{}lc@{}}
        \toprule
        \multicolumn{2}{c}{\textbf{COCO-20k (CorLoc)}} \\
        \midrule
        DeepSpectral & 52.2 \\
        TokenCut & 58.8 \\
        \midrule
        \dinosaur & \meanandstd{67.15}{1.46} \\
        \bottomrule
        \end{tabular}
    \end{subtable}\hfill
    \begin{subtable}[t]{.33\linewidth}
        \centering
        \caption{Unsup.~Object Segmentation.}
        \label{tab:results-unsup-object-seg-short}
        \vspace{-0.3em}
        \begin{tabular}{@{}lc@{}}
        \toprule
        \multicolumn{2}{c}{\textbf{PASCAL VOC 2012 (mIoU)}} \\
        \midrule
        MaskContrast & 35.0 \\
        COMUS & 50.0 \\
        \midrule
        \dinosaur & \meanandstd{37.15}{1.84} \\
        \bottomrule
        \end{tabular}
    \end{subtable}\hfill
    \begin{subtable}[t]{.33\linewidth}
      \centering
        \caption{Unsup.~Scene Decomposition.}
        \label{tab:results-unsup-scene-decomp-short}
        \vspace{-0.3em}
        \begin{tabular}{@{}lc@{}}
        \toprule
        \multicolumn{2}{c}{\textbf{COCO-Stuff 27 (mIoU)}} \\
        \midrule
        SlotCon & 18.3 \\
        STEGO & 26.8 \\
        \midrule
        \dinosaur & \meanandstd{23.99}{0.85} \\
        \bottomrule
        \end{tabular}
    \end{subtable}
\end{table}

\subsection{Analysis}
\label{subsec:analysis}

In this section, we analyze different aspects of our approach: the importance of feature reconstruction, the impact of the method for self-supervised pre-training, and the role of the decoder.
Additional experiments are included in \app{app:sec:additional-experiments}.

\vspace{-0.75em}
\paragraph{Insufficiency of Image Reconstruction}\looseness-1
We first test the hypothesis if a scaled-up Slot Attention model trained with image reconstruction could lead to real-world object grouping.
Our experiments from~\sec{subsec:unsup-object-discovery} already show that a ResNet encoder is not sufficient.
We additionally test a ViT-B/16 encoder under different training modes: training from scratch, frozen, or finetuning DINO pre-trained weights.
We find that training from scratch results in divergence of the training process, and that both the frozen and finetuning setting fail to yield meaningful objects, resulting in striped mask patterns (see \fig{app:fig:analysis-image-recon}).
Thus, even when starting from features that are highly semantic, image reconstruction does not give enough signal towards semantic grouping.

\vspace{-0.75em}
\paragraph{ResNet and Pre-Trained ViT Encoders Perform Similar}\looseness-1
Second, we analyze whether \emph{pre-training the encoder} plays a crucial role for our method.
To do so, we compare ResNet34 encoders trained from scratch with pre-trained ViT encoders, and find that the performance is overall similar (see \tab{app:tab:analysis-encoder-scratch}).
This also suggests that the feature reconstruction signal is the key component in our approach that allows object-centricness to emerge on real-world data.
We expand in \app{app:subsec:training-encoder-scratch}.

\begin{figure}[tbh]
    \begin{minipage}[t]{0.48\textwidth}
        \centering
        \captionof{table}{Comparing self-supervised reconstruction targets produced by a ViT-B/16 on COCO object discovery, with a ResNet34 encoder and the MLP decoder.}
        \label{tab:analysis-self-supervised-targets}
        \small
        \begin{tabular}{@{}lccc@{}}
        \toprule
        Algorithm & FG-ARI & \mBOinst & \mBOclass \\
        \midrule
        DINO & \meanandstd{40.91}{0.19} & \meanandstd{27.93}{0.04} & \meanandstd{31.14}{0.05}  \\
        MoCo-v3 & \meanandstd{40.40}{0.57} & \meanandstd{28.06}{0.16} & \meanandstd{31.12}{0.11} \\
        MSN & \meanandstd{40.73}{0.30} & \meanandstd{27.57}{0.12} & \meanandstd{30.65}{0.12} \\
        MAE & \meanandstd{37.73}{0.12} & \meanandstd{28.10}{0.07} & \meanandstd{31.70}{0.03} \\
        \bottomrule
        \end{tabular}
    \end{minipage}\hfill
    \begin{minipage}[t]{0.48\textwidth}
        \centering
        \captionof{table}{Comparing different decoders on object discovery, with a ViT-B/16 encoder. We also list mean squared reconstruction error (MSE).}
        \label{tab:analysis-decoders}
        \small
        \setlength{\tabcolsep}{0.5em}
        \begin{tabular}{@{}llcc@{\hskip 0.6em}cc@{}}
        \toprule
        Dataset & Decoder & ARI & \multicolumn{2}{c}{mBO$^{(i,c)}$} & MSE \\
        \midrule
        \multirow{2}{*}{MOVi-C} &
        MLP & \meanandstd[m]{65.97}{0.43} & \multicolumn{2}{c}{\meanandstd[m]{35.00}{0.16}} & \meanandstdnoround[m]{0.24}{0.0001} \\
        & Transformer & \meanandstd[m]{55.65}{2.62} & \multicolumn{2}{c}{\meanandstd[m]{42.36}{1.08}} & \meanandstdnoround[m]{0.14}{0.0010} \\
        \midrule
        \multirow{2}{*}{PASCAL} &
        MLP & \meanandstd[m]{24.61}{0.21} & \meanandstd[m]{39.54}{0.12} & \meanandstd[m]{40.85}{0.10} & \meanandstdnoround[m]{0.33}{0.0000} \\
        & Transformer & \meanandstd[m]{24.84}{2.21} & \meanandstd[m]{44.02}{1.85} & \meanandstd[m]{51.18}{1.87} & \meanandstdnoround[m]{0.17}{0.0007} \\
        \midrule
        \multirow{2}{*}{COCO} &
        MLP & \meanandstd[m]{40.54}{0.03} & \meanandstd[m]{27.67}{0.19} & \meanandstd[m]{30.87}{0.21} & \meanandstdnoround[m]{0.31}{0.0002} \\
        & Transformer & \meanandstd[m]{34.14}{1.03} & \meanandstd[m]{31.63}{0.69} & \meanandstd[m]{39.72}{0.94} & \meanandstdnoround[m]{0.16}{0.0016} \\
        \bottomrule
        \end{tabular}
    \end{minipage}
\end{figure}

\vspace{-0.75em}
\paragraph[Choice of Self-Supervised Targets]{Choice of Self-Supervised Targets (\tab{tab:analysis-self-supervised-targets})}
We now analyze the role of the self-supervised pre-training algorithm.
To this end, we train \dinosaur with a ResNet34 encoder (from scratch) on COCO, but reconstruct targets obtained from ViTs pre-trained with different methods: DINO, MoCo-v3, MSN, and MAE.
Remarkably, all self-supervised schemes perform well for the task of object discovery (examples in \fig{app:fig:examples-coco-encoders}).
This demonstrates that self-supervised pre-training on ImageNet translates into a useful, general bias for discovering objects.

\vspace{-0.75em}
\paragraph[Choice of Decoder]{Choice of Decoder (\tab{tab:analysis-decoders})}

\begin{figure}[t]
    \centering
    \begin{minipage}[c]{0.48\textwidth}
        \centering
        \vspace{0.5em}
        \includegraphics{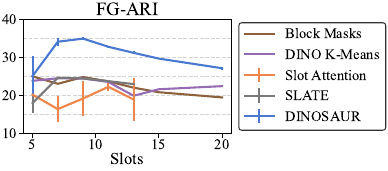}%
        \captionof{figure}{Sensitivity to number of slots on COCO object discovery (see also \app{app:subsec:exp-number-of-slots}).}
        \label{fig:slot-sensitivity}
    \end{minipage}\hfill
    \begin{minipage}[c]{0.48\textwidth}
        \centering
        {
\centering
\newcommand{\myig}[1]{\includegraphics[width=0.2\textwidth,valign=c]{#1}}
\newcommand{\mycaption}[1]{{\rotatebox[origin=c]{90}{\makecell{\footnotesize#1}}}}
\renewcommand{\arraystretch}{3}
\setlength{\tabcolsep}{1.5pt}
\begin{tabular}{@{}rcccc@{}}
    \mycaption{MLP\\[-0.25em]\footnotesize Decoder} & \myig{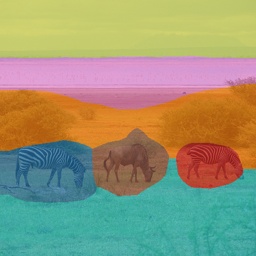} & \myig{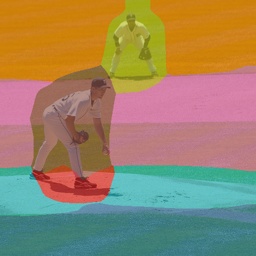} & \myig{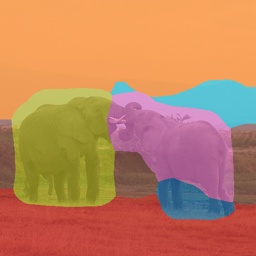} & \myig{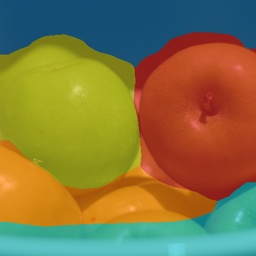} \\
    \mycaption{Transformer\\[-0.25em]\footnotesize Decoder} & \myig{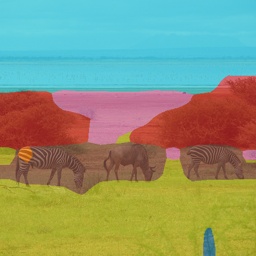} & \myig{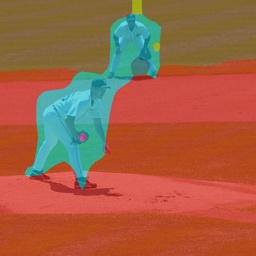} & \myig{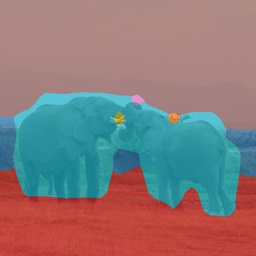} & \myig{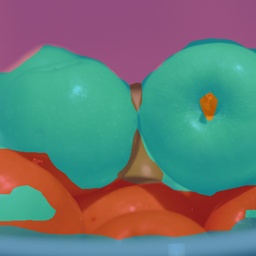} \\
\end{tabular}
}
        \vspace{0.5em}
        \captionof{figure}{MLP and Transformer decoder have different biases in how they group objects.}
        \label{fig:analysis-decoders}
    \end{minipage}
\end{figure}

We compare the choice of MLP \vs Transformer decoder for object discovery.
Both options use a ViT-B/16 encoder.
Generally, we find that the MLP decoder is better on ARI whereas the Transformer decoder is better on mBO.
For MOVi-C, visual inspection (see \fig{app:fig:examples-extended-movi-c}) shows that the Transformer tends to produce tighter masks and a cleaner background separation, but uses excess slots to split objects.
For PASCAL and COCO, what's striking is the Transformer decoders' improvement of 9--10 class mBO.
This reveals that the Transformer decoder is biased towards grouping semantically related instances into the same slot, which we suggest stems from its global view on the image, but also from the generally increased expressiveness of the architecture (\cf the lower reconstruction loss).
In contrast, the MLP decoder is able to separate instances better (see \fig{fig:analysis-decoders} and also \fig{app:fig:examples-extended-coco}), which is reflected in the higher ARI scores.
Researching how different decoder designs affect semantic \vs instance-level grouping is an interesting avenue for future work.
In \app{app:subsec:exp-decoder-analysis} and \app{app:subsec:exp-transformer-decoder}, we further study different decoder properties. %

\section{Conclusion}

We presented the first image-based fully unsupervised approach for object-centric learning that scales to real-world data.
Our experiments demonstrate significant improvements on both simulated and real-world data compared to previously suggested approaches and even achieve competitive performance with more involved pipeline methods from the computer vision literature.

This work only takes a first step towards the goal of representing the world in terms of objects.
As such, some problems remain open.
One issue concerns semantic \vs instance-level grouping.
As evident from the presented examples, our approach covers a mix of both, with semantically related objects sometimes being grouped into a single slot.
While we found the type of decoder to influence this behavior, more fine-grained control is needed.
A related issue is the detail of the decomposition, \eg whether objects are split into parts or stay whole.
We found this to be dependent on the number of slots, with a fixed number often being inappropriate (see \fig{fig:slot-sensitivity} and \app{app:subsec:exp-number-of-slots}).
How models can dynamically choose a suitable level of detail while staying unsupervised but controllable will be an important challenge to fully master the ambiguities the real world inherently presents.

In this work, we mainly focused on object discovery.
Future work could further examine the properties of the learned slot representations, for instance robustness to distribution shifts, generalization and usefulness for downstream tasks.
Another interesting direction is how our approach can be combined with image generation to build flexible and compositional generative models of natural data.

\newpage

\section*{Reproducibility Statement}
\App{app:subsec-detail-architecture} contains detailed information about the \dinosaur architecture and all hyperparamers used for all experiments.
\App{app:subsec:details-baselines} contains details about how baselines were trained.
\App{app:sec:dataset-and-eval-settings} contains information about task evaluation and datasets.
All datasets used in this work (MOVi, PASCAL VOC 2012, COCO, KITTI) are public and can be obtained on their respective web pages.
Source code will be made available under \url{https://github.com/amazon-science/object-centric-learning-framework}.

\bibliography{iclr2023_conference}

\begin{thebibliography}{79}
\providecommand{\natexlab}[1]{#1}
\providecommand{\url}[1]{\texttt{#1}}
\expandafter\ifx\csname urlstyle\endcsname\relax
  \providecommand{\doi}[1]{doi: #1}\else
  \providecommand{\doi}{doi: \begingroup \urlstyle{rm}\Url}\fi

\bibitem[Alayrac et~al.(2022)Alayrac, Donahue, Luc, Miech, Barr, Hasson, Lenc,
  Mensch, Millican, Reynolds, Ring, Rutherford, Cabi, Han, Gong, Samangooei,
  Monteiro, Menick, Borgeaud, Brock, Nematzadeh, Sharifzadeh, Binkowski,
  Barreira, Vinyals, Zisserman, and Simonyan]{Alayrac2022Flamingo}
Jean-Baptiste Alayrac, Jeff Donahue, Pauline Luc, Antoine Miech, Iain Barr,
  Yana Hasson, Karel Lenc, Arthur Mensch, Katherine Millican, Malcolm Reynolds,
  Roman Ring, Eliza Rutherford, Serkan Cabi, Tengda Han, Zhitao Gong, Sina
  Samangooei, Marianne Monteiro, Jacob Menick, Sebastian Borgeaud, Andrew
  Brock, Aida Nematzadeh, Sahand Sharifzadeh, Mikolaj Binkowski, Ricardo
  Barreira, Oriol Vinyals, Andrew Zisserman, and Karen Simonyan.
\newblock {Flamingo: a Visual Language Model for Few-Shot Learning}.
\newblock In \emph{NeurIPS}, 2022.
\newblock URL \url{https://openreview.net/forum?id=EbMuimAbPbs}.

\bibitem[Alhaija et~al.(2018)Alhaija, Mustikovela, Mescheder, Geiger, and
  Rother]{Alhaija2018Kitti}
Hassan Alhaija, Siva Mustikovela, Lars Mescheder, Andreas Geiger, and Carsten
  Rother.
\newblock {Augmented Reality Meets Computer Vision: Efficient Data Generation
  for Urban Driving Scenes}.
\newblock \emph{International Journal of Computer Vision}, 2018.
\newblock URL \url{https://arxiv.org/abs/1708.01566}.

\bibitem[Assouel et~al.(2022)Assouel, Rodriguez, Taslakian, Vazquez, and
  Bengio]{assouel2022object}
Rim Assouel, Pau Rodriguez, Perouz Taslakian, David Vazquez, and Yoshua Bengio.
\newblock {Object-centric Compositional Imagination for Visual Abstract
  Reasoning}.
\newblock In \emph{{ICLR} Workshop on the Elements of Reasoning: Objects,
  Structure and Causality}, 2022.
\newblock URL \url{https://openreview.net/forum?id=rCzfIruU5x5}.

\bibitem[Assran et~al.(2022)Assran, Caron, Misra, Bojanowski, Bordes, Vincent,
  Joulin, Rabbat, and Ballas]{Assran2022MSN}
Mahmoud Assran, Mathilde Caron, Ishan Misra, Piotr Bojanowski, Florian Bordes,
  Pascal Vincent, Armand Joulin, Michael~G. Rabbat, and Nicolas Ballas.
\newblock {Masked Siamese Networks for Label-Efficient Learning}.
\newblock In \emph{ECCV}, 2022.
\newblock URL \url{https://arxiv.org/abs/2204.07141}.

\bibitem[Bao et~al.(2022)Bao, Tokmakov, Jabri, Wang, Gaidon, and
  Hebert]{Bao2022ObjectsThatMove}
Zhipeng Bao, Pavel Tokmakov, Allan Jabri, Yu-Xiong Wang, Adrien Gaidon, and
  Martial Hebert.
\newblock {Discovering Objects that Can Move}.
\newblock \emph{CVPR}, 2022.
\newblock URL \url{https://arxiv.org/abs/2203.10159}.

\bibitem[Biza et~al.(2022)Biza, Platt, van~de Meent, Wong, and
  Kipf]{biza2022binding}
Ondrej Biza, Robert Platt, Jan-Willem van~de Meent, Lawson~LS Wong, and Thomas
  Kipf.
\newblock {Binding Actions to Objects in World Models}.
\newblock In \emph{{ICLR} Workshop on the Elements of Reasoning: Objects,
  Structure and Causality}, 2022.
\newblock URL \url{https://openreview.net/forum?id=HImz8BuUclc}.

\bibitem[Burgess et~al.(2019)Burgess, Matthey, Watters, Kabra, Higgins,
  Botvinick, and Lerchner]{burgess2019monet}
Christopher~P. Burgess, Loic Matthey, Nicholas Watters, Rishabh Kabra, Irina
  Higgins, Matt Botvinick, and Alexander Lerchner.
\newblock {MONet: Unsupervised Scene Decomposition and Representation}.
\newblock \emph{arXiv:1901.11390}, 2019.
\newblock URL \url{https://arxiv.org/abs/1901.11390}.

\bibitem[Caesar et~al.(2018)Caesar, Uijlings, and Ferrari]{Caesar2018COCOStuff}
Holger Caesar, Jasper Uijlings, and Vittorio Ferrari.
\newblock {COCO-Stuff: Thing and Stuff Classes in Context}.
\newblock In \emph{CVPR}, 2018.
\newblock URL \url{https://arxiv.org/abs/1612.03716}.

\bibitem[Caron et~al.(2021)Caron, Touvron, Misra, Jégou, Mairal, Bojanowski,
  and Joulin]{Caron2021DINO}
Mathilde Caron, Hugo Touvron, Ishan Misra, Hervé Jégou, Julien Mairal, Piotr
  Bojanowski, and Armand Joulin.
\newblock {Emerging Properties in Self-Supervised Vision Transformers}.
\newblock \emph{ICCV}, 2021.
\newblock URL \url{https://arxiv.org/abs/2104.14294}.

\bibitem[Chen et~al.(2020{\natexlab{a}})Chen, Radford, Wu, Jun, Dhariwal, Luan,
  and Sutskever]{Chen2020ImageGPT}
Mark Chen, Alec Radford, Jeff Wu, Heewoo Jun, Prafulla Dhariwal, David Luan,
  and Ilya Sutskever.
\newblock {Generative Pretraining from Pixels}.
\newblock In \emph{ICML}, 2020{\natexlab{a}}.
\newblock URL \url{https://proceedings.mlr.press/v119/chen20s.html}.

\bibitem[Chen et~al.(2020{\natexlab{b}})Chen, Kornblith, Norouzi, and
  Hinton]{Chen2020SimCLR}
Ting Chen, Simon Kornblith, Mohammad Norouzi, and Geoffrey Hinton.
\newblock {A Simple Framework for Contrastive Learning of Visual
  Representations}.
\newblock In \emph{ICML}, 2020{\natexlab{b}}.
\newblock URL \url{https://proceedings.mlr.press/v119/chen20j.html}.

\bibitem[Chen et~al.(2021)Chen, Xie, and He]{Chen2021MoCov3}
Xinlei Chen, Saining Xie, and Kaiming He.
\newblock {An Empirical Study of Training Self-Supervised Vision Transformers}.
\newblock \emph{ICCV}, 2021.
\newblock URL \url{https://arxiv.org/abs/2104.02057}.

\bibitem[Cho et~al.(2021)Cho, Mall, Bala, and Hariharan]{Cho2021Picie}
Jang~Hyun Cho, Utkarsh Mall, Kavita Bala, and Bharath Hariharan.
\newblock {PiCIE: Unsupervised Semantic Segmentation Using Invariance and
  Equivariance in Clustering}.
\newblock In \emph{CVPR}, 2021.
\newblock URL \url{https://arxiv.org/abs/2103.17070}.

\bibitem[Cho et~al.(2014)Cho, van Merrienboer, G{\"u}lçehre, Bahdanau,
  Bougares, Schwenk, and Bengio]{Cho2014GRU}
Kyunghyun Cho, Bart van Merrienboer, Çaglar G{\"u}lçehre, Dzmitry Bahdanau,
  Fethi Bougares, Holger Schwenk, and Yoshua Bengio.
\newblock {Learning Phrase Representations using RNN Encoder–Decoder for
  Statistical Machine Translation}.
\newblock In \emph{EMNLP}, 2014.
\newblock URL \url{https://arxiv.org/abs/1406.1078}.

\bibitem[Cho et~al.(2015)Cho, Kwak, Schmid, and Ponce]{Cho2015UnsupervisedOD}
Minsu Cho, Suha Kwak, Cordelia Schmid, and Jean Ponce.
\newblock {Unsupervised Object Discovery and Localization in the Wild:
  Part-based Matching with Bottom-up Region Proposals}.
\newblock In \emph{CVPR}, 2015.
\newblock URL \url{https://arxiv.org/abs/1501.06170}.

\bibitem[Deng et~al.(2009)Deng, Dong, Socher, Li, Li, and
  Fei-Fei]{Deng2009ImageNet}
Jia Deng, Wei Dong, Richard Socher, Li-Jia Li, Kai Li, and Li~Fei-Fei.
\newblock {ImageNet: A Large-scale Hierarchical Image Database}.
\newblock In \emph{CVPR}, 2009.
\newblock \doi{10.1109/CVPR.2009.5206848}.
\newblock URL \url{https://ieeexplore.ieee.org/document/5206848}.

\bibitem[Dittadi et~al.(2022)Dittadi, Papa, De~Vita, Sch{\"o}lkopf, Winther,
  and Locatello]{dittadi2021generalization}
Andrea Dittadi, Samuele Papa, Michele De~Vita, Bernhard Sch{\"o}lkopf, Ole
  Winther, and Francesco Locatello.
\newblock {Generalization and Robustness Implications in Object-Centric
  Learning}.
\newblock In \emph{ICML}, 2022.
\newblock URL \url{https://proceedings.mlr.press/v162/dittadi22a.html}.

\bibitem[Dosovitskiy \& Brox(2016)Dosovitskiy and
  Brox]{dosovitskiy2016generating}
Alexey Dosovitskiy and Thomas Brox.
\newblock {Generating Images with Perceptual Similarity Metrics based on Deep
  Networks}.
\newblock In \emph{NeurIPS}, 2016.
\newblock URL
  \url{https://proceedings.neurips.cc/paper/2016/hash/371bce7dc83817b7893bcdeed13799b5-Abstract.html}.

\bibitem[Dosovitskiy et~al.(2021)Dosovitskiy, Beyer, Kolesnikov, Weissenborn,
  Zhai, Unterthiner, Dehghani, Minderer, Heigold, Gelly, Uszkoreit, and
  Houlsby]{Dosovitskiy2021ViT}
Alexey Dosovitskiy, Lucas Beyer, Alexander Kolesnikov, Dirk Weissenborn,
  Xiaohua Zhai, Thomas Unterthiner, Mostafa Dehghani, Matthias Minderer, Georg
  Heigold, Sylvain Gelly, Jakob Uszkoreit, and Neil Houlsby.
\newblock {An Image is Worth 16x16 Words: Transformers for Image Recognition at
  Scale}.
\newblock In \emph{ICLR}, 2021.
\newblock URL \url{https://openreview.net/forum?id=YicbFdNTTy}.

\bibitem[Elsayed et~al.(2022)Elsayed, Mahendran, van Steenkiste, Greff, Mozer,
  and Kipf]{elsayed2022savi++}
Gamaleldin~Fathy Elsayed, Aravindh Mahendran, Sjoerd van Steenkiste, Klaus
  Greff, Michael~Curtis Mozer, and Thomas Kipf.
\newblock {SAVi++: Towards End-to-End Object-Centric Learning from Real-World
  Videos}.
\newblock In \emph{NeurIPS}, 2022.
\newblock URL \url{https://openreview.net/forum?id=fT9W53lLxNS}.

\bibitem[Engelcke et~al.(2020)Engelcke, Kosiorek, Jones, and
  Posner]{Engelcke2020GENESIS}
Martin Engelcke, Adam~R. Kosiorek, Oiwi~Parker Jones, and Ingmar Posner.
\newblock {GENESIS: Generative Scene Inference and Sampling with Object-Centric
  Latent Representations}.
\newblock In \emph{ICLR}, 2020.
\newblock URL \url{https://openreview.net/forum?id=BkxfaTVFwH}.

\bibitem[Engelcke et~al.(2021)Engelcke, Jones, and
  Posner]{engelcke2021genesis2}
Martin Engelcke, Oiwi~Parker Jones, and Ingmar Posner.
\newblock {GENESIS-V2: Inferring Unordered Object Representations without
  Iterative Refinement}.
\newblock In \emph{NeurIPS}, 2021.
\newblock URL \url{https://openreview.net/forum?id=nRBZWEUhIhW}.

\bibitem[Eslami et~al.(2016)Eslami, Heess, Weber, Tassa, Szepesvari,
  Kavukcuoglu, and Hinton]{eslami2016attend}
S.~M.~Ali Eslami, Nicolas Heess, Theophane Weber, Yuval Tassa, David
  Szepesvari, Koray Kavukcuoglu, and Geoffrey~E. Hinton.
\newblock {Attend, Infer, Repeat: Fast Scene Understanding with Generative
  Models}.
\newblock In \emph{NeurIPS}, 2016.
\newblock URL
  \url{https://proceedings.neurips.cc/paper/2016/hash/52947e0ade57a09e4a1386d08f17b656-Abstract.html}.

\bibitem[Everingham et~al.(2012)Everingham, Van~Gool, Williams, Winn, and
  Zisserman]{Everingham2012PASCALVOC}
M.~Everingham, L.~Van~Gool, C.~K.~I. Williams, J.~Winn, and A.~Zisserman.
\newblock {The PASCAL Visual Object Classes Challenge 2012 (VOC2012)}, 2012.
\newblock URL
  \url{http://www.pascal-network.org/challenges/VOC/voc2012/workshop/index.html}.

\bibitem[Geiger et~al.(2013)Geiger, Lenz, Stiller, and
  Urtasun]{Geiger2013KITTI}
Andreas Geiger, Philip Lenz, Christoph Stiller, and Raquel Urtasun.
\newblock {Vision meets Robotics: The KITTI Dataset}.
\newblock \emph{International Journal of Robotics Research}, 2013.
\newblock URL \url{https://www.cvlibs.net/publications/Geiger2013IJRR.pdf}.

\bibitem[Goyal et~al.(2021)Goyal, Lamb, Hoffmann, Sodhani, Levine, Bengio, and
  Sch{\"o}lkopf]{Goyal2021RIMs}
Anirudh Goyal, Alex Lamb, Jordan Hoffmann, Shagun Sodhani, Sergey Levine,
  Yoshua Bengio, and Bernhard Sch{\"o}lkopf.
\newblock {Recurrent Independent Mechanisms}.
\newblock In \emph{ICLR}, 2021.
\newblock URL \url{https://openreview.net/forum?id=mLcmdlEUxy-}.

\bibitem[Greff et~al.(2019)Greff, Kaufman, Kabra, Watters, Burgess, Zoran,
  Matthey, Botvinick, and Lerchner]{greff2019multi}
Klaus Greff, Rapha{\"e}l~Lopez Kaufman, Rishabh Kabra, Nick Watters,
  Christopher Burgess, Daniel Zoran, Loic Matthey, Matthew Botvinick, and
  Alexander Lerchner.
\newblock {Multi-Object Representation Learning with Iterative Variational
  Inference}.
\newblock In \emph{ICML}, 2019.
\newblock URL \url{https://arxiv.org/abs/1903.00450}.

\bibitem[Greff et~al.(2020)Greff, Van~Steenkiste, and
  Schmidhuber]{greff2020binding}
Klaus Greff, Sjoerd Van~Steenkiste, and J{\"u}rgen Schmidhuber.
\newblock {On the Binding Problem in Artificial Neural Networks}.
\newblock \emph{arXiv:2012.05208}, 2020.
\newblock URL \url{https://arxiv.org/abs/2012.05208}.

\bibitem[Greff et~al.(2022)Greff, Belletti, Beyer, Doersch, Du, Duckworth,
  Fleet, Gnanapragasam, Golemo, Herrmann, Kipf, Kundu, Lagun, Laradji, Liu,
  Meyer, Miao, Nowrouzezahrai, Oztireli, Pot, Radwan, Rebain, Sabour, Sajjadi,
  Sela, Sitzmann, Stone, Sun, Vora, Wang, Wu, Yi, Zhong, and
  Tagliasacchi]{Greff2021Kubric}
Klaus Greff, Francois Belletti, Lucas Beyer, Carl Doersch, Yilun Du, Daniel
  Duckworth, David~J. Fleet, Dan Gnanapragasam, Florian Golemo, Charles
  Herrmann, Thomas Kipf, Abhijit Kundu, Dmitry Lagun, Issam Laradji,
  Hsueh-Ti~(Derek) Liu, Henning Meyer, Yishu Miao, Derek Nowrouzezahrai, Cengiz
  Oztireli, Etienne Pot, Noha Radwan, Daniel Rebain, Sara Sabour, Mehdi S.~M.
  Sajjadi, Matan Sela, Vincent Sitzmann, Austin Stone, Deqing Sun, Suhani Vora,
  Ziyu Wang, Tianhao Wu, Kwang~Moo Yi, Fangcheng Zhong, and Andrea
  Tagliasacchi.
\newblock {Kubric: A Scalable Dataset Generator}.
\newblock In \emph{CVPR}, 2022.
\newblock URL \url{https://arxiv.org/abs/2203.03570}.

\bibitem[Grill et~al.(2020)Grill, Strub, Altché, Tallec, Richemond,
  Buchatskaya, Doersch, Avila~Pires, Guo, Gheshlaghi~Azar, Piot, Kavukcuoglu,
  Munos, and Valko]{Grill2020BYOL}
Jean-Bastien Grill, Florian Strub, Florent Altché, Corentin Tallec, Pierre
  Richemond, Elena Buchatskaya, Carl Doersch, Bernardo Avila~Pires, Zhaohan
  Guo, Mohammad Gheshlaghi~Azar, Bilal Piot, Koray Kavukcuoglu, Remi Munos, and
  Michal Valko.
\newblock {Bootstrap Your Own Latent - A New Approach to Self-Supervised
  Learning}.
\newblock In \emph{NeurIPS}, 2020.
\newblock URL \url{https://arxiv.org/abs/2006.07733}.

\bibitem[Hamilton et~al.(2022)Hamilton, Zhang, Hariharan, Snavely, and
  Freeman]{Hamilton2022Stego}
Mark Hamilton, Zhoutong Zhang, Bharath Hariharan, Noah Snavely, and William~T.
  Freeman.
\newblock {Unsupervised Semantic Segmentation by Distilling Feature
  Correspondences}.
\newblock In \emph{ICLR}, 2022.
\newblock URL \url{https://openreview.net/forum?id=SaKO6z6Hl0c}.

\bibitem[Hariharan et~al.(2011)Hariharan, Arbelaez, Bourdev, Maji, and
  Malik]{Bharath2011SBD}
Bharath Hariharan, Pablo Arbelaez, Lubomir Bourdev, Subhransu Maji, and
  Jitendra Malik.
\newblock {Semantic Contours from Inverse Detectors}.
\newblock In \emph{ICCV}, 2011.
\newblock URL \url{https://ieeexplore.ieee.org/document/6126343}.

\bibitem[He et~al.(2015)He, Zhang, Ren, and Sun]{He2015ResNet}
Kaiming He, Xiangyu Zhang, Shaoqing Ren, and Jian Sun.
\newblock {Deep Residual Learning for Image Recognition}.
\newblock In \emph{CVPR}, 2015.
\newblock URL \url{https://arxiv.org/abs/1512.03385}.

\bibitem[He et~al.(2022)He, Chen, Xie, Li, Doll\'ar, and Girshick]{He2022MAE}
Kaiming He, Xinlei Chen, Saining Xie, Yanghao Li, Piotr Doll\'ar, and Ross
  Girshick.
\newblock {Masked Autoencoders are Scalable Vision Learners}.
\newblock In \emph{CVPR}, 2022.
\newblock URL \url{https://arxiv.org/abs/2111.06377}.

\bibitem[Hinton et~al.(2015)Hinton, Vinyals, and Dean]{Hinton2015DistillingTK}
Geoffrey~E. Hinton, Oriol Vinyals, and Jeffrey Dean.
\newblock {Distilling the Knowledge in a Neural Network}.
\newblock In \emph{NeurIPS 2014 Deep Learning Workshop}, 2015.
\newblock URL \url{https://arxiv.org/abs/1503.02531}.

\bibitem[Huang et~al.(2022)Huang, Chen, and Rudin]{Huang2022SegDiscover}
Haiyang Huang, Zhi Chen, and Cynthia Rudin.
\newblock {SegDiscover: Visual Concept Discovery via Unsupervised Semantic
  Segmentation}.
\newblock \emph{arXiv:2204.10926}, 2022.
\newblock URL \url{https://arxiv.org/abs/2204.10926}.

\bibitem[Hénaff et~al.(2022)Hénaff, Koppula, Shelhamer, Zoran, Jaegle,
  Zisserman, Carreira, and Arandjelovi'c]{Henaff2022Odin}
Olivier~J. Hénaff, Skanda Koppula, Evan Shelhamer, Daniel Zoran, Andrew
  Jaegle, Andrew Zisserman, Jo{\~a}o Carreira, and Relja Arandjelovi'c.
\newblock {Object Discovery and Representation Networks}.
\newblock In \emph{ECCV}, 2022.
\newblock URL \url{https://arxiv.org/abs/2203.08777}.

\bibitem[Jaegle et~al.(2022)Jaegle, Borgeaud, Alayrac, Doersch, Ionescu, Ding,
  Koppula, Zoran, Brock, Shelhamer, Henaff, Botvinick, Zisserman, Vinyals, and
  Carreira]{Jaegle2022PerceiverIO}
Andrew Jaegle, Sebastian Borgeaud, Jean-Baptiste Alayrac, Carl Doersch, Catalin
  Ionescu, David Ding, Skanda Koppula, Daniel Zoran, Andrew Brock, Evan
  Shelhamer, Olivier~J. Henaff, Matthew Botvinick, Andrew Zisserman, Oriol
  Vinyals, and Joao Carreira.
\newblock {Perceiver IO: A General Architecture for Structured Inputs \&
  Outputs}.
\newblock In \emph{ICLR}, 2022.
\newblock URL \url{https://openreview.net/forum?id=fILj7WpI-g}.

\bibitem[Ji et~al.(2019)Ji, Henriques, and Vedaldi]{Ji2019IIC}
Xu~Ji, Jo{\~a}o~F. Henriques, and Andrea Vedaldi.
\newblock {Invariant Information Clustering for Unsupervised Image
  Classification and Segmentation}.
\newblock In \emph{ICCV}, 2019.
\newblock URL \url{https://arxiv.org/abs/1807.06653}.

\bibitem[Jiang et~al.(2020)Jiang, Janghorbani, {de Melo}, and
  Ahn]{Jiang2020SCALOR}
Jindong Jiang, Sepehr Janghorbani, Gerard {de Melo}, and Sungjin Ahn.
\newblock {SCALOR: Generative World Models with Scalable Object
  Representations}.
\newblock In \emph{ICLR}, 2020.
\newblock URL \url{https://openreview.net/pdf?id=SJxrKgStDH}.

\bibitem[Kahneman et~al.(1992)Kahneman, Treisman, and
  Gibbs]{kahneman1992reviewing}
Daniel Kahneman, Anne Treisman, and Brian~J. Gibbs.
\newblock {The Reviewing of Object Files: Object-specific Integration of
  Information}.
\newblock \emph{Cognitive psychology}, 1992.
\newblock URL
  \url{https://www.sciencedirect.com/science/article/abs/pii/001002859290007O}.

\bibitem[Karazija et~al.(2021)Karazija, Laina, and
  Rupprecht]{Karazija2021clevrtex}
Laurynas Karazija, Iro Laina, and Christian Rupprecht.
\newblock {ClevrTex: A Texture-Rich Benchmark for Unsupervised Multi-Object
  Segmentation}.
\newblock In \emph{NeurIPS Track on Datasets and Benchmarks}, 2021.
\newblock URL \url{https://arxiv.org/abs/2111.10265}.

\bibitem[Kingma \& Ba(2015)Kingma and Ba]{kingma2014adam}
Diederik~P. Kingma and Jimmy Ba.
\newblock {Adam: A Method for Stochastic Optimization}.
\newblock In \emph{ICLR}, 2015.
\newblock URL \url{https://arxiv.org/abs/1412.6980}.

\bibitem[Kipf et~al.(2022)Kipf, Elsayed, Mahendran, Stone, Sabour, Heigold,
  Jonschkowski, Dosovitskiy, and Greff]{kipf2022conditional}
Thomas Kipf, Gamaleldin~Fathy Elsayed, Aravindh Mahendran, Austin Stone, Sara
  Sabour, Georg Heigold, Rico Jonschkowski, Alexey Dosovitskiy, and Klaus
  Greff.
\newblock {Conditional Object-centric Learning from Video}.
\newblock In \emph{ICLR}, 2022.
\newblock URL \url{https://openreview.net/forum?id=aD7uesX1GF_}.

\bibitem[Kosiorek et~al.(2018)Kosiorek, Kim, Teh, and
  Posner]{Kosiorek2018SQAIR}
Adam Kosiorek, Hyunjik Kim, Yee~Whye Teh, and Ingmar Posner.
\newblock {Sequential Attend, Infer, Repeat: Generative Modelling of Moving
  Objects}.
\newblock In \emph{NeurIPS}, 2018.
\newblock URL
  \url{https://proceedings.neurips.cc/paper/2018/hash/7417744a2bac776fabe5a09b21c707a2-Abstract.html}.

\bibitem[Kuhn(1955)]{Kuhn95Hungarian}
Harold~W. Kuhn.
\newblock {The Hungarian Method for the Assignment Problem}.
\newblock \emph{Naval Research Logistics Quarterly}, 1955.
\newblock URL
  \url{https://onlinelibrary.wiley.com/doi/abs/10.1002/nav.3800020109}.

\bibitem[Lin et~al.(2014)Lin, Maire, Belongie, Hays, Perona, Ramanan,
  Doll{\'a}r, and Zitnick]{Lin2014COCO}
Tsung-Yi Lin, Michael Maire, Serge~J. Belongie, James Hays, Pietro Perona, Deva
  Ramanan, Piotr Doll{\'a}r, and C.~Lawrence Zitnick.
\newblock {Microsoft COCO: Common Objects in Context}.
\newblock In \emph{ECCV}, 2014.
\newblock URL \url{https://arxiv.org/abs/1405.0312}.

\bibitem[Lin et~al.(2020)Lin, Fu, Peri, Sun, Singh, Deng, Jiang, and
  Ahn]{Lin2020SPACE}
Zhixuan Lin, Yi-Wu Fu, Skand~Vishwanath Peri, Weihao Sun, Gautam Singh, Fei
  Deng, Jindong Jiang, and Sungjin Ahn.
\newblock {SPACE: Unsupervised Object-Oriented Scene Representation via Spatial
  Attention and Decomposition}.
\newblock In \emph{ICLR}, 2020.
\newblock URL \url{https://openreview.net/forum?id=rkl03ySYDH}.

\bibitem[Locatello et~al.(2020)Locatello, Weissenborn, Unterthiner, Mahendran,
  Heigold, Uszkoreit, Dosovitskiy, and Kipf]{Locatello2020SlotAttention}
Francesco Locatello, Dirk Weissenborn, Thomas Unterthiner, Aravindh Mahendran,
  Georg Heigold, Jakob Uszkoreit, Alexey Dosovitskiy, and Thomas Kipf.
\newblock {Object-Centric Learning with Slot Attention}.
\newblock In \emph{NeurIPS}, 2020.
\newblock URL
  \url{https://proceedings.neurips.cc/paper/2020/file/8511df98c02ab60aea1b2356c013bc0f-Paper.pdf}.

\bibitem[Mambelli et~al.(2022)Mambelli, Tr{\"a}uble, Bauer, Sch{\"o}lkopf, and
  Locatello]{mambelli2022compositional}
Davide Mambelli, Frederik Tr{\"a}uble, Stefan Bauer, Bernhard Sch{\"o}lkopf,
  and Francesco Locatello.
\newblock {Compositional Multi-object Reinforcement Learning with Linear
  Relation Networks}.
\newblock In \emph{ICLR Workshop on the Elements of Reasoning: Objects,
  Structure and Causality}, 2022.
\newblock URL \url{https://openreview.net/forum?id=HFUxPr_I5ec}.

\bibitem[Melas{-}Kyriazi et~al.(2022)Melas{-}Kyriazi, Rupprecht, Laina, and
  Vedaldi]{Melaskyriazi2022DeepSpectral}
Luke Melas{-}Kyriazi, Christian Rupprecht, Iro Laina, and Andrea Vedaldi.
\newblock {Deep Spectral Methods: A Surprisingly Strong Baseline for
  Unsupervised Semantic Segmentation and Localization}.
\newblock In \emph{CVPR}, 2022.
\newblock URL \url{https://arxiv.org/abs/2205.07839}.

\bibitem[Mottaghi et~al.(2014)Mottaghi, Chen, Liu, Cho, Lee, Fidler, Urtasun,
  and Yuille]{Mottaghi14PASCALcontext}
Roozbeh Mottaghi, Xianjie Chen, Xiaobai Liu, Nam-Gyu Cho, Seong-Whan Lee, Sanja
  Fidler, Raquel Urtasun, and Alan Yuille.
\newblock {The Role of Context for Object Detection and Semantic Segmentation
  in the Wild}.
\newblock In \emph{CVPR}, 2014.
\newblock URL \url{https://ieeexplore.ieee.org/document/6909514}.

\bibitem[Ouali et~al.(2020)Ouali, Hudelot, and
  Tami]{oualiAutoregressiveUnsupervisedImage2020a}
Yassine Ouali, Céline Hudelot, and Myriam Tami.
\newblock {Autoregressive Unsupervised Image Segmentation}.
\newblock In \emph{ECCV}, 2020.
\newblock URL \url{https://arxiv.org/abs/2007.08247}.

\bibitem[Pont-Tuset et~al.(2017)Pont-Tuset, Arbeláez, Barron, Marques, and
  Malik]{Arbelaez2014MCG}
Jordi Pont-Tuset, Pablo Arbeláez, Jonathan~T. Barron, Ferran Marques, and
  Jitendra Malik.
\newblock {Multiscale Combinatorial Grouping for Image Segmentation and Object
  Proposal Generation}.
\newblock \emph{{IEEE} Transactions on Pattern Analysis and Machine
  Intelligence}, 39\penalty0 (1), 2017.
\newblock \doi{10.1109/TPAMI.2016.2537320}.
\newblock URL \url{https://ieeexplore.ieee.org/document/7423791}.

\bibitem[Reed et~al.(2022)Reed, Zolna, Parisotto, Colmenarejo, Novikov, maron
  Gabriel, Gim{\'e}nez, Sulsky, Kay, Springenberg, Eccles, Bruce, Razavi,
  Edwards, Heess, Chen, Hadsell, Vinyals, Bordbar, and
  de~Freitas]{Reed2022Gato}
Scott Reed, Konrad Zolna, Emilio Parisotto, Sergio~G{\'o}mez Colmenarejo,
  Alexander Novikov, Barth maron Gabriel, Mai Gim{\'e}nez, Yury Sulsky, Jackie
  Kay, Jost~Tobias Springenberg, Tom Eccles, Jake Bruce, Ali Razavi, Ashley
  Edwards, Nicolas Heess, Yutian Chen, Raia Hadsell, Oriol Vinyals, Mahyar
  Bordbar, and Nando de~Freitas.
\newblock {A Generalist Agent}.
\newblock \emph{TMLR}, 2022.
\newblock URL \url{https://openreview.net/forum?id=1ikK0kHjvj}.

\bibitem[Sch{\"o}lkopf et~al.(2021)Sch{\"o}lkopf, Locatello, Bauer, Ke,
  Kalchbrenner, Goyal, and Bengio]{scholkopf2021toward}
Bernhard Sch{\"o}lkopf, Francesco Locatello, Stefan Bauer, Nan~Rosemary Ke, Nal
  Kalchbrenner, Anirudh Goyal, and Yoshua Bengio.
\newblock {Towards Causal Representation Learning}.
\newblock \emph{{IEEE} - Advances in Machine Learning and Deep Neural
  Networks}, 2021.
\newblock URL \url{https://arxiv.org/abs/2102.11107}.

\bibitem[Simeoni et~al.(2021)Simeoni, Puy, Vo, Roburin, Gidaris, Bursuc,
  Pérez, Marlet, and Ponce]{Simeoni2021LOST}
Oriane Simeoni, Gilles Puy, Huy~V. Vo, Simon Roburin, Spyros Gidaris, Andrei
  Bursuc, Patrick Pérez, Renaud Marlet, and Jean Ponce.
\newblock {Localizing Objects with Self-Supervised Transformers and no Labels}.
\newblock In \emph{BMVC}, 2021.
\newblock URL \url{https://arxiv.org/abs/2109.14279}.

\bibitem[Singh et~al.(2022{\natexlab{a}})Singh, Deng, and Ahn]{Singh2022SLATE}
Gautam Singh, Fei Deng, and Sungjin Ahn.
\newblock {Illiterate DALL-E Learns to Compose}.
\newblock In \emph{ICLR}, 2022{\natexlab{a}}.
\newblock URL \url{https://openreview.net/forum?id=h0OYV0We3oh}.

\bibitem[Singh et~al.(2022{\natexlab{b}})Singh, Wu, and Ahn]{Singh2022STEVE}
Gautam Singh, Yi-Fu Wu, and Sungjin Ahn.
\newblock {Simple Unsupervised Object-Centric Learning for Complex and
  Naturalistic Videos}.
\newblock In \emph{NeurIPS}, 2022{\natexlab{b}}.
\newblock URL \url{https://openreview.net/forum?id=eYfIM88MTUE}.

\bibitem[Spelke \& Kinzler(2007)Spelke and Kinzler]{spelke2007core}
Elizabeth~S. Spelke and Katherine~D. Kinzler.
\newblock {Core Knowledge}.
\newblock \emph{Developmental Science}, 2007.
\newblock URL
  \url{https://onlinelibrary.wiley.com/doi/abs/10.1111/j.1467-7687.2007.00569.x}.

\bibitem[Traub et~al.(2023)Traub, Otte, Menge, Karlbauer, Thuemmel, and
  Butz]{traub2022learning}
Manuel Traub, Sebastian Otte, Tobias Menge, Matthias Karlbauer, Jannik
  Thuemmel, and Martin~V. Butz.
\newblock {Learning What and Where: Disentangling Location and Identity
  Tracking Without Supervision}.
\newblock In \emph{ICLR}, 2023.
\newblock URL \url{https://openreview.net/forum?id=NeDc-Ak-H_}.

\bibitem[van~den Oord et~al.(2017)van~den Oord, Vinyals, and
  Kavukcuoglu]{Oord2017VQVAE}
A{\"a}ron van~den Oord, Oriol Vinyals, and Koray Kavukcuoglu.
\newblock {Neural Discrete Representation Learning}.
\newblock In \emph{NeurIPS}, 2017.
\newblock URL
  \url{https://proceedings.neurips.cc/paper/2017/hash/7a98af17e63a0ac09ce2e96d03992fbc-Abstract.html}.

\bibitem[Van~Gansbeke et~al.(2021)Van~Gansbeke, Vandenhende, Georgoulis, and
  Van~Gool]{VanGansbeke2021MaskContrast}
Wouter Van~Gansbeke, Simon Vandenhende, Stamatios Georgoulis, and Luc Van~Gool.
\newblock {Unsupervised Semantic Segmentation by Contrasting Object Mask
  Proposals}.
\newblock \emph{ICCV}, 2021.
\newblock URL \url{https://arxiv.org/abs/2102.06191}.

\bibitem[Van~Gansbeke et~al.(2022)Van~Gansbeke, Vandenhende, and
  Van~Gool]{VanGansbeke2022MaskDistill}
Wouter Van~Gansbeke, Simon Vandenhende, and Luc Van~Gool.
\newblock {Discovering Object Masks with Transformers for Unsupervised Semantic
  Segmentation}.
\newblock \emph{arXiv:2206.06363}, 2022.
\newblock URL \url{https://arxiv.org/abs/2206.06363}.

\bibitem[Vaswani et~al.(2017)Vaswani, Shazeer, Parmar, Uszkoreit, Jones, Gomez,
  Kaiser, and Polosukhin]{Vaswani2017Attention}
Ashish Vaswani, Noam Shazeer, Niki Parmar, Jakob Uszkoreit, Llion Jones,
  Aidan~N. Gomez, Lukasz Kaiser, and Illia Polosukhin.
\newblock {Attention is All you Need}.
\newblock In \emph{NeurIPS}, 2017.
\newblock URL
  \url{https://papers.nips.cc/paper/2017/hash/3f5ee243547dee91fbd053c1c4a845aa-Abstract.html}.

\bibitem[Vo et~al.(2020)Vo, Pérez, and Ponce]{Vo2020OSD}
Huy~V. Vo, Patrick Pérez, and Jean Ponce.
\newblock {Toward Unsupervised, Multi-object Discovery in Large-scale Image
  Collections}.
\newblock In \emph{ECCV}, 2020.
\newblock URL \url{https://arxiv.org/abs/2007.02662}.

\bibitem[Vo et~al.(2021)Vo, Sizikova, Schmid, Pérez, and Ponce]{Vo2021LOD}
Huy~V. Vo, Elena Sizikova, Cordelia Schmid, Patrick Pérez, and Jean Ponce.
\newblock {Large-scale Unsupervised Object Discovery}.
\newblock In \emph{NeurIPS}, 2021.
\newblock URL \url{https://arxiv.org/abs/2106.06650}.

\bibitem[Wang et~al.(2022)Wang, Shen, Hu, Yuan, Crowley, and
  Vaufreydaz]{Wang2022TokenCut}
Yangtao Wang, Xi~Shen, Shell~Xu Hu, Yuan Yuan, James~L. Crowley, and Dominique
  Vaufreydaz.
\newblock {Self-Supervised Transformers for Unsupervised Object Discovery Using
  Normalized Cut}.
\newblock In \emph{CVPR}, 2022.
\newblock URL \url{https://arxiv.org/abs/2202.11539}.

\bibitem[Watters et~al.(2019)Watters, Matthey, Burgess, and
  Lerchner]{Watters2019SpatialBD}
Nick Watters, Loic Matthey, Chris~P. Burgess, and Alexander Lerchner.
\newblock {Spatial Broadcast Decoder: A Simple Architecture for Disentangled
  Representations in VAEs}.
\newblock In \emph{ICLR Learning from Limited Labeled Data Workshop}, 2019.
\newblock URL \url{https://openreview.net/forum?id=S1x7WjnzdV}.

\bibitem[Weis et~al.(2021)Weis, Chitta, Sharma, Brendel, Bethge, Geiger, and
  Ecker]{weis2021benchmarking}
Marissa~A Weis, Kashyap Chitta, Yash Sharma, Wieland Brendel, Matthias Bethge,
  Andreas Geiger, and Alexander~S. Ecker.
\newblock {Benchmarking Unsupervised Object Representations for Video
  Sequences}.
\newblock \emph{JMLR}, 2021.
\newblock URL \url{https://jmlr.org/papers/v22/21-0199.html}.

\bibitem[Wen et~al.(2022)Wen, Zhao, Zheng, Zhang, and Qi]{Wen2022SlotCon}
Xin Wen, Bingchen Zhao, Anlin Zheng, Xiangyu Zhang, and Xiaojuan Qi.
\newblock {Self-Supervised Visual Representation Learning with Semantic
  Grouping}.
\newblock In \emph{NeurIPS}, 2022.
\newblock URL \url{https://openreview.net/forum?id=H3JObxjd8S}.

\bibitem[Wightman(2019)]{Wightman2019timm}
Ross Wightman.
\newblock {PyTorch Image Models}, 2019.
\newblock URL \url{https://github.com/rwightman/pytorch-image-models}.

\bibitem[Xie et~al.(2021)Xie, Zhan, Liu, Ong, and Change]{xie2021unsupervised}
Jiahao Xie, Xiaohang Zhan, Ziwei Liu, Yew-Soon Ong, and Chen~Loy Change.
\newblock {Unsupervised Object-Level Representation Learning from Scene
  Images}.
\newblock In \emph{NeurIPS}, 2021.
\newblock URL \url{https://openreview.net/forum?id=X2K8KVEaAXG}.

\bibitem[Xiong et~al.(2020)Xiong, Yang, He, Zheng, Zheng, Xing, Zhang, Lan,
  Wang, and Liu]{Xiong2020OnLN}
Ruibin Xiong, Yunchang Yang, Di~He, Kai Zheng, Shuxin Zheng, Chen Xing,
  Huishuai Zhang, Yanyan Lan, Liwei Wang, and Tie-Yan Liu.
\newblock {On Layer Normalization in the Transformer Architecture}.
\newblock In \emph{ICML}, 2020.
\newblock URL \url{https://proceedings.mlr.press/v119/xiong20b.html}.

\bibitem[Xu et~al.(2022)Xu, De~Mello, Liu, Byeon, Breuel, Kautz, and
  Wang]{Xu2022GroupViT}
Jiarui Xu, Shalini De~Mello, Sifei Liu, Wonmin Byeon, Thomas Breuel, Jan Kautz,
  and Xiaolong Wang.
\newblock {GroupViT: Semantic Segmentation Emerges from Text Supervision}.
\newblock In \emph{CVPR}, 2022.
\newblock URL \url{https://arxiv.org/abs/2202.11094}.

\bibitem[Yang et~al.(2020)Yang, Mao, Wu, Parikh, Cox, Tenenbaum, and
  Gan]{yang2020object}
Jianwei Yang, Jiayuan Mao, Jiajun Wu, Devi Parikh, David~D Cox, Joshua~B.
  Tenenbaum, and Chuang Gan.
\newblock {Object-centric Diagnosis of Visual Reasoning}.
\newblock \emph{arXiv:2012.11587}, 2020.
\newblock URL \url{https://arxiv.org/abs/2012.11587}.

\bibitem[Yang \& Yang(2022)Yang and Yang]{yang2022promising}
Yafei Yang and Bo~Yang.
\newblock {Promising or Elusive? Unsupervised Object Segmentation from
  Real-world Single Images}.
\newblock In \emph{NeurIPS}, 2022.
\newblock URL \url{https://openreview.net/forum?id=DzPWTwfby5d}.

\bibitem[Zadaianchuk et~al.(2020)Zadaianchuk, Seitzer, and
  Martius]{zadaianchuk2020self}
Andrii Zadaianchuk, Maximilian Seitzer, and Georg Martius.
\newblock {Self-supervised Visual Reinforcement Learning with Object-centric
  Representations}.
\newblock In \emph{ICLR}, 2020.
\newblock URL \url{https://openreview.net/forum?id=xppLmXCbOw1}.

\bibitem[Zadaianchuk et~al.(2023)Zadaianchuk, Kleindessner, Zhu, Locatello, and
  Brox]{Zadaianchuk2022COMUS}
Andrii Zadaianchuk, Matthaeus Kleindessner, Yi~Zhu, Francesco Locatello, and
  Thomas Brox.
\newblock {Unsupervised Semantic Segmentation with Self-supervised
  Object-centric Representations}.
\newblock In \emph{ICLR}, 2023.
\newblock URL \url{https://openreview.net/forum?id=1_jFneF07YC}.

\end{thebibliography}
\bibliographystyle{iclr2023_conference}

\newpage
\appendix

\addcontentsline{toc}{section}{Appendix} %
\part{Appendix} %
\parttoc %
\FloatBarrier

\newpage

\section{Extended Related Work}
\label{app:expanded-related-work}

\looseness-1
\textbf{Unsupervised Semantic Segmentation} is the challenging task of structuring natural scene into semantically coherent regions without any human annotations.
Methods for this task~\citep{Ji2019IIC,oualiAutoregressiveUnsupervisedImage2020a,Cho2021Picie,VanGansbeke2021MaskContrast,VanGansbeke2022MaskDistill,Hamilton2022Stego, Zadaianchuk2022COMUS} are typically separated into two independent steps: first, representations for each pixel by a pre-trained self-supervised model are refined; second, these representations are clustered on the whole dataset to assign a class category to each pixel.
\dinosaur combines those steps into a simple method that groups semantically meaningful regions jointly with learning region representations.

Approaches that try to learn and cluster dense representations without additional assumptions~\citep{Ji2019IIC, oualiAutoregressiveUnsupervisedImage2020a} were shown to be effective only on small-scale datasets with narrow visual domains.
More advanced methods incorporate geometric consistency~\citep{Cho2021Picie} or unsupervised saliency detection~\citep{VanGansbeke2022MaskDistill, Zadaianchuk2022COMUS} to simplify dense representation learning and the grouping task.
Additionally, using self-supervised DINO representations~\citep{Caron2021DINO} by contrasting them~\citep{Hamilton2022Stego} or to self-train a semantic segmentation model~\citep{VanGansbeke2022MaskDistill, Melaskyriazi2022DeepSpectral,Zadaianchuk2022COMUS} allows obtaining a more accurate image decomposition to semantic categories.

\vspace{0.5em}
\textbf{Object Localization} is another research direction that aims at structuring the scene from images, but by predicting object-level features such as bounding boxes or category.
Recent works~\citep{Vo2020OSD, Vo2021LOD, Simeoni2021LOST, Melaskyriazi2022DeepSpectral, Wang2022TokenCut} often group pre-trained, self-supervised features of individual images into initial object proposals, and then aggregate and cluster these proposals across images to assign them labels.
Self-training can then be used on top to achieve multi-object detection on natural scenes.

At first, rOSD~\citep{Vo2020OSD, Vo2021LOD} used features obtained from a supervised classifier to localize salient objects in the image.
Next, LOST~\citep{Simeoni2021LOST} showed that dense, self-supervised DINO features can also be used for localization of the most prominent object in the image.
Finally, DeepSpectral~\citep{Melaskyriazi2022DeepSpectral} and TokenCut~\citep{Wang2022TokenCut} showed that additionally transforming DINO features to a spectral embedding can significantly improve the quality of the object proposals.
While those methods can be combined with self-training of object detectors like R-CNN, they are limited by the quality of the set of salient objects used as pseudo-labels.
In contrast, \dinosaur splits each scene to a set of object proposals enabling multi-object scene decomposition without an additional recognition network.

\vspace{0.5em}
\textbf{Architectures making use of slot-like components} have also enjoyed popularity aside from the research explicitly focused on object-centric learning.
For instance, the Perceiver model learns to distill a large set of input tokens from different modalities into a reduced set of embeddings, which can be seen as slots~\citep{Jaegle2022PerceiverIO}.
Prominent recent successes of large-scale multi-modal modeling such as Flamingo~\citep{Alayrac2022Flamingo} or Gato~\citep{Reed2022Gato} are based on the Perceiver architecture.
In dynamics modeling, recurrent independent mechanisms~\citep{Goyal2021RIMs} use a competition between independent modules to read inputs into a slot-structured latent state.

Recent work on contrastive learning such as SlotCon~\citep{Wen2022SlotCon} or Odin~\citep{Henaff2022Odin} also can be seen as learning slot representations, although with a focus on semantic (class-based) slots rather than instances.
In the case of SlotCon, this is due to the reliance on prototypes which naturally lead to semantically focused representations whereas in the case of Odin, the contrastive objective does not encourage differentiation of one or multiple instances.
Further, Odin is positioned as a strategy to pre-train a backbone which can then be finetuned on downstream tasks and is not intended to be an unsupervised method that directly leads to instance-level grouping.
Note that even though Odin also evaluates object discovery on COCO using the mBO metric, the obtained numbers are not comparable to ours as Odin generates 255 proposal masks by running K-Means 128 times with different numbers of clusters.
Finally, ORL~\citep{xie2021unsupervised} addresses the reliance of many contrastive methods on ImageNet-specific biases, which leads to intra-image contrasting of the same object under different augmentations.
This is done by identifying correspondences of objects in different images using models trained via intra-image contrasting and subsequently training the model using inter-image contrasting by selecting objects of the same type from different images.

\section{Expanded Comparison with Object-Centric Methods}
\label{app:comparison-object-centric-methods}

\looseness-1
In this section, we present additional results on object discovery.
In particular, we compare to two state-of-the-art video-based object-centric methods: STEVE~\citep{Singh2022STEVE} and SAVi++~\citep{elsayed2022savi++} (\sec{app:subsec:comparison-video-methods}).
We also present preliminary results on the KITTI driving dataset~\citep{Alhaija2018Kitti} (\sec{app:results-kitti}), and a study of the learned slot representations on COCO (\sec{app:subsec:coco-property-prediction}).

\begin{table*}[tb]
    \centering
    \caption{Object discovery on synthetic datasets (mean $\pm$ standard dev., 5 seeds), corresponding to results from \fig{fig:results-object-discovery-synthetic-quant}. We additionally add results (over 3 seeds) for the video methods STEVE~\citep{Singh2022STEVE} and SAVi++~\citep{elsayed2022savi++}.
    Note that this version of SAVi++ is unconditional, \ie it does not have access to the objects of the first frame.
    We train SAVi++ with (unsupervised) image reconstruction, and (supervised) optical flow reconstruction.
    To ensure a valid comparison, STEVE and SAVi++ are trained on video, but evaluated by processing each video frame independently.
    }
    \small
    \label{tab:results-object-discovery-synthetic}
    \begin{tabular}{@{}lcccccc@{}}
    \toprule
    & \multicolumn{2}{c}{\textbf{MOVi-C}} & {} & \multicolumn{2}{c}{\textbf{MOVi-E}} \\
    \cmidrule{2-3} \cmidrule{5-6}
    & FG-ARI & mBO & \phantom & FG-ARI & mBO \\
    \midrule
    \emph{Block Masks} & \emph{43.8} & \emph{27.5} && \emph{45.4} & \emph{24.3} \\
    \emph{DINO K-Means} & \emph{48.9} & \emph{38.5} && \emph{50.2} & \emph{23.6} \\
    Slot Attention & \meanandstd{43.76}{0.30}
 & \meanandstd{26.22}{0.96} && \meanandstd{44.98}{1.67} & \meanandstd{23.98}{1.20} \\
    SLATE & \meanandstd{43.56}{1.26} & \meanandstd{26.53}{1.11} && \meanandstd{44.36}{0.86} & \meanandstd{23.60}{0.54} \\
    \dinosaur (ViT-S/8) & \meanandstd{67.21}{0.33} & \meanandstd{38.63}{0.14} && \meanandstd{64.66}{0.72} & \meanandstd{34.10}{0.12} \\
    \dinosaur (ViT-B/8) & \meanandstd{68.63}{0.44} & \meanandstd{39.07}{0.19} && \meanandstd{65.13}{1.23} & \meanandstd{35.52}{0.17} \\
    \midrule
    STEVE & \meanandstd{44.25}{0.75} & && \meanandstd{44.52}{2.81} & \\
    SAVi++ (Image Recon.) & \meanandstd{10.39}{0.07} & && \meanandstd{15.93}{1.13} & \\ %
    SAVi++ (Flow Recon.) & \meanandstd{29.98}{1.48} & && \meanandstd{19.79}{1.99} & \\  %
    \bottomrule
    \end{tabular}
\end{table*}

\subsection{Comparison to Video Methods}
\label{app:subsec:comparison-video-methods}
Having access to motion information is often thought to be advantageous for discovering objects~\citep{greff2020binding,kipf2022conditional}, as object identity and motions are stable through time.
It is thus interesting how our method compares against models that have been trained on videos.
To this end, we test two recent video-based object-centric: STEVE~\citep{Singh2022STEVE} and SAVi++~\citep{elsayed2022savi++}.
STEVE is an extension of SLATE~\citep{Singh2022SLATE} to videos, whereas SAVi++ is a scaled-up version of SAVi~\citep{kipf2022conditional}, which in turn is the video extension of Slot Attention~\citep{Locatello2020SlotAttention}.
We train both of them on the synthetic MOVi-C (with 11 slots) and MOVi-E (with 24 slots), using the original hyperparameters proposed for those datasets.
However, we train SAVi++ unconditionally, that is, using random learnt slot initializations instead of providing first-frame supervision of the object positions.
We also had to reduce the batch size of SAVi++ from 64 to 32 due to computational reasons.
STEVE is trained for 200k steps, and SAVi++ for 300k steps.
To ensure a valid comparison to the image-based methods, we evaluate FG-ARI on the individual frames of the MOVi datasets, reinitializing slots for each frame.

\textbf{Results (\tab{tab:results-object-discovery-synthetic})\quad}
Somewhat surprisingly, the video-based methods do not perform better than the image-based methods.
For both STEVE and SAVi++, there is a large gap to our method.
STEVE performs similar to its image-based counterpart SLATE.
Compared to the results reported by~\citet{elsayed2022savi++}, SAVi++ performs considerably worse, falling behind its image counterpart Slot Attention.
Even when having access to optical flow supervision, SAVi++ does not produce meaningful object groupings.
We hypothesize this is because SAVi++ was not designed for the unconditional setting, and model and/or hyperparameter changes are needed to alleviate this.

\subsection{Results on the KITTI Driving Dataset}
\label{app:results-kitti}

We present prelimary results on KITTI~\citep{Alhaija2018Kitti}, consisting of real-world videos of city-street driving.
In particular, we evaluate on the instance segmentation annotations, containing 200 independently annotated frames.
For training, we use all 147 videos by randomly sampling frames from them, resulting in \numprint{95869} training images.
We compare with ``Discovering Objects that Can Move''~\citep{Bao2022ObjectsThatMove}, a recent video-based object-centric method that uses motion masks to supervise the attention masks.
To be consistent, we evaluate at the same mask resolution as them, namely at $1242 \times 375$ pixels.

\begin{wrapfigure}[13]{r}{0.37\textwidth}
    \centering
    \vspace{-1\intextsep}
    \captionof{table}{Object discovery on KITTI (mean $\pm$ standard dev., 5 seeds).}
    \label{app:tab:object-discovery-kitti}
    \setlength{\tabcolsep}{0.25em}
    \begin{tabular}{@{}lcc@{}}
    \toprule
    Method & FG-ARI & mBO \\
    \midrule
    \citet{Bao2022ObjectsThatMove} & 47.1 & \\
    \dinosaur & \meanandstd{70.28}{0.76} & \meanandstd{19.41}{0.26} \\
    \bottomrule
    \end{tabular}

    {
    \vspace{0.7em}
    \captionof{figure}{Example masks on KITTI.}
    \label{app:fig:example-kitti}
    {
    \centering
    \renewcommand{\arraystretch}{4.7}
    \setlength{\tabcolsep}{2pt}
    \vspace{-0.2em}
    \begin{tabular}{@{}cc@{}}
        \includegraphics[width=0.43\linewidth]{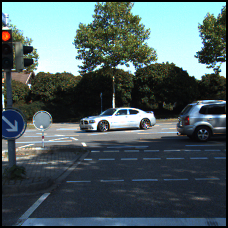} & \includegraphics[width=0.43\linewidth]{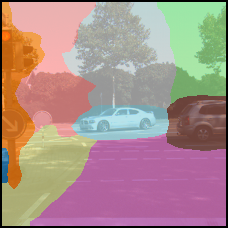} \\
    \end{tabular}
    }
    }
\end{wrapfigure}

Because of the skewed aspect ratio and high resolution, and because our model is trained on square images, we evaluate in a sliding window fashion.
In particular, we partition the image into 4 non-overlapping windows of size $375 \times 375$ (adding padding of 129 black pixels to both sides), process each image individually into $K$ masks of size $375 \times 375$ using our model, then join the masks spatially and removing padding.
See \fig{app:fig:example-kitti} for an example window.
We use the resulting $4 \cdot K$ masks of size $1242 \times 375$ for evaluation.
This kind of sliding window approach has the obvious flaw that objects can not span more than window.
We leave it to future work to improve working with high-resolution images.

\looseness-1
Tuning \dinosaur to work on KITTI was more difficult than for COCO or PASCAL VOC.
For instance, using the Transformer decoder did not result in object groupings, and we had to resort to the MLP decoder with a small slot size of 32.
We also had difficulties training a ResNet encoder on KITTI.
We hypothesize this is because the KITTI dataset does not have a lot of diversity, and so high-capacity architectures can easily overfit to the expected scene layout instead of learning to discover objects.

\textbf{Results (\tab{app:tab:object-discovery-kitti})\quad}
We train the model with $K=9$ slots, yielding $36$ slots for evaluation.
Our model clearly outperforms \citet{Bao2022ObjectsThatMove}'s method in terms of FG-ARI, even though it is trained with motion supervision.
However, there is ample room for improvement:
visual inspection shows that while our method manages to capture larger objects in the foreground (mainly cars), the many small objects in the background of the scene are grouped together with non-objects, \eg houses.
This explains why FG-ARI is relatively high (as large objects influence this metric more), while mBO is relatively low (as the metric is averaged over \emph{all} objects).
This hints at another difficulty of real-world data which current object-centric methods are not well equipped to deal with: diversity of object scales.

\subsection{Object Property Prediction on COCO}
\label{app:subsec:coco-property-prediction}

We study the usefulness of the learned representations for downstream applications by predicting \emph{object properties} from the slots on COCO.
To this end, we supervise shallow predictors using the slot representations from unsupervised object-centric models.
As properties, we use the class, and the x- and y-coordinates of each object (center-of-mass).

We closely follow the experimental design of \cite{dittadi2021generalization}, and did not attempt to tune their settings to optimize for performance.
As downstream models, we test a linear predictor, and a one-hidden layer MLP.
Each model is applied in parallel to all slots for an image, \ie it receives a slot as input, and outputs a vector of property predictions for this slot.
We use the cross-entropy loss for the categorical class prediction, and the MSE loss for the continuous x- and y-coordinates, where the overall loss is just the sum of the individual losses.
We match the predicted set of vectors to the target set of vectors using the Hungarian method~\citep{Kuhn95Hungarian}, minimizing the total loss of the assignment.
As metrics, we report Top-1 and Top-5 average-class accuracy for the 80-way class prediction and R$^2$ (coefficient of determination) for the x- and y-coordinates.

We use \numprint{10000} random images from the COCO dataset to train, and another \numprint{1000} random images as a validation set.
We report metrics on the \numprint{5000} validation images of COCO.
Images that contain no objects due to center cropping are filtered out.
As COCO images can contain many objects, but our method typically only uses a low number of slots (\eg 7), we keep only the 6 largest objects in the image.
This way, all ground truth objects can in principle be matched by a slot.
We train for \numprint{15000} steps, and use the model with the lowest validation loss for testing.

\begin{figure}
    \centering
    \includegraphics{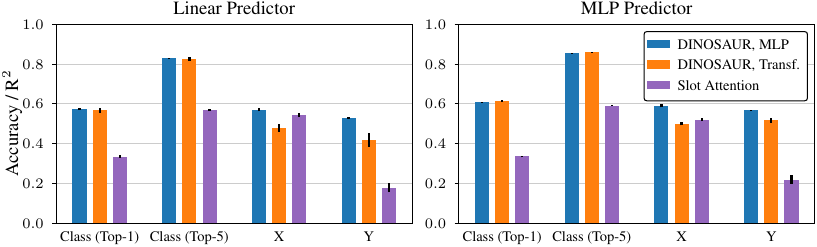}
    \caption{\emph{Object property prediction} on the validation set of COCO, showing mean and standard deviation over 3 seeds.
    We train a linear predictor (left) and a one-hidden layer MLP (right) to predict object properties from the slot representations.
    For class prediction, we show Top-1 and Top-5 accuracy.
    For coordinate prediction, we show R$^2$ (coefficient of determination, higher is better).
    We compare our method using MLP and Transformer decoders with Slot Attention.
    }
    \label{app:fig:coco-property-prediction}
\end{figure}

The results are listed in \fig{app:fig:coco-property-prediction}.
We compare \dinosaur using a DINO-pretrained ViT-B/16 encoder with a Slot Attention model, all using 7 slots.
On class prediction, our method has 25\% higher accuracy than Slot Attention.
On coordinate prediction, \dinosaur reaches similar performance to Slot Attention on the x-coordinate, but exceeds it on the y-coordinate, where Slot Attention drops-off.
We also find that on coordinate prediction, the representations from the Transformer decoder are a bit worse than from the MLP decoder.
This may be because the Transformer decoder often captures several disconnected objects in one slot, making it more difficult to predict the center-of-mass.
Using a non-linear instead of a linear predictor slightly improves accuracy on all properties; using more hidden layers than one did not improve the performance further in our experiments.
Overall, we conclude that the representation \dinosaur learns are more useful than Slot Attention's for downstream tasks on real-world data.
However, our method also still leaves room for improvement.
Finally, we note that this experiment only constitutes a first step in evaluating the learned representations, and that further investigations could be conducted, \eg taking into account robustness to distribution shifts.

\section{Expanded Comparison with Computer Vision Methods}
\label{app:comparison-cv-methods}

In this section, we extend the comparison to computer vision methods with a more detailed discussion.
In addition to the results from \sec{subsec:comp-cv-methods}, we present results on more datasets (PASCAL VOC 2012 for object localization, COCO for object segmentation, COCO-Stuff 171 for scene decomposition), additional metrics (detection rate for object localization, per-pixel accuracy for scene decomposition), and list further baseline methods.
Finally, we also run the unsupervised semantic segmentation method STEGO on the MOVi datasets (\sec{app:subsec:movi-stego}).

\subsection{Unsupervised Object Localization}
\label{subsec:unsup-object-localization}

\textbf{Datasets and Metrics\quad}
In the computer vision literature, the task of object localization refers to finding the location of one or more objects that repeatedly appear in a collection of images~\citep{Cho2015UnsupervisedOD}.
In particular, objects have to be discovered by proposing bounding boxes that ``correctly localize'' them.
An object counts as correctly localized if its bounding box has an IoU-overlap of at least 50\% with a proposed bounding box.
Like prior work~\citep{Vo2020OSD,Melaskyriazi2022DeepSpectral}, we report the fraction of images on which at least one object was correctly localized (CorLoc), and the average fraction of objects correctly localized per image (detection rate).
As our method produces masks, we use the tightest bounding box around each mask as the proposal.
We evaluate on the \texttt{trainval} split of PASCAL VOC 2012 and COCO-20k, as is standard in the literature~\citep{Vo2020OSD,Simeoni2021LOST}.
COCO-20k is a subset of COCO 2014 filtered to contain no instances annotated as crowd.

\textbf{Results (\tab{tab:results-unsup-object-localization})\quad}\looseness-1
The results show that our method manages to focus on the salient parts of the data.
On the majority of images, at least one object is captured by a slot.
Also, our method reaches results comparable or better to what has been reported in the literature on this task.
Interestingly, zero-shot transferring a model trained on COCO to PASCAL yields better results than a model trained on PASCAL itself; we explain this by the size of the training data, which is around $10\times$ larger for COCO.

\begin{figure}
    \begin{minipage}[t]{0.64\textwidth}
        \centering
        \captionof{table}{Object Localization (mean $\pm$ standard dev., 5 seeds). We report CorLoc and detection rate and compare with rOSD \citep{Vo2020OSD}, DINO-Seg \citep{Simeoni2021LOST}, LOST \citep{Simeoni2021LOST}, DeepSpectral \citep{Melaskyriazi2022DeepSpectral} and TokenCut \citep{Wang2022TokenCut}}
        \label{tab:results-unsup-object-localization}
        \setlength{\tabcolsep}{0.32em}
        \footnotesize
        \begin{tabular}{@{}lccccc@{}}
        \toprule
        & \multicolumn{2}{c}{\textbf{PASCAL VOC 2012}} & {} & \multicolumn{2}{c}{\textbf{COCO-20k}} \\
        \cmidrule{2-3} \cmidrule{5-6}
        & CorLoc & DetRate & \phantom{} & CorLoc & DetRate \\
        \midrule
        rOSD & 51.9 & 41.2 & & 48.5 & 12.0 \\
        DINO-Seg & 46.2 & -- & & 42.1 & -- \\
        LOST & 64.0 & -- & & 50.7 & -- \\
        DeepSpectral & 66.4 & -- & & 52.2 & -- \\
        TokenCut & 72.1 & -- & & 58.8 & -- \\
        \midrule
        Ours & \meanandstd{69.76}{4.85} & \meanandstd{52.06}{3.93} && \meanandstd{67.15}{1.46} & \meanandstd{31.03}{0.86} \\
        Ours (COCO Transfer) & \meanandstd{77.88}{0.97} & \meanandstd{58.36}{0.90} && & \\
        \bottomrule
        \end{tabular}
    \end{minipage}\hfill
    \begin{minipage}[t]{0.32\textwidth}
        \vspace{-3em}
        {
\centering
\newcommand{\myig}[1]{\includegraphics[width=0.49\linewidth,valign=c]{#1}}
\newcommand{\mycaption}[1]{{\footnotesize#1}}
\renewcommand{\arraystretch}{4.7}
\setlength{\tabcolsep}{1pt}
\begin{tabular}{@{}cc@{}}
    \mycaption{Ground Truth} & \mycaption{\dinosaur} \\[-1.4em]
    \myig{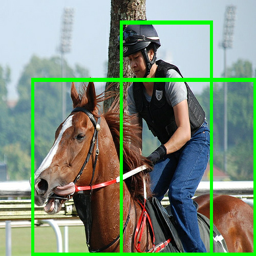} & \myig{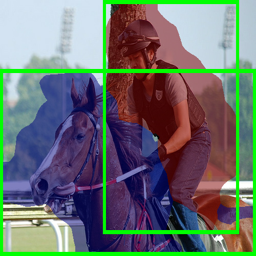} \\
    \myig{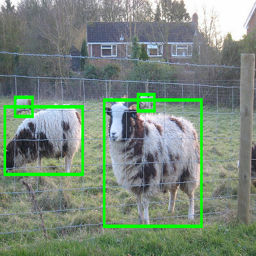} & \myig{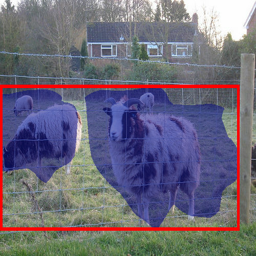} \\
\end{tabular}
}
        \vspace{-1.5em}
        \captionof{figure}{Object Localization on PASCAL VOC 2012.}
    \end{minipage}
\end{figure}

\subsection{Unsupervised Semantic Segmentation}
\label{subsec:unsup-semantic-segmentation}

\begin{table}[tb]
    \caption{Unsupervised Semantic Segmentation (mean $\pm$ standard dev., 5 seeds). We report mean intersection over union (mIoU). For scene decomposition, we additionally report per-pixel accuracy (pAcc). STEGO results are presented without CRF post-processing for fair comparison with other methods. Results marked $\dagger$ were produced ourselves using the official implementation.}
    \begin{subtable}[t]{.461\linewidth}
        \centering
        \caption{Object Segmentation. We compare with MaskContrast \citep{VanGansbeke2021MaskContrast}, DeepSpectral \citep{Melaskyriazi2022DeepSpectral}, MaskDistill \citep{VanGansbeke2022MaskDistill}, and COMUS \citep{Zadaianchuk2022COMUS}.}%
        \label{tab:results-unsup-object-seg}
        \setlength{\tabcolsep}{0.3em}
        \footnotesize
        \begin{tabular}{@{}lcc@{}}
        \toprule
        & \textbf{PASCAL VOC 2012} & \textbf{COCO} \\
        \midrule
        MaskContrast & 35.0 & --  \\
        DeepSpectral & 37.2  & -- \\
        MaskDistill & 45.8 & -- \\ %
        COMUS & 50.0 & 19.6 \\
        \midrule
        Ours & \meanandstd{37.15}{1.84} & \meanandstd{13.92}{0.40} \\
        \bottomrule
        \end{tabular}
    \end{subtable}\hfill
    \begin{subtable}[t]{.524\linewidth}
      \centering
        \caption{Scene Decomposition. We compare with IIC~\citep{Ji2019IIC}, SegDiscover~\citep{Huang2022SegDiscover}, PiCIE~\citep{Cho2021Picie}, SlotCon~\citep{Wen2022SlotCon}, and STEGO~\citep{Hamilton2022Stego}.}
        \label{tab:results-unsup-scene-decomp}
        \setlength{\tabcolsep}{0.18em}
        \footnotesize
        \begin{tabular}{@{}lccccc@{}}
        \toprule
        & \multicolumn{2}{c}{\textbf{COCO-Stuff 27}} & {} & \multicolumn{2}{c}{\textbf{COCO-Stuff 171}} \\
        \cmidrule{2-3} \cmidrule{5-6}
        & mIoU & pAcc & \phantom & mIoU & pAcc \\
        \midrule
        IIC & 6.7 & 21.8 & & -- & -- \\
        SegDiscover & 14.3 & 56.5 & & -- & -- \\
        PiCIE + H. & 14.4 & 50.0 & & -- & -- \\
        SlotCon & 18.3 & 42.4 & & -- & -- \\
        STEGO & 26.8 & 54.8 & & 10.0$^\dagger$ & 32.5$^\dagger$  \\  %
        \midrule
        Ours & \meanandstd{23.99}{0.85} & \meanandstd{44.85}{1.17} & & \meanandstd{13.13}{0.26} & \meanandstd{27.03}{0.31} \\
        \bottomrule
        \end{tabular}
    \end{subtable}
\end{table}

In this task, each pixel of the image has to be assigned a semantic class label.
As in the unsupervised setting, no direct correspondence to ground truth classes are given, a successful model needs to output segments that consistently cover the same semantic content on different images.

\textbf{Tasks, Metrics and Baselines}\quad
We report results on two different versions of this task: in object segmentation, only foreground classes have to segmented; in scene decomposition, the background has to be split into semantic categories as well.
For object segmentation, we evaluate on PASCAL VOC 2012 and COCO 2017, with instance masks merged to semantic class masks.
For scene decomposition, we evaluate on the COCO-Stuff dataset~\citep{Caesar2018COCOStuff}, in two different variants:
COCO-27 uses a reduced set of $27$ coarse classes from the COCO class hierarchy, whereas COCO-171 uses the full set of $80$ foreground and $91$ background classes available in the dataset.
To the best of our knowledge, we are the first to report results on this challenging setting.
We evaluate with mean intersection-over-union over classes (mIoU) and per-pixel accuracy (pAcc).

\textbf{Details on Obtaining a Semantic Segmentation with DINOSAUR~\quad}
For unsupervised semantic segmentation, we need to assign a discrete label to each discovered mask.
As our method does not directly output labels required for semantic segmentation, we obtain them by running K-Means clustering on features associated with each slot after training the model, then assigning clusters to ground truth classes by maximizing IoU using Hungarian matching.
Similar ways of evaluating are commonly used in the literature~\citep{VanGansbeke2021MaskContrast,Melaskyriazi2022DeepSpectral}.
In particular, for each slot generated by the model, we compute a feature vector by taking a weighted average between the slot's mask and ViT features (see below).
Each vector is then L2-normalized, and the set of vectors from all images is clustered by running k-means clustering, and each slot is associated with a cluster.
Finally, the clusters are assigned to ground truth classes by maximizing the total IoU between clusters and classes using the Hungarian method~\citep{Kuhn95Hungarian}.
For K-Means clustering, to avoid sensitivity to initialization, we run 20 repetitions and use the run with the lowest sum of squared distances to the cluster centers.
As we noticed there is still some variance between different evaluation runs, we run the inference and clustering procedure three times and use the average as the final value for the training run.

\begin{wrapfigure}[10]{r}{0.39\textwidth}
    \centering
    \vspace{-1\intextsep}
    \captionof{table}{Comparing ViT-B/16 blocks and slot features for clustering on COCO-27 (mIoU).}
    \vspace{-0.2em}
    \small
    \label{app:tab:clustering-features}
    \setlength{\tabcolsep}{0.25em}
    \begin{tabular}{@{}lccc@{}}
    \toprule
    Method & Slots & Block 9 & Block 12 \\
    \midrule
    Supervised & \meanandstd[m]{8.04}{1.26} & \meanandstd[m]{18.01}{0.31} & \meanandstd[m]{13.30}{0.15} \\
    DINO & \meanandstd[m]{13.43}{1.19} & \meanandstd[m]{23.99}{0.85} & \meanandstd[m]{21.11}{0.70} \\
    MoCo-v3 & \meanandstd[m]{13.46}{0.37} & \meanandstd[m]{20.48}{0.44} & \meanandstd[m]{22.37}{0.47} \\
    MSN & \meanandstd[m]{11.85}{0.40} & \meanandstd[m]{21.76}{0.33} & \meanandstd[m]{16.76}{0.33} \\
    MAE & \meanandstd[m]{6.67}{0.21} & \meanandstd[m]{10.08}{0.28} & \meanandstd[m]{11.20}{0.74} \\
    \bottomrule
    \end{tabular}
\end{wrapfigure}
\emph{Features for clustering (\tab{app:tab:clustering-features}).} \looseness-1
Which features should we use for clustering the slots?
We initially experimented with the slots themselves, but found that they do not cluster well.
Instead, we use a weighted average of the slots masks and pre-trained ViT features to produce a feature to cluster for each slot.
We compare features from ViT block 9 and ViT block 12 and find that the best block depends on the training method.
In the following, we use block 9 for supervised training, DINO and MSN, and block 12 for MoCo-v3 and MAE.

\vspace{0.5em}
\emph{Number of clusters.}
Results for unsupervised semantic segmentation are dependent on the number of clusters used.
Prior methods~\citep{Hamilton2022Stego,VanGansbeke2022MaskDistill} typically use the number of classes labeled in the datasets as the number of clusters.
For unsupervised object segmentation, this presents an issue for our method, as it captures entities in both foreground and background, and thus clustering using the number of (foreground) object classes leads to over-merging of classes.
To ensure a fairer comparison, we instead chose the number of clusters by estimating the how many foreground and background classes can appear on the different datasets.
In particular, for PASCAL VOC 2012, we use 105 clusters (the number of classes occurring on at least 0.5\% of images, using the PASCAL-Context~\citep{Mottaghi14PASCALcontext} labeling and class statistics), and for COCO, we use 172 clusters (the number of classes labeled on the COCO-Stuff dataset~\citep{Caesar2018COCOStuff}).
As this results in more clusters than ground truth classes, Hungarian matching leaves some clusters unassigned; we treat all unassigned clusters as being assigned to the ``background'' class.
For scene decomposition, we simply use 27 clusters for COCO-Stuff 27 and 172 clusters for COCO-Stuff 172.

\textbf{Results on Object Segmentation (\tab{tab:results-unsup-object-seg})\quad}
This task is not naturally suited for our method, which decomposes the scene into semantically meaningful parts without distinguishing fore- and background.
We still obtain competitive results with recently published work~\citep{VanGansbeke2021MaskContrast,Melaskyriazi2022DeepSpectral}.
Moreover, DeepSpectral, MaskDistill and COMUS employ additional steps of training segmentation networks which improves results and allows them to run at the original image resolution.
In contrast, the masks we evaluate are only of size $14 \times 14$; we leave it to future work to improve the resolution of the produced masks.

\textbf{Results on Scene Decomposition (\tab{tab:results-unsup-scene-decomp})\quad}
We are competitive with the state-of-the-art method STEGO on the setting with 27 coarse classes.
On the novel setting with all 171 classes, our method improves upon STEGO by 3.2 IoU points.
This demonstrates that \dinosaur is able to handle the full complexity of real-world scenes somewhat better.

\subsection{Unsupervised Semantic Segmentation Baseline (STEGO) on MOVi}
\label{app:subsec:movi-stego}

\begin{wrapfigure}[9]{r}{0.37\textwidth}
    \centering
    \vspace{-1\intextsep}
    \captionof{table}{Results for STEGO on the MOVi datasets.}
    \vspace{-0.2em}
    \small
    \label{app:tab:movi-stego}
    \setlength{\tabcolsep}{0.25em}
    \begin{tabular}{@{}lcccc@{}}
    \toprule
    Dataset & Clusters & ARI & FG-ARI & mBO \\
    \midrule
    \multirow{2}{*}{MOVi-C} & 22 & 40.0 & 20.1 & 24.1 \\
     & 27 & 39.5 & 19.8 & 23.8 \\
    \midrule
    \multirow{2}{*}{MOVi-E} & 22 & 41.1 & 15.6 & 11.5 \\
     & 27 & 41.6 & 15.8 & 12.0 \\
    \bottomrule
    \end{tabular}
\end{wrapfigure}
An interesting question is to what degree unsupervised semantic segmentation methods (which we compare to on real-word data in \sec{subsec:unsup-semantic-segmentation}) can already solve the object discovery task on synthetic datasets designed to benchmark object-centric learning methods.
If this were the case, it would showcase that those datasets are flawed in the sense that instance-level grouping (one goal of object-centric learning) is not needed to tackle them.
To this end, we test the state-of-the-art scene segmentation method STEGO~\citep{Hamilton2022Stego} on the MOVi datasets.

STEGO assigns each pixel to one of several clusters/classes, with the clustering performed over the full dataset.
We run STEGO with 22 and 27 clusters, corresponding to the 17 semantic categories of the MOVi dataset (see \sec{app:sec:dataset-and-eval-settings}), plus 5 or 10 clusters which can be used for the backgrounds.
We use a DINO ViT-B/8 backbone for feature extraction, other hyperparameters are listed in \sec{app:subsec:details-baselines}.
To evaluate on the object discovery task, we use the segmentation masks produced by STEGO as if they were object masks.
That is, each cluster/class mask is used as a separate object mask.

\begin{figure}
    \centering
    {
\centering
\newcommand{\myig}[1]{\includegraphics[width=0.115\textwidth,valign=c]{#1}}
\newcommand{\mycaption}[1]{{\rotatebox[origin=c]{90}{\footnotesize#1}}}
\renewcommand{\arraystretch}{3.5}
\setlength{\tabcolsep}{1pt}
\begin{tabular}{@{}rccccccccc@{}}
    & \multicolumn{4}{c}{\textbf{MOVi-C}} & \phantom{x} & \multicolumn{4}{c}{\textbf{MOVi-E}} \\[-0.5em]
    \mycaption{Image} & \myig{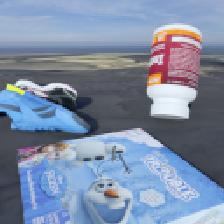} & \myig{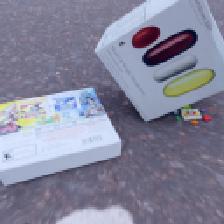} & \myig{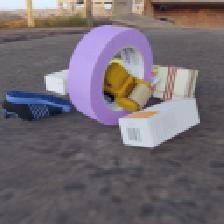} & \myig{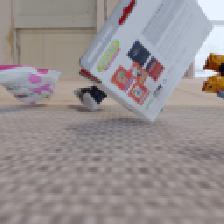} & \phantom & \myig{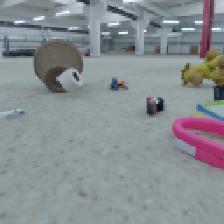} & \myig{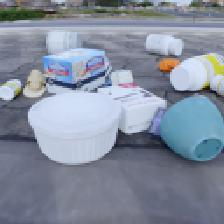} & \myig{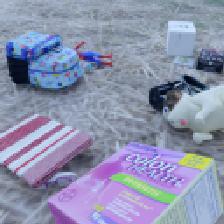} & \myig{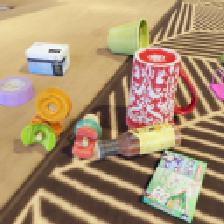} \\
    \mycaption{STEGO} & \myig{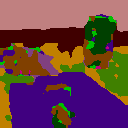} & \myig{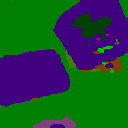} & \myig{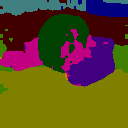} & \myig{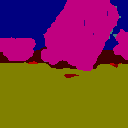} & \phantom & \myig{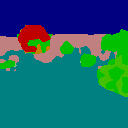} & \myig{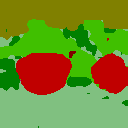} & \myig{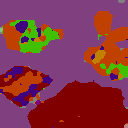} & \myig{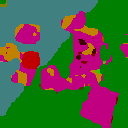}\\
\end{tabular}
}
    \vspace{-1.5em}
    \caption{Running the unsupervised semantic segmentation method STEGO~\citep{Hamilton2022Stego} on the MOVi datasets, using 22 clusters for MOVi-C and 27 clusters for MOVi-E. Color codes correspond to the different clusters. Results are presented without CRF post-processing.}
    \label{app:tab:movi-stego-examples}
\end{figure}

We list quantitative results in \tab{app:tab:movi-stego} and show examples in \fig{app:tab:movi-stego-examples}.
In addition to foreground ARI (FG-ARI) and mean best overlap (mBO), we list ARI, which takes the background pixels into account as well.
STEGO is able to segment objects and background to some degree.
However, objects are not properly split, with a single cluster often covering several objects from different semantic categories.
Compared to our method, STEGO's masks are also significantly more cluttered and less geometrically consistent.
Quantitatively, STEGO performs significantly worse than all object-centric methods.
There is a large difference between ARI and FG-ARI, indicating that STEGO is often able to separate out the background, but struggles to separate the objects.
For mBO on MOVi-C, STEGO is only 2 points worse than our Slot Attention and STEVE baselines, showing that on scenes with a moderate amount of objects, the masks can capture the objects to some degree (but do not succeed in separating them, see \fig{app:tab:movi-stego-examples}).
Overall, this result suggests that a current state-of-the-art unsupervised semantic segmentation method can not satisfactorily solve the object discovery task, justifying the existence of datasets like MOVi to study object-centric learning methods.

\begin{figure}[th]
    \centering
    \newcommand{\myig}[1]{\includegraphics[width=0.55\textwidth,valign=c]{#1}}
    \renewcommand{\arraystretch}{2.7}
    \setlength{\tabcolsep}{3pt}
    \begin{tabular}{@{}rc@{}}
        Inputs & \myig{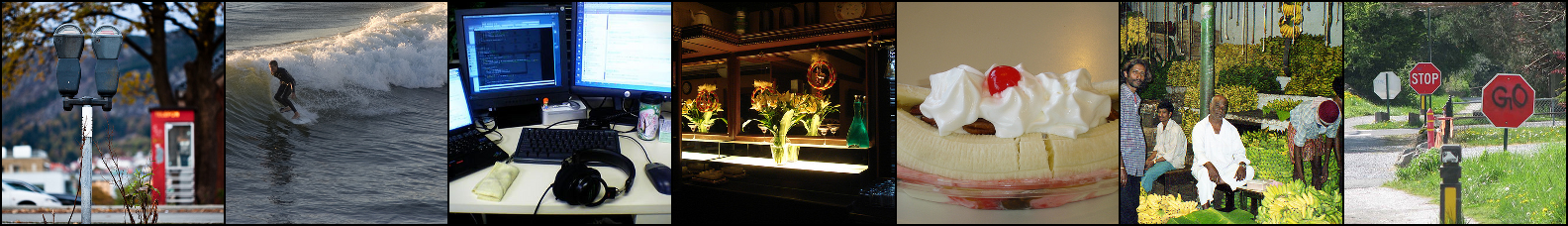} \\
        Reconstructions & \myig{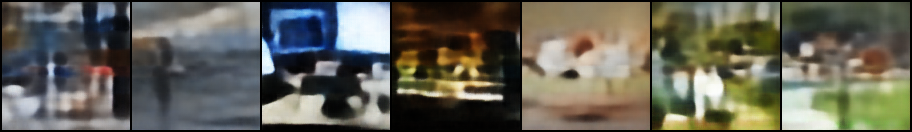} \\
        Masks & \myig{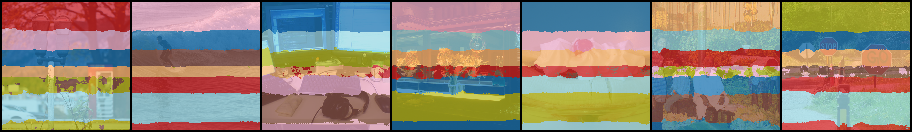} \\
    \end{tabular}
    \caption{Finetuning a DINO pre-trained ViT encoder using image reconstruction as a training signal fails to yield meaningful results and results in striped mask patterns (see discussion in \sec{subsec:analysis}).}
    \label{app:fig:analysis-image-recon}
\end{figure}

\newpage

\section{Additional Analysis}
\label{app:sec:additional-experiments}

In this section, we conduct several experiments that did not fit into the main part.
A brief overview:

\begin{itemize}[topsep=0pt,align=parleft,left=1pt..4em]
    \item[\tab{app:tab:decoder-ablation-additional-results}:] Results on object localization and semantic segmentation for decoder analysis from \sec{subsec:analysis}.
    \item[\sec{app:subsec:choice-pretraining-method}:] Comparing different pre-training schemes for feature extraction and reconstruction.
    \item[\sec{app:subsec:training-encoder-scratch}:] Comparing training a ResNet encoder from scratch \vs using a pre-trained ViT encoder.
    \item[\sec{app:subsec:exp-number-of-slots}:] How does the number of slots affect the quantitative and qualitative results of our method, both at training and test time?
    \item[\sec{app:subsec:exp-decoder-analysis}:] How does decoder scale affect the quality of object discovery?
    \item[\sec{app:subsec:exp-transformer-decoder}:] Analyzing the poor performance of the Transformer decoder with larger ViTs.
    \item[\sec{app:subsec:exp-mask-type}:] Comparing the quality of masks from slot attention and the Transformer decoder.
    \item[\sec{app:subsec:movi-semantic}:] Does \dinosaur perform \emph{semantic segmentation} on the MOVi datasets?
\end{itemize}

\begin{table*}[thb]
    \centering
    \caption{Additional results for decoder analysis in \fig{tab:analysis-decoders} on \emph{Object Localization} and \emph{Unsupervised Semantic Segmentation} (mean $\pm$ standard dev., 3 seeds), using a ViT-B/16 encoder and 6 slots (PASCAL VOC 2012) or 7 slots (COCO).
    For localization, results are given on PASCAL VOC 2012 \texttt{trainval} (training on PASCAL VOC 2012) and COCO-20k (training on COCO).
    For segmentation, results are given for object segmentation (PASCAL VOC 2012, COCO) and scene decomposition (Stuff-27, Stuff-171); when the training dataset is different from the evaluation dataset, this tests zero-shot transfer (\ie PASCAL $\rightarrow$ COCO and COCO $\rightarrow$ PASCAL).
    We also refer to \figs{app:fig:examples-extended-pascal} and \ref{app:fig:examples-extended-coco} for side-by-side comparisons of the two types of decoders.
    }
    \label{app:tab:decoder-ablation-additional-results}
    \setlength{\tabcolsep}{0.5em}
    \begin{tabular}{@{}llccccccc@{}}
    \toprule
    & & \multicolumn{2}{c}{\textbf{Object Localization}} & {} & \multicolumn{4}{c}{\textbf{Semantic Segmentation (mIoU)}} \\
    \cmidrule{3-4} \cmidrule{6-9}
    Training & Decoder & CorLoc & DetRate & \phantom & COCO & Stuff-27 & Stuff-171 & PASCAL\\
    \midrule
    \multirow{2}{*}{PASCAL} & MLP & \meanandstd{76.03}{0.26} & \meanandstd{56.33}{0.16} && \meanandstd{12.45}{0.38} & \meanandstd{20.97}{0.34} & \meanandstd{11.31}{0.13} & \meanandstd{33.69}{0.92} \\
    & Transformer & \meanandstd{69.76}{4.85} & \meanandstd{52.06}{3.93} && \meanandstd{11.87}{1.00} & \meanandstd{22.22}{0.75} & \meanandstd{11.59}{0.64} & \meanandstd{37.15}{1.84} \\
    \midrule
    \multirow{2}{*}{COCO} & MLP & \meanandstd{69.33}{0.48} & \meanandstd{29.80}{0.29} && \meanandstd{12.39}{0.24} & \meanandstd{20.32}{0.22} & \meanandstd{11.68}{0.03} & \meanandstd{32.28}{0.80} \\
    & Transformer & \meanandstd{67.15}{1.46} & \meanandstd{31.03}{0.86} && \meanandstd{13.92}{0.40} & \meanandstd{23.99}{0.85} & \meanandstd{13.13}{0.26} & \meanandstd{38.64}{1.12} \\
    \bottomrule
    \end{tabular}
\end{table*}

\subsection{Choice of Pre-Training Method}
\label{app:subsec:choice-pretraining-method}

We compare different different backbones and pre-training schemes in \tab{app:tab:encoder-ablation-additional-results}.
We list results on object discovery, object localization and unsupervised semantic segmentation for a comprehensive overview.
Different from~\tab{tab:analysis-self-supervised-targets} in the main part, here we consider the setting with pre-trained, frozen encoders.
We also investigate target features from ResNet-50 encoders and differently-sized ViTs.
When using ResNet50 encoders, we resorted to the MLP decoder as the Transformer decoder did not yield good performance when reconstructing ResNet features.

As in \tab{tab:analysis-self-supervised-targets}, we find that all self-supervised schemes perform well for the task of object discovery (\cf \fig{app:fig:examples-coco-encoders}).
In comparison, supervised training via ImageNet classification is a worse source of features.
Moreover, there is a gap between ViT and ResNet features: while ARI is high (likely a by-product of using the MLP decoder), mBO is poor.
Visual inspection shows that ResNet-trained models seem to rely on a mix of semantic and geometric grouping of the image (\cf \fig{app:fig:examples-coco-encoders-mlp}).
Generally, larger ViT architectures also yield better results.
We conclude from this experiment that scaling the architecture used for the encoder and target features yields better results.
Furthermore, we find that using \emph{self-supervised} training is important for real-world grouping to emerge, but that the specific method is less important.

Finally, we also point to \tab{app:tab:analysis-encoder-scratch}, where we list results for different self-supervised schemes, but training with the MLP decoder instead of the Transformer decoder.
This is interesting because MAE, the worst among the self-supervised methods with the Transformer decoder, yields the overall best results in terms of ARI on COCO object discovery: 42.3.
This suggests that there are still details about how exactly self-supervised training methods interact with object-centric feature reconstruction left open for investigation.

\begin{table*}[htb]
    \centering
    \caption{
    Comparing different pre-trained models for feature extraction \emph{and} reconstruction target (mean $\pm$ standard dev., 3 seeds), using 7 slots, the Transformer decoder for ViTs, and the MLP decoder for ResNets.
    For discovery, results are given on COCO.
    For localization, results are given on COCO-20k.
    For segmentation, results are given for object segmentation (COCO) and scene decomposition (Stuff-27, Stuff-171).
    We refer to \fig{app:fig:examples-coco-encoders} for side-by-side comparisons.
    }
    \label{app:tab:encoder-ablation-additional-results}
    \setlength{\tabcolsep}{0.25em}
    \resizebox{\textwidth}{!}{
    \begin{tabular}{@{}llcccccccccc@{}}
    \toprule
    & & \multicolumn{3}{c}{\textbf{Object Discovery}} & {} & \multicolumn{2}{c}{\textbf{Object Localization}} & {} & \multicolumn{3}{c}{\textbf{Semantic Segmentation (mIoU)}} \\
    \cmidrule{3-5} \cmidrule{7-8} \cmidrule{10-12}
    Encoder & Pretraining & FG-ARI & \mBOinst & \mBOclass & \phantom & CorLoc & DetRate & \phantom & COCO & Stuff-27 & Stuff-171 \\
    \midrule
    ResNet50 & Supervised & \meanandstd{36.03}{0.25} & \meanandstd{22.05}{0.11} & \meanandstd{24.16}{0.11} && \meanandstd{44.21}{0.64} & \meanandstd{15.01}{0.31} && \meanandstd{9.05}{0.09} & \meanandstd{13.73}{0.26} & \meanandstd{7.70}{0.05} \\
    ResNet50 & DINO & \meanandstd{36.01}{0.54} & \meanandstd{22.87}{0.40} & \meanandstd{25.09}{0.44} && \meanandstd{44.28}{2.94} & \meanandstd{15.44}{1.29} && \meanandstd{7.01}{0.14} & \meanandstd{12.76}{0.14} & \meanandstd{6.35}{0.07} \\
    ResNet50 & MoCo-v3 & \meanandstd{33.69}{0.48} & \meanandstd{21.05}{0.15} & \meanandstd{23.20}{0.17} && \meanandstd{35.16}{1.02} & \meanandstd{11.22}{0.45} && \meanandstd{6.30}{0.04} & \meanandstd{11.87}{0.23} & \meanandstd{5.40}{0.08} \\
    ViT-S/16 & Supervised & \meanandstd{26.04}{0.15} & \meanandstd{24.28}{0.11} & \meanandstd{30.20}{0.05} && \meanandstd{56.23}{0.65} & \meanandstd{24.49}{0.32} && \meanandstd{18.20}{0.20} & \meanandstd{14.42}{0.50} & \meanandstd{13.47}{0.09} \\ %
    ViT-S/16 & DINO & \meanandstd{36.87}{0.61} & \meanandstd{29.68}{0.65} & \meanandstd{33.85}{0.97} && \meanandstd{68.64}{1.90} & \meanandstd{30.23}{1.34} && \meanandstd{11.24}{0.07} & \meanandstd{20.33}{0.06} & \meanandstd{10.60}{0.23} \\ %
    ViT-B/16 & Supervised & \meanandstd{32.25}{0.80} & \meanandstd{26.32}{0.32} & \meanandstd{31.53}{0.28} && \meanandstd{59.11}{1.06} & \meanandstd{25.49}{0.40} && \meanandstd{18.59}{0.22} & \meanandstd{13.30}{0.15} & \meanandstd{13.46}{0.08} \\ %
    ViT-B/16 & DINO & \meanandstd{34.14}{1.03} & \meanandstd{31.63}{0.69} & \meanandstd{39.72}{0.94} && \meanandstd{67.19}{1.45} & \meanandstd{31.06}{0.85} && \meanandstd{13.92}{0.40} & \meanandstd{23.99}{0.85} & \meanandstd{13.13}{0.26} \\ %
    ViT-B/16 & MoCo-v3 & \meanandstd{35.21}{0.24} & \meanandstd{31.41}{0.18} & \meanandstd{38.49}{0.49} && \meanandstd{69.95}{0.53} & \meanandstd{32.30}{0.13} && \meanandstd{14.27}{0.21} & \meanandstd{22.37}{0.47} & \meanandstd{13.37}{0.23} \\ %
    ViT-B/16 & MSN & \meanandstd{35.20}{0.34} & \meanandstd{31.05}{0.20} & \meanandstd{37.23}{0.38} && \meanandstd{69.67}{0.65} & \meanandstd{31.85}{0.38} && \meanandstd{13.28}{0.22} & \meanandstd{21.76}{0.33} & \meanandstd{12.22}{0.21} \\ %
    ViT-B/16 & MAE & \meanandstd{32.78}{3.66} & \meanandstd{30.21}{1.78} & \meanandstd{33.19}{1.81} && \meanandstd{68.91}{5.15} & \meanandstd{29.78}{3.01} && \meanandstd{9.21}{0.39} & \meanandstd{11.20}{0.74} & \meanandstd{7.25}{0.17} \\ %
    ViT-S/8 & DINO & \meanandstd{34.32}{0.49} & \meanandstd{32.29}{0.35} & \meanandstd{38.83}{0.38} && \meanandstd{72.96}{0.16} & \meanandstd{33.80}{0.03} && \meanandstd{13.41}{0.11} & \meanandstd{23.71}{0.29} & \meanandstd{13.24}{0.11} \\ %
    \bottomrule
    \end{tabular}
    }
\end{table*}

\subsection{Comparison of ResNet and Pre-Trained ViT Encoders}
\label{app:subsec:training-encoder-scratch}
In this experiment, we compare training a ResNet34 encoder from scratch \vs using a pre-trained, frozen ViT encoder (\ie our primary setup).
As the reconstruction target, we still use features from a pre-trained ViT.
Notably, we were not able to optimize the ResNet from scratch using the Transformer decoder, as the model degraded to only use a single slot for the full image.
Instead, here we use the MLP decoder with both ResNet34 and pre-trained ViT encoders.

\tab{app:tab:analysis-encoder-scratch} lists the results.
Example predictions on COCO are included in \tab{app:fig:examples-coco-encoders-mlp} and \tab{app:fig:examples-coco-encoders-mlp-2}.
We find that training from scratch is a viable alternative to using a pre-trained encoder, in some cases improving over the pre-trained encoder.
We think this is because training from scratch allows adaptation to the domain, whereas the pre-trained encoder may to some degree be misaligned.
On COCO, the improvement may also partially stem from the ResNet's higher spatial resolution ($28 \times 28)$ --- although the decoder's mask resolution is the same ($14 \times 14$).
However, using a frozen encoder has computational advantages: no backward pass is needed, and only one forward pass through the encoder is needed (instead of one pass through the encoder being trained, and one pass through the encoder providing the target features).
Finally, we find that pre-trained ResNet features are a worse reconstruction target then ViT features.
Overall, this experiment shows that the crucial element making our approach work is feature reconstruction.
While the type of encoder and the use of pre-trained weights can yield further improvements, they are less important compared to the use of feature reconstruction.

\begin{table*}[tb]
    \centering
    \captionof{table}{Training a ResNet34 encoder \emph{from scratch} vs.~using a \emph{pre-trained frozen} ViT encoder on \emph{object discovery} (mean $\pm$ standard dev., 3 seeds). Both options use the same target features for reconstruction and the MLP decoder.
    Training from scratch yields overall similar results compared to using a pre-trained encoder.
    We refer to \figs{app:fig:examples-coco-encoders-mlp} and \ref{app:fig:examples-coco-encoders-mlp-2} for side-by-side comparisons.
    }
    \label{app:tab:analysis-encoder-scratch}
    \begin{tabular}{@{}ll@{\hskip 0.4em}ll@{\hskip 0.6em}lcc@{\hskip 0.7em}cc@{}}
    \toprule
    Dataset & \multicolumn{2}{l}{Target Features} & Encoder & Training & FG-ARI & \multicolumn{2}{c}{mBO$^{(i,c)}$} \\
    \midrule
    \multirow{2}{*}{MOVi-C} & \multirow{2}{*}{DINO} & \multirow{2}{*}{ViT-S/8} &
    ResNet34 & Scratch & \meanandstd{65.11}{2.07} & \multicolumn{2}{c}{\meanandstd{29.11}{1.95}} \\
    & & & ViT-S/8 & Frozen & \meanandstd{67.21}{0.33} & \multicolumn{2}{c}{\meanandstd{38.63}{0.14}} \\
    \midrule
    \multirow{2}{*}{MOVi-E} & \multirow{2}{*}{DINO} & \multirow{2}{*}{ViT-S/8} &
    ResNet34 & Scratch & \meanandstd{64.49}{0.60} & \multicolumn{2}{c}{\meanandstd{36.52}{0.11}} \\
    & & & ViT-S/8 & Frozen & \meanandstd{64.66}{0.72} & \multicolumn{2}{c}{\meanandstd{34.10}{0.12}} \\
    \midrule
    \multirow{2}{*}{PASCAL} & \multirow{2}{*}{DINO} & \multirow{2}{*}{ViT-B/16} &
    ResNet34 & Scratch & \meanandstd{27.36}{0.74} & \meanandstd{38.32}{0.46} & \meanandstd{39.15}{0.45} \\
    & & & ViT-B/16 & Frozen & \meanandstd{24.61}{0.21} & \meanandstd{39.54}{0.12} & \meanandstd{40.85}{0.10} \\
    \midrule
    \multirow{12}{*}{COCO} & \multirow{4}{*}{DINO} & \multirow{2}{*}{ResNet50} &
    ResNet34 & Scratch & \meanandstd{36.57}{0.34} & \meanandstd{26.18}{0.15} & \meanandstd{28.80}{0.14} \\
    & & & ResNet50 & Frozen & \meanandstd{36.01}{0.54} & \meanandstd{22.87}{0.40} & \meanandstd{25.09}{0.44} \\
    \cmidrule{3-8}
    & & \multirow{2}{*}{ViT-B/16} & ResNet34 & Scratch & \meanandstd{40.91}{0.19} & \meanandstd{27.93}{0.04} & \meanandstd{31.14}{0.05} \\
    & & & ViT-B/16 & Frozen & \meanandstd{40.54}{0.03} & \meanandstd{27.67}{0.19} & \meanandstd{30.87}{0.21} \\
    \cmidrule{2-8}
     & \multirow{4}{*}{MoCo-v3} & \multirow{2}{*}{ResNet50} &
    ResNet34 & Scratch & \meanandstd{31.90}{1.06} & \meanandstd{23.87}{0.51} & \meanandstd{26.66}{0.40}  \\
    & & & ResNet50 & Frozen & \meanandstd{33.69}{0.48} & \meanandstd{21.05}{0.15} & \meanandstd{23.20}{0.17} \\
    \cmidrule{3-8}
    & & \multirow{2}{*}{ViT-B/16} & ResNet34 & Scratch & \meanandstd{40.40}{0.57} & \meanandstd{28.06}{0.16} & \meanandstd{31.12}{0.11}  \\
    & & & ViT-B/16 & Frozen & \meanandstd{39.79}{0.19} & \meanandstd{27.42}{0.10} & \meanandstd{30.44}{0.10} \\
    \cmidrule{2-8}
     & \multirow{2}{*}{MSN} & \multirow{2}{*}{ViT-B/16} &
    ResNet34 & Scratch & \meanandstd{40.73}{0.30} & \meanandstd{27.57}{0.12} & \meanandstd{30.65}{0.12} \\
    & & & ViT-B/16 & Frozen & \meanandstd{38.99}{0.10} & \meanandstd{28.14}{0.03} & \meanandstd{31.62}{0.07} \\
    \cmidrule{2-8}
     & \multirow{2}{*}{MAE} & \multirow{2}{*}{ViT-B/16} &
    ResNet34 & Scratch & \meanandstd{37.73}{0.12} & \meanandstd{28.10}{0.07} & \meanandstd{31.70}{0.03} \\
    & & & ViT-B/16 & Frozen & \meanandstd{42.30}{0.27} & \meanandstd{29.05}{0.12} & \meanandstd{32.18}{0.15} \\
    \bottomrule
    \end{tabular}

\end{table*}

\subsection{Number of Slots Analysis} \label{app:subsec:exp-number-of-slots}

\subsubsection{Training Sensitivity}
\label{app:subsubsec:slots-training-sensitivity}

\textbf{MOVi-E (\tab{app:tab:results-movi-e-num-slots})\quad}
On the synthetic MOVi-E dataset, we found that changing the number of slots can greatly impact performance.
We attribute this to the strong impact object splitting (distributing an object over two or more slots) can have on the ARI score.
While setting the number of slots to the maximum number of objects that can appear in the scene (23 on MOVi-E) allows the model to fully cover crowded scenes, most scenes have fewer objects, leading to impaired performance on those.
We show examples on MOVi-E with 11 slots for all tested methods in \fig{app:fig:examples-extended-movi-e-11slots}.

\begin{figure}[thb]
    \centering
    {
\centering
\newcommand{\myig}[1]{\includegraphics[width=0.15\textwidth,valign=c]{#1}}
\renewcommand{\arraystretch}{4.5}
\setlength{\tabcolsep}{1.5pt}
\vspace{-2em}
\begin{tabular}{@{}ccccccc@{}}
    5 slots & 7 slots & 9 slots & 11 slots & 13 slots & 15 slots \\[-1em]
    \myig{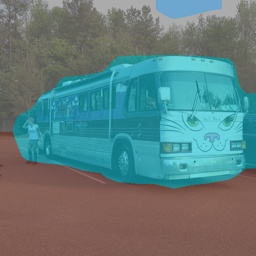} & \myig{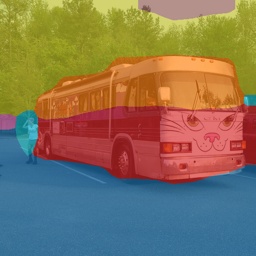} & \myig{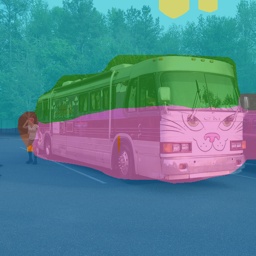} & \myig{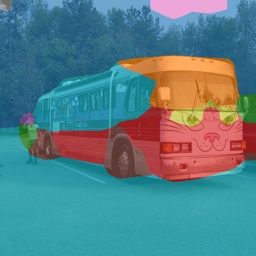} & \myig{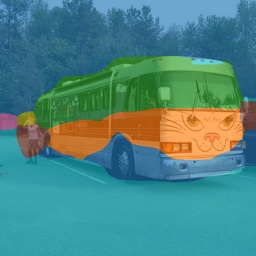}& \myig{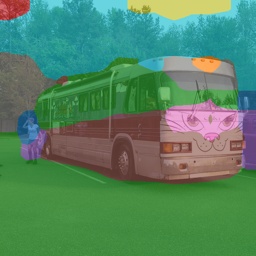} \\
    \myig{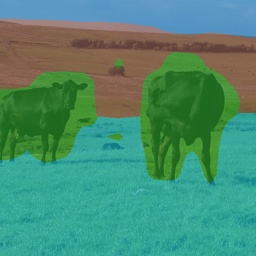} & \myig{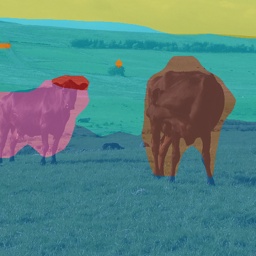} & \myig{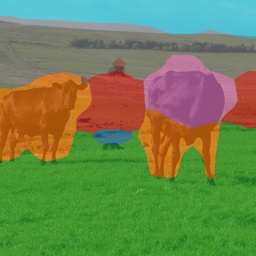} & \myig{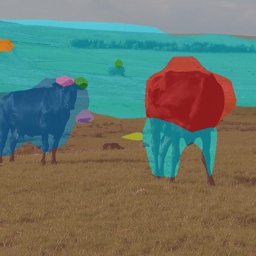} & \myig{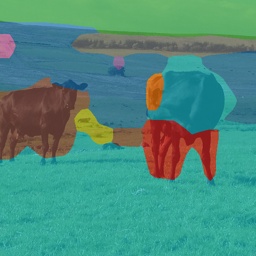}& \myig{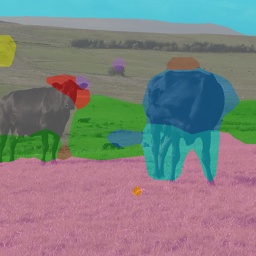} \\
    \myig{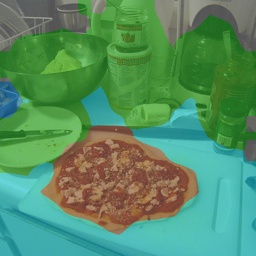} & \myig{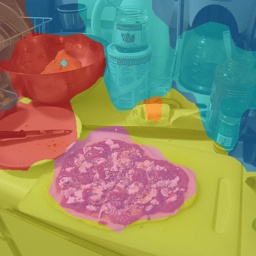} & \myig{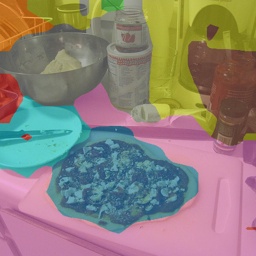} & \myig{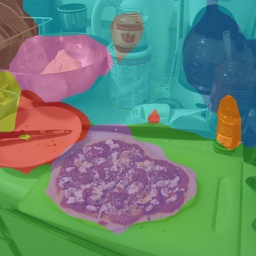} & \myig{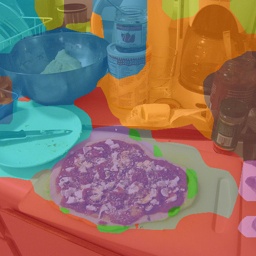}& \myig{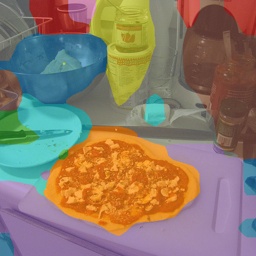} \\
\end{tabular}
}
    \caption{Three scenes of low/medium/high complexity on COCO 2017, varying the number of slots the model is trained with, using a ViT-B/16 encoder and the Transformer decoder.
    For an appropriate grouping to emerge, the number of slots has to fit the complexity of the scene.
    However, the groupings found by our method are still semantically meaningful, grouping together related objects (too few slots) or separating objects into their constituents parts (too many slots).
    }
    \label{app:fig:examples-coco-num-slots}
\end{figure}

\begin{figure}[tbh]
    \centering
    \includegraphics[width=\textwidth]{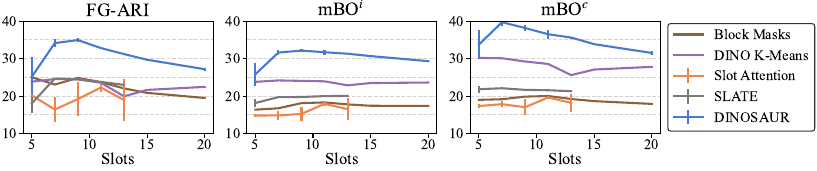}
    \vspace{-1em}
    \caption{Varying the number of slots the model is trained with on \emph{COCO object discovery} (mean $\pm$ standard dev., 3 seeds), using a ViT-B/16 encoder and the Transformer decoder.
    Our method shows a small trade-off between instance- (more slots) and class-level segmentation (less slots).
    Slot attention fails to produce meaningful masks over a range of slots.}
    \label{app:fig:coco-discovery-num-slots}
\end{figure}

\begin{figure}[tbh]
    \centering
    \includegraphics[width=\textwidth]{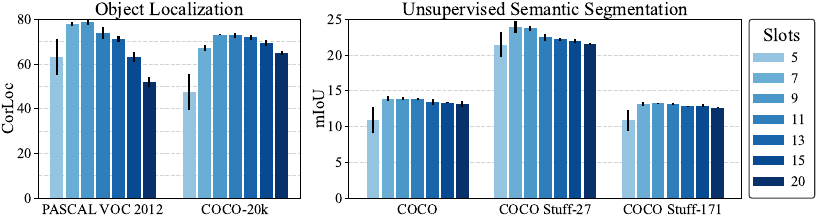}
    \vspace{-1em}
    \caption{Varying the number of slots the model is trained with on \emph{object localization} and \emph{unsupervised semantic segmentation} (mean $\pm$ standard dev., 3 seeds), using a ViT-B/16 encoder and the Transformer decoder.
    The models are trained on COCO and tested on the respective datasets.
    }
    \label{app:fig:coco-loc-segm-num-slots}
\end{figure}

\looseness-1
\textbf{PASCAL (\tab{app:tab:results-pascal-num-slots}) \& COCO (\fig{app:fig:coco-discovery-num-slots}, \fig{app:fig:coco-loc-segm-num-slots})\quad}
On both PASCAL VOC 2012 and COCO, our method performs well over a range of slots, as long as a minimum number of slots is given.
Slot Attention, in contrast, fails to produce meaningful results for all tested number of slots.
For our method on COCO, the analysis shows that there are different sweet spots for instance-level (9--11 slots) and class-level segmentation (7 slots).
Visual inspection (see \fig{app:fig:examples-coco-num-slots}) suggests that the optimal number of slots (unsurprisingly) depends on the complexity of the specific image: using few slots tends to lead to grouping of related objects, whereas using many slots leads to oversegmentation of objects into parts.
Choosing the number of slots per-dataset rather than per-image thus only optimizes results for the average scene.
An interesting question for future work is how models could deal with the variance of visual complexity and number of instances that they are faced with in real-world situations.

\begin{figure}[th]
    \begin{minipage}[t]{0.52\textwidth}
        \centering
        \captionof{table}{Number of slots analysis on \emph{MOVi-E object discovery} (mean over 3 seeds). All methods perform better when training with less slots.}
        \label{app:tab:results-movi-e-num-slots}
        \setlength{\tabcolsep}{0.6em}
        \begin{tabular}{@{}lcccccc@{}}
        \toprule
        & \multicolumn{2}{c}{\textbf{11 slots}} & {} & \multicolumn{2}{c}{\textbf{24 slots}} \\
        \cmidrule{2-3} \cmidrule{5-6}
        & ARI & mBO & \phantom & ARI & mBO \\
        \midrule
        \emph{Block Masks} & \emph{48.9} & \emph{14.7} && \emph{45.4} & \emph{24.3} \\
        \emph{DINO K-Means} & \emph{50.1} & \emph{35.2} && \emph{50.2} & \emph{23.6} \\
        Slot Attention & \meanandstd[m]{52.61}{2.60} & \meanandstd[m]{16.25}{1.10} && \meanandstd[m]{44.98}{1.67} & \meanandstd[m]{23.98}{1.20} \\
        SLATE & \meanandstd[m]{49.35}{0.43} & \meanandstd[m]{16.59}{0.32} && \meanandstd[m]{44.36}{0.86} & \meanandstd[m]{23.60}{0.54} \\
        Ours (ViT-S/8) & \meanandstd[m]{76.69}{0.75} & \meanandstd[m]{29.70}{0.10} && \meanandstd[m]{64.66}{0.72} & \meanandstd[m]{34.10}{0.12} \\
        Ours (ViT-B/8) & \meanandstd[m]{79.28}{0.52} & \meanandstd[m]{32.68}{0.14} && \meanandstd[m]{65.13}{1.23} & \meanandstd[m]{35.52}{0.17} \\
        \bottomrule
        \end{tabular}
    \end{minipage}\hfill
    \begin{minipage}[t]{0.44\textwidth}
        \centering
        \captionof{table}{Number of slots analysis on \emph{PASCAL VOC 2012 object discovery} (mean $\pm$ standard dev., 3 seeds), using a ViT-B/16 encoder and the Transformer decoder.}
        \label{app:tab:results-pascal-num-slots}
        \begin{tabular}{@{}lccc@{}}
        \toprule
        Slots & FG-ARI & \mBOinst & \mBOclass \\
        \midrule
        4 & \meanandstd{29.59}{1.43} & \meanandstd{34.22}{0.97} & \meanandstd{40.36}{1.13} \\
        5 & \meanandstd{26.61}{1.13} & \meanandstd{39.12}{0.60} & \meanandstd{45.94}{0.66} \\
        6 & \meanandstd{24.84}{2.21} & \meanandstd{44.02}{1.85} & \meanandstd{51.18}{1.87} \\
        7 & \meanandstd{19.06}{1.31} & \meanandstd{42.73}{0.72} & \meanandstd{47.88}{0.55} \\
        9 & \meanandstd{20.53}{0.69} & \meanandstd{43.47}{0.28} & \meanandstd{47.39}{0.35} \\
        \bottomrule
        \end{tabular}
    \end{minipage}
\end{figure}

\subsubsection{Test-Time Robustness}
In the last section, we found that the quality of results is affected by the number of slots the model is trained with.
An attractive quality of Slot Attention-based models\footnote{This is only true when using randomly sampled, and not fixed, learned slot initializations.
Recent models~\citep{kipf2022conditional,elsayed2022savi++} depart from the original random sampling design, as it appears to be easier to train the model with fixed initializations on more complex data.
We found no such difficulties with \dinosaur.
} is that the number of slots \emph{can be changed after the model is trained}~\citep{Locatello2020SlotAttention}, and we inherit this feature.
However, this constitutes a form of distribution shift for the model, and so it is important to know how robust the model is to such changes.
To this end, we compare models on trained on a certain number of slots, with models trained on a \emph{different} number of slots, but tested with that number of slots.
If the performance is similar, we have evidence for the robustness of the model to this type of shift.

\Tab{app:fig:analysis-num-slots-transfer} shows the results of this experiment on MOVi-E, PASCAL VOC 2012, and COCO.
We find that the model is fairly robust to changing the number of slots at test time.
Interestingly, in many instances, there is even a large improvement from changing the number of slots (\eg on MOVi-E from 11 to 24 slots, or on PASCAL, from 6 to 4 slots.
This can be explained by our observations in \sec{app:subsubsec:slots-training-sensitivity}: depending on the number of slots at training time, the model learns to focus on object of different scales.
Some object scales fit better to the dataset annotations than others.
Applying a better model with a different number of slots can thus improve over models that did not perform well with the original slot configuration.
We can also explain the cases where performance drops from changing the number of slots (\eg on MOVi-E from 24 to 11 slots, or on COCO from 20 slots) with this: those configurations did not work well in the first place, so they also do not transfer well to a different number of slots.
Overall, this result shows that \dinosaur generalizes well when applied with a different number of slots.

\begin{figure}[tbh]
    \centering
    \includegraphics[width=\textwidth]{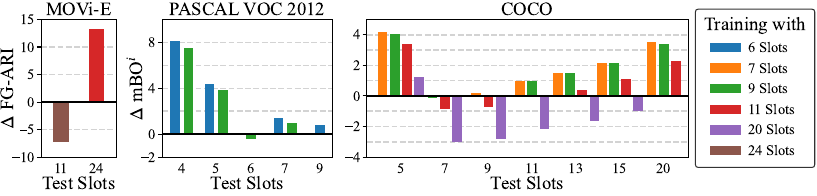}
    \vspace{-1em}
    \caption{Varying the \emph{test time} number of slots at on \emph{object discovery} (mean $\pm$ standard dev., 3 seeds).
    We plot the \emph{difference} between testing at a certain number of slots and directly training at that number of slots, for models trained with different number of slots.
    We list FG-ARI for MOVi-E, and \mBOinst for COCO and PASCAL VOC 2012.
    For MOVi-E, we use a ViT-B/8 encoder with the MLP decoder.
    For COCO and PASCAL VOC 2012, we use a ViT-B/16 encoder with the Transformer decoder.
    The performance of the original models can be gauged from \tab{app:tab:results-movi-e-num-slots}, \tab{app:tab:results-pascal-num-slots} and \fig{app:fig:coco-discovery-num-slots}.
    }
    \label{app:fig:analysis-num-slots-transfer}
\end{figure}

\subsection{How Does Decoder Scale Interact With Object Discovery?} \label{app:subsec:exp-decoder-analysis}

\looseness-1
Extending our decoder analysis from \sec{subsec:analysis}, we investigate how the capacity of the decoder and the slot dimensionality affects the object discovery performance on COCO.
In particular, we compare the MLP with the Transformer decoder:
for the MLP decoder, we use a 4-layer MLP and test hidden layer dimensionalities 1024, 2048, and 4096.
For the Transformer decoder, we test different layer numbers 2, 3, and 4.
For each decoder setting, we test it under slot dimensionalities 64, 128, 256, and 512.

\begin{figure}[tb]
    \centering
    \includegraphics[width=\textwidth]{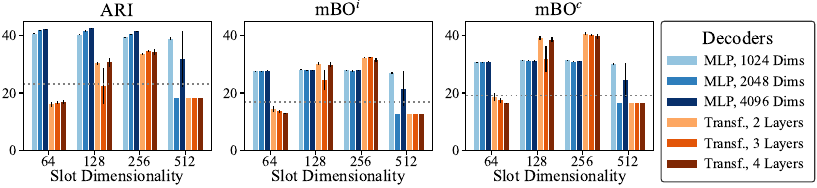}
    \vspace{-1em}
    \caption{\emph{COCO object discovery} performance when varying decoder capacity and slot dimensionality, using 7 slots (mean $\pm$ standard dev., 3 seeds).
    For the MLP decoder, we vary the dimensionality of a 4-layer MLP, whereas for the Transformer decoder, we vary the number of layers.
    The dotted line represents the block masks baseline.
    The analysis reveals instabilities of the Transformer decoder: for too small or too large slot dimensionalities, the performance degrades below the block masks baseline.
    The MLP decoder is generally robust, but also becomes unstable at slot dimensionality 512.
    }
    \label{app:fig:coco-discovery-decoders}
\end{figure}

\looseness-1
We show the results in \fig{app:fig:coco-discovery-decoders}.
The general trend of \tab{tab:analysis-decoders} stays unchanged: the MLP decoder has higher ARI (indicating better instance grouping), whereas the Transformer decoder has higher mBO (indicating better mask quality).
For the MLP decoder, we find that scaling the MLP decoder has slight benefits in terms of ARI, and that it is robust to the slot dimensionality (with the exception of slot dimensionality 512).
For the Transformer decoder, scaling appears to have little impact.
However, we find that the Transformer decoder is sensitive to the slot dimensionality:
setting it too low (64, 128) can cause the model to collapse to just use one or two slots to model the whole image.
Setting it too high (512) causes training instabilities (\ie NaNs or loss spikes).
These instabilities under slot dimensionality 512 also occur with the MLP decoder.
We conjecture this might be related to the use of 16-bit mixed precision and the attention mechanism of slot attention, but we did not investigate further.

\subsection{The Transformer Decoder Performs Poor with Larger ViTs} \label{app:subsec:exp-transformer-decoder}
In \tab{app:tab:results-transformer-dec-analysis}, we compare the Transformer decoder with different ViT encoders on COCO object discovery.
We find that scaling up the encoder leads to better results.
However, specifically combining the Transformer decoder with the ViT-B/8 encoder leads to poor results -- the MLP decoder has no such issues (\cf the results on the MOVi datasets in \fig{fig:results-object-discovery-synthetic-quant}).
Visual inspection shows that the model collapses to use just one or two slots for the whole image.
We attribute this to the \emph{high capacity of the decoder}, which leads to it being able to reconstruct the token map well while using the conditioning information from only a few slots.
Note that the capacity of the decoder is related to the encoder through the token dimensionality, and the number of tokens (which has strong influence on a Transformer's capacity as it directly influences the amount of memory and computation available throughout the blocks).
Indeed, when reducing the decoder's capacity by either reducing the feature dimensionality (ViT-S/8) or the number of tokens the decoder operates on (using bicubic interpolation on the feature map to reconstruct), the problem disappears.
In principle, the issue could also arise from optimization becoming unstable due to switching to a larger model, but the loss curves do not suggest such instabilities.

\begin{table*}[tbh]
    \centering
    \caption{Comparing different encoders with the Transformer decoder on \emph{COCO object discovery}.
    For ViT-B/8, we also list results for \emph{scaling the target token map} by a factor of $0.5$ (from 784 to 196 tokens) and by a factor of $0.75$ (to 441 tokens).
    Object-centricness does no longer emerge for ViT-B/8, which we attribute to the high capacity of the decoder: when reducing the capacity (by reducing the feature dimensionality or the number of tokens), the problems disappear.
    }
    \label{app:tab:results-transformer-dec-analysis}
    \begin{tabular}{@{}lccccc@{}}
    \toprule
    Encoder & Target Tokens & $D_\text{feat}$ & FG-ARI & \mBOinst & \mBOclass \\
    \midrule
    ViT-S/16 & 196 & 384 & \meanandstd{36.87}{0.61} & \meanandstd{29.68}{0.65} & \meanandstd{33.85}{0.97} \\
    ViT-B/16 & 196 & 768 & \meanandstd{34.14}{1.03} & \meanandstd{31.63}{0.69} & \meanandstd{39.72}{0.94} \\
    ViT-S/8 & 784 & 384 & \meanandstd{34.32}{0.49} & \meanandstd{32.29}{0.35} & \meanandstd{38.83}{0.38} \\
    ViT-B/8 & 196 (scaled) & 768 & \meanandstd{35.84}{0.65} & \meanandstd{32.03}{0.61} & \meanandstd{38.55}{0.20} \\
    ViT-B/8 & 441 (scaled) & 768 & \meanandstd{20.06}{9.79} & \meanandstd{19.06}{8.56} & \meanandstd{23.54}{9.91} \\
    ViT-B/8 & 784 & 768 & \meanandstd{13.07}{3.46} & \meanandstd{16.58}{3.90} & \meanandstd{21.25}{5.14} \\
    \bottomrule
    \end{tabular}

\end{table*}

\subsection{Mask Type Analysis} \label{app:subsec:exp-mask-type}
As noted in \sec{sec:method}, the Transformer decoder offers a choice between attention masks from the slot attention module, and attention masks from the decoder.
In \tab{app:tab:results-transformer-mask-type}, we compare the two types of masks and find that decoder attention masks are consistently better on all tasks, and especially outperform the masks from slot attention on segmentation.
Note that we could in principle also use masks from slot attention with the MLP decoder.
We point to~\citet{Singh2022STEVE}, who analyze this choice and find that alpha masks produced by reconstruction are superior to attention masks.

\begin{table*}[th]
    \centering
    \caption{Comparing masks from Slot Attention and the Transformer decoder on \emph{COCO} (mean $\pm$ standard dev., 5 seeds), using a ViT-B/16 encoder. For segmentation, we list results for object segmentation (COCO) and scene decomposition (Stuff-27, Stuff-171).}
    \label{app:tab:results-transformer-mask-type}
    \setlength{\tabcolsep}{0.25em}
    \begin{tabular}{@{}lcccccccccc@{}}
    \toprule
    & \multicolumn{3}{c}{\textbf{Object Discovery}} & {} & \multicolumn{2}{c}{\textbf{Object Localization}} & {} & \multicolumn{3}{c}{\textbf{Segmentation (mIoU)}} \\
    \cmidrule{2-4} \cmidrule{6-7} \cmidrule{9-11}
    Mask Type & FG-ARI & \mBOinst & \mBOclass & \phantom & CorLoc & DetRate & \phantom & COCO & Stuff-27 & Stuff-171 \\
    \midrule
    Slot Attention & \meanandstd{31.72}{1.38} & \meanandstd{30.27}{0.79} & \meanandstd{38.08}{1.14} && \meanandstd{65.99}{1.48} & \meanandstd{30.19}{0.81} && \meanandstd{12.02}{0.17} & \meanandstd{21.09}{0.58} & \meanandstd{11.68}{0.18} \\
    Decoder & \meanandstd{34.14}{1.03} & \meanandstd{31.63}{0.69}
 & \meanandstd{39.72}{0.94} && \meanandstd{67.19}{1.45} & \meanandstd{31.06}{0.85} && \meanandstd{13.92}{0.40} & \meanandstd{23.99}{0.85} & \meanandstd{13.13}{0.26} \\
    \bottomrule
    \end{tabular}

\end{table*}

\subsection{Does DINOSAUR Perform Semantic Grouping on MOVi?}
\label{app:subsec:movi-semantic}
On the MOVi datasets, \dinosaur seemingly separates objects into slots, \ie instance grouping.
As \dinosaur exhibits some bias towards semantic grouping on real-world datasets, an interesting question is whether our approach does in fact also perform semantic grouping on MOVi.
If this would be the case, it would be hard to detect by evaluating on the original MOVi datasets, because objects of the same category rarely appear on the same image on those datasets.

To answer this question, we designed two variants of the MOVi-C dataset.
In the first, only objects of \emph{the same semantic category} appear in each image.
In the second, each image contains \emph{the exact same object} several times.
All other characteristics of the dataset stay unchanged.
If our method would perform semantic grouping on MOVi, objects would be grouped together on those datasets.
In \tab{app:tab:movi-semantic}, we list the results for training on MOVi-C and testing on the new ``semantic'' variants.
We find that the results do not change significantly compared to evaluating on the original MOVi-C dataset, indicating that our method does in fact perform instance grouping on the MOVi datasets.

\begin{table*}[tbh]
    \centering
    \caption{Testing a model trained on MOVi-C on two variants of MOVi-C: in ``semantic objects'', each image only contains objects of the same semantic category; in ``duplicate objects'', each image only contains the same object, but several times. We show results for our method with ViT-B/8 encoder and MLP decoder, with mean $\pm$ standard dev. over 5 seeds.
    We find that the results do not change significantly, indicating that \dinosaur does not apply semantic grouping on MOVi.}
    \label{app:tab:movi-semantic}
    \begin{tabular}{@{}lcc@{}}
    \toprule
    Dataset & FG-ARI & mBO \\
    \midrule
    MOVi-C & \meanandstd{68.63}{0.44} & \meanandstd{40.78}{0.20} \\
    Semantic Objects & \meanandstd{68.63}{0.44} & \meanandstd{39.07}{0.19} \\
    Duplicate Objects & \meanandstd{68.10}{0.23} & \meanandstd{38.95}{0.17} \\
    \bottomrule
    \end{tabular}
    {%
    \newcommand{\myig}[1]{\includegraphics[width=0.12\textwidth,valign=c]{#1}}
    \newcommand{\mycaption}[1]{{\rotatebox[origin=c]{90}{\footnotesize#1}}}
    \renewcommand{\arraystretch}{3.9}
    \setlength{\tabcolsep}{1pt}
    \begin{tabular}{@{}rcccc@{}}
        \mycaption{Semantic} & \myig{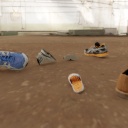} & \myig{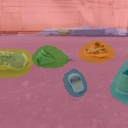} & \myig{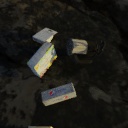} & \myig{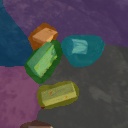}  \\
        \mycaption{Duplicate} & \myig{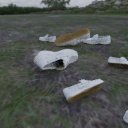} & \myig{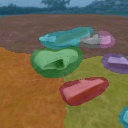} & \myig{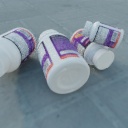} & \myig{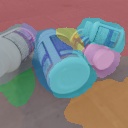} \\
    \end{tabular}}
\end{table*}

\section{Implementation Details}
\label{app:sec-implementation-details}

\subsection{Architecture and Hyperparameters}
\label{app:subsec-detail-architecture}
In the following, we describe implementation details of the \dinosaur architecture.
The hyperparameters for our experiments are given in \tab{app:tab:dinosaur_params}.

\textbf{ViT Encoder\quad}
We use the Vision Transformer implementation and provided by the timm library~\citep{Wightman2019timm}\footnote{\url{https://github.com/rwightman/pytorch-image-models}, v0.6.7}.
Depending on the experiment, we use the following configurations:
ViT-S/16 (token dimensionality 384, 6 heads, patch size 16), ViT-S/8 (token dimensionality 384, 6 heads, patch size 8), ViT-B/16 (token dimensionality 768, 12 heads, patch size 16), ViT-B/8 (token dimensionality 768, 12 heads, patch size 8).
The specific timm model names (for DINO weights) used are \texttt{vit\_small\_patch16\_224\_dino} (ViT-S/16), \texttt{vit\_small\_patch8\_224\_dino} (ViT-S/8), \texttt{vit\_base\_patch16\_224\_dino} (ViT-B/16), \texttt{vit\_base\_patch8\_224\_dino} (ViT-B/8).
All models use 12 Transformer blocks, linear patch embedding and additive positional encoding.
The output of the last block (not applying the final layer norm) is passed on to the Slot Attention module and used in the feature reconstruction loss, after removing the entry corresponding to the CLS token.
We always use images of size $224 \times 224$ as input to the ViT; this yields $N=14^2=196$ patches for patch size 16 and $N=28^2=784$ patches for patch size 8.
If the image size is not equal to $224$, we use bicubic interpolation for resizing.
For the pre-trained weights, we use the timm library for DINO and third-party releases for MoCo-v3\footnote{\url{https://github.com/facebookresearch/moco-v3}}, MSN\footnote{\url{https://github.com/facebookresearch/msn}}, and MAE\footnote{\url{https://github.com/facebookresearch/mae}}.

\textbf{ResNet34 Encoder\quad}
We use a ResNet34 using the modifications suggested by~\citet{kipf2022conditional} and \citet{elsayed2022savi++}.
That is, we modify the basic ResNet34 architecture (\texttt{resnet34} in the timm library) to use a stride of $1$ in the first convolution, and replace batch normalization with group normalization.
This results in a $512$-dimensional feature map of size $28 \times 28$ for images of size $224 \times 224$.
Like \citet{kipf2022conditional}, we add a linear positional encoding to the feature map passed to the Slot Attention module (coordinates in the four cardinal directions scaled to $[-1, 1]$, linearly projected to the feature dimension and added to the features).

\textbf{Slot Attention\quad}
The Slot Attention module largely follows the original formulation from \citet{Locatello2020SlotAttention}.
After applying a layer norm, the set of $N$ input features is transformed to slot dimensionality $D_\text{slots}$ by a two-layer MLP with hidden size equal to the feature size $D_\text{feat}$, followed by another layer norm.
An initial set of $K$ slots is randomly sampled from a normal distribution parametrized by learnable mean $\mu \in \mathbb{R}^{D_\text{slots}}$ and log standard deviation $\log\sigma \in \mathbb{R}^{D_\text{slots}}$.
The slots are then updated over several iterations, where each iteration involves a layer norm step on the current slots, an attention step where slots compete for input features, updating the slots using a GRU~\citep{Cho2014GRU}, and applying a residual two-layer MLP.
The query/key/value projections' size is the same as the slot dimensionality $D_\text{slots}$; they do not use a bias.
The key-value dot product is scaled by $D_\text{slots}^{-0.5}$, and an epsilon of $10^{-8}$ is added to the softmax weights for stability reasons.
The MLP has a hidden size equal to $4 \cdot D_\text{slots}$.

On MOVi-C and MOVi-E, we use 11 and 24 slots respectively, which corresponds to the maximal number of objects that can appear in the scene plus one slot that can capture the background.
On PASCAL VOC and COCO, we use 6 and 7 slots respectively, as we found those values to perform quantitatively well and give visually appealing results (\cf \app{app:subsec:exp-number-of-slots}).

\textbf{MLP Decoder\quad}
We use a four-layer MLP with ReLU activations, with output dimensionality $D_\text{feat} + 1$, where the last dimension is used for the alpha value.
The MLP has hidden layer sizes of 1024 for the MOVi datasets, and 2048 for COCO.
A learnable positional encoding of size $N \times D_\text{slots}$ is added to the slots after broadcasting them to the number of patches $N$.

\looseness-1
\textbf{Transformer Decoder\quad}
We use the standard decoder design from~\citet{Vaswani2017Attention} with pre-normalization~\citep{Xiong2020OnLN}, conditioning on the set of slots output by the slot attention module.
Similar to ImageGPT~\citep{Chen2020ImageGPT}, the decoder autoregressively reconstructs $N$ patch features of dimensionality $D_\text{feat}$ starting from the top-left and proceeding row-by-row over the image (``raster order'').
To this end, its initial input is the set of target features shifted back by one, removing the last feature and inserting a learned ``beginning-of-sentence'' feature at the start.
Each Transformer block then consists of self-attention on the set of tokens (using causal masking to prevent attending to future tokens), cross-attention with the set of slots, and a residual two-layer MLP with hidden size $4 \cdot D_\text{feat}$.
Before the Transformer blocks, both the initial input and the slots are linearly transformed to $D_\text{feat}$, followed by a layer norm.
The Transformer thus operates on tokens of dimensionality $D_\text{feat}$.
The output of the last Transformer block is directly used as the reconstruction $\vy$.
We do not apply dropout as we did not find it to be beneficial.
Also, we do not add positional encoding as the target features forming the input already contain positional information (from the ViT encoder).

\textbf{Masks for Evaluation\quad}
As mentioned in the main text, we use the attention mask from the decoder as the slot mask used for evaluation.
In particular, the cross attention step involves a softmax over the slots for each of the $N$ tokens, resulting in $N$ attention weights per slot, which can be reshaped to form a $\sqrt{N} \times \sqrt{N}$ mask.
Intuitively, each mask entry measures how important the corresponding slot was for reconstructing the patch feature at that position.
For the evaluation, this mask is bilinearly resized to the image size.
We use the attention mask from the last decoder block, which we found to perform best against the alternatives of using masks from earlier blocks or averaging the masks across blocks.
The soft slot masks are converted to hard masks using the $\argmax$.

\textbf{Details for Specific Experiments\quad}
Here we list settings for experiments aside from our main experiments.

\begin{itemize}[topsep=0pt]
    \item Insufficiency of Image Reconstruction (\sec{subsec:analysis}). In this experiment, we train on COCO using image reconstruction and combine a ViT-B/16 encoder with Slot Attention's spatial broadcast decoder (as detailed in \app{app:subsec:details-baselines}).
    Image reconstruction uses $128 \times 128$ as the target resolution.
    For finetuning, we use a lower learning rate of $4\cdot 10^{-5}$ for the encoder, compared to $4\cdot 10^{-4}$ for Slot Attention and the decoder.
    All models use 7 slots.
    Other training and architecture settings are the same as for the main experiments.
    \item \looseness-1 Choice of Self-Supervised Method (\sec{subsec:analysis}) and Choice of Pre-Training Method (\sec{app:subsec:choice-pretraining-method}). In these experiments, we use different pre-trained networks for feature extraction and reconstruction. The specific timm model names used are \texttt{resnet50} (ResNet50, supervised), \texttt{vit\_small\_patch16\_224} (ViT-S/16, supervised), \texttt{vit\_base\_patch16\_224} (ViT-B/16, supervised), \texttt{vit\_small\_patch16\_224\_dino} (ViT-S/16, DINO), \texttt{vit\_base\_patch16\_224\_dino} (ViT-B/16, DINO).
    For MoCo-v3, MSN and MAE, we use the ViT-B/16 implementation of timm with third-party weight releases (referenced above).
    All networks were pre-trained on the ImageNet dataset.
    The ResNet50 encoder uses the features output from the second last block (feature level 3 in timm), resulting in a $1024$-dimensional feature map of size $14 \times 14$.
    Like~\citet{kipf2022conditional}, we add linear positional encoding to this feature map before Slot Attention.
    The ViT encoders use the Transformer decoder whereas for the ResNet encoder, we use the MLP decoder as training with the Transformer decoder did not yield good results.
    All models use 7 slots.
    \item Choice of Decoder (\sec{subsec:analysis}). This experiment uses 11 slots for MOVi-C, 6 slots for PASCAL VOC 2012, and 7 slots for COCO.
    \item Object Property Prediction (\sec{app:subsec:coco-property-prediction}). We train the downstream models for \numprint{15000} steps with the Adam optimizer with a starting learning rate of $10^{-3}$ and batch size $64$.
    The learning rate is halved every \numprint{2000} training steps.
    The non-linear predictor is a one-hidden layer MLP with 256 hidden neurons and leaky ReLU activation (slope $0.01$).
    Continuous targets are standardized by computing mean and variance over the training dataset.
    We use center crops as input and filter objects that are not contained in the center crop.
    \item Comparison of ResNet and Pre-Trained ViT Encoders (\sec{app:subsec:training-encoder-scratch}). We train the ResNet models for 300k steps (instead of 500k steps as the models in our main experiments) due to computational constraints, but we found that they can also benefit from longer training.
\end{itemize}

\begin{table}[th!]
\caption{Hyperparameters for \dinosaur used for the results on MOVi-C, MOVi-E, PASCAL VOC 2012, COCO and KITTI datasets.}
\label{app:tab:dinosaur_params}
\small
\centering
\setlength{\tabcolsep}{0.47em}
\begin{tabular}{@{}llrrrrrr@{}}
\toprule
Dataset                            &                           & \textbf{MOVi-C}  & \textbf{MOVi-E}    & \textbf{PASCAL} &  \textbf{COCO} & \textbf{KITTI}  \\ \midrule
Training Steps                     &                           & 500k       & 500k       & 250k         & 500k   & 500k   \\
Batch Size                         &                           & 64         & 64         & 64           & 64     & 64           \\
LR Warmup Steps                    &                           & 10000      & 10000      & 10000        & 10000  & 10000     \\
Peak LR                            &                           & 0.0004     & 0.0004     & 0.0004       & 0.0004 & 0.0004           \\
Exp.~Decay Half-Life               & & 100k       & 100k       & 100k       & 100k     & 100k     \\
ViT Architecture                   & & ViT-B      & ViT-B      & ViT-B        & ViT-B  & ViT-B           \\
Patch Size                         &                           & 8          & 8          & 16           & 16    & 8          \\
Feature Dim. $D_\text{feat}$       &       & 768       & 768        & 768         & 768  & 768        \\
Gradient Norm Clipping                                                                              &                          & 1.0        & 1.0        & 1.0          & 1.0     & 1.0          \\ \midrule
Image/Crop Size                                                                                     &                           & 224        & 224        & 224 & 224  & 224 \\
Cropping Strategy                                                                              &                           & Full         & Full         & Random       & Center  & Random           \\
Augmentations                                                                             &                           & --         & --         & \multicolumn{3}{c}{\hspace{2em}Random Horizontal Flip}              \\
Image Tokens                                                                                   &                           & 784      & 784       & 196  & 196   & 784  \\ \midrule
\multirow{4}{*}{Decoder}            & Type                      & MLP          & MLP      & Transformer  & Transformer    & MLP   \\
& Layers                    & 4          & 4          & 4            & 4       & 4 \\
& Heads                     & --        & --          & 8            & 8       & --  \\
& MLP Hidden Dim.               & 1024       & 1024        & 3072         & 3072     & 2048         \\    \midrule
\multirow{4}{*}{Slot Attention}      & Slots                     & 11         & 24         & 6            & 7  & 9     \\
& Iterations                & 3          & 3          & 3           & 3         & 3        \\
& Slot Dim.~$D_\text{slots}$          & 128        & 128        & 256          & 256   & 32  \\
& MLP Hidden Dim.          & 512        & 512        & 1024          & 1024    & 128   \\\midrule
\end{tabular}
\end{table}

\newpage
\subsection{Baselines}
\label{app:subsec:details-baselines}

\begin{wrapfigure}[8]{r}{0.25\textwidth}
    \centering
    \vspace{-1\intextsep}
    \includegraphics[width=0.11\textwidth]{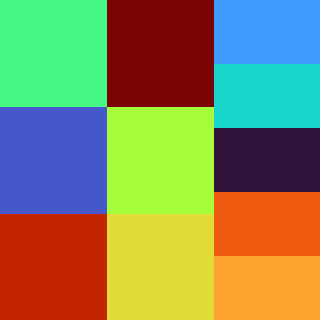}\hspace{0.5em}
    \includegraphics[width=0.11\textwidth]{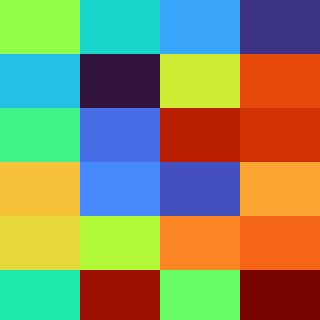}
    \vspace{-0.5em}
    \caption{Block masks for 11 and 24 slots.}
    \label{app:fig:block_patterns}
\end{wrapfigure}
\paragraph{Block Masks}
The block mask patterns are generated by equally dividing the image into a specified number of columns, then equally dividing the columns such that a target number of slot masks is reached (see \fig{app:fig:block_patterns}).
For less than 9 masks, 2 columns are used, for 9--15 masks, 3 columns, and for more than 15 masks, 4 columns are used.
For the block masks results given in the main text, MOVi-C uses 11 masks, MOVi-E uses 24 masks, PASCAL VOC 2012 uses 6 masks and COCO uses 7 masks.
The number of masks is consistent with the number of slots used by the other methods on the respective datasets.

\paragraph{DINO K-Means}
We run the K-Means algorithm on the patch tokens output by the last DINO ViT block (before the final layer normalization).
For the MOVi experiments, we use ViT-B/8, whereas for the PASCAL VOC 2012 and COCO experiments, we use ViT-B/16.
The resulting low-dimensional clustering masks are resized to the target mask resolution required for evaluation, where we treat each cluster as an object instance.
For K-Means, we use the standard scikit-learn implementation (\textit{sklearn.cluster.KMeans}) with default parameters (10 repetitions, k-means++ init, maximum of 300 iterations).
We only vary the number of clusters: for the results given in the main text, MOVi-C uses 11 clusters, MOVi-E uses 24 clusters, PASCAL VOC 2012 uses 6 clusters and COCO uses 7 clusters.
The number of clusters is consistent with the number of slots used by the other methods on the respective datasets.

\paragraph{Slot Attention} \looseness-1
The hyperparameters used to run the experiments can be found in \tab{app:tab:slot_attention_params}.
For the encoder, we use the ResNet34 as described above.
The slot attention module largely follows the description given in \app{app:subsec-detail-architecture} for the \dinosaur architecture.
However, like \citet{kipf2022conditional}, we add a linear positional encoding to the feature map (coordinates in the four cardinal directions scaled to $[-1, 1]$, linearly projected to the feature dimension and added to the features).
The number of slots was set to 11 and 24 for MOVi-C and MOVi-E, whereas for PASCAL and COCO we used 6 and 7 slots, to be consistent with \dinosaur.
For PASCAL and COCO, we verified that Slot Attention does not succeeded in successfully discovering objects over a range of number of slots (\cf \fig{app:fig:coco-discovery-num-slots}).
For the decoder, we use the same spatial broadcast decoder as \citep{kipf2022conditional}, \ie 5 transposed convolutions with $5 \times 5$ kernels, channel size $64$ and ReLU activations, upsampling from size $8 \times 8$ to $128 \times 128$.
Input and target images are scaled to range $[-1, 1]$.

\begin{table}[h!]
\caption{Hyperparameters used for Slot Attention training on MOVi-C, MOVi-E, PASCAL VOC 2012 and COCO datasets.}
\label{app:tab:slot_attention_params}
\small
\centering
\setlength{\tabcolsep}{0.45em}
\begin{tabular}{@{}llrrrr@{}}
\toprule
Dataset                                                                                        &                           & \textbf{MOVi-C}  & \textbf{MOVi-E}    & \textbf{PASCAL VOC 2012} \&  \textbf{COCO}     \\ \midrule
Training Steps                                                                                 &                           & 250k       & 250k       & 250k    \\
Batch Size                                                                                     &                           & 64         & 64         & 64                  \\
LR Warmup Steps                                                                                &                           & 2500      & 2500      & 2500          \\
Peak LR                                                                                        &                           & 0.0002     & 0.0002     & 0.0002           \\
Annealing                                                                                      &                           & Cosine     & Cosine      & Cosine            \\
Gradient Norm Clipping                                                                              &                           & 1.0        & 1.0        & 1.0               \\ \midrule
Image/Crop Size                                                                                     &                           & 128         & 128         & 128                           \\
Cropping Strategy                                                                              &                           & Full         & Full         & Random (PASCAL) / Center (COCO)             \\
Augmentations                                                                             &                           & --         & --         & Random Horizontal Flip             \\
\midrule
\multirow{4}{*}{Slot Attention}      & Slots                     & 11         & 24         & 6 \& 7      \\
                                                                                               & Iterations                & 3          & 3          & 3                 \\
                                                                                               & Slot Dim.                 & 256        & 256         & 256               \\
                                                                                               & MLP Hidden Dim.                 & 512        & 512         & 512               \\ \midrule
\end{tabular}
\end{table}

\paragraph{SLATE}
We use the official SLATE implementation\footnote{\url{https://github.com/singhgautam/slate}}.
The hyperparameters are close to the ones used in SLATE~\citep{Singh2022SLATE} and STEVE~\citep{Singh2022STEVE} and are presented in~\tab{app:tab:slate_params}.
We noticed that SLATE's performance metrics began to degrade significantly after some training time; we picked the training time such that best performance was obtained.

\begin{table}[h!]
\caption{Hyperparameters used for SLATE training on MOVi-C, MOVi-E, PASCAL VOC 2012 and COCO datasets.}
\label{app:tab:slate_params}
\small
\centering
\begin{tabular}{@{}llrrrr@{}}
\toprule
Dataset                                                                                        &                           & \textbf{MOVi-C}  & \textbf{MOVi-E}    & \makecell{\textbf{PASCAL VOC 2012} \\\& \textbf{COCO}}     \\ \midrule
Training Steps                                                                                 &                           & 40k       & 40k       & 50k \& 40k   \\
Batch Size                                                                                     &                           & 64         & 64         & 64                  \\
LR Warmup Steps                                                                                &                           & 30000      & 30000      & 30000          \\
Peak LR                                                                                        &                           & 0.0001     & 0.0001     & 0.0001           \\
Dropout                                                                                        &                           & 0.1        & 0.1        & 0.1               \\
Gradient Norm Clipping                                                                              &                           & 1.0        & 1.0        & 1.0               \\ \midrule
\multirow{4}{*}{DVAE}                                                                          & Vocabulary Size           & 4096       & 4096       & 4096         \\
                                                                                               & Temp. Cooldown            & 1.0 to 0.1 & 1.0 to 0.1 & 1.0 to 0.1  \\
                                                                                               & Temp. Cooldown Steps      & 30000      & 30000      & 30000            \\
                                                                                               & LR                        & 0.0003     & 0.0003     & 0.0003         \\ \midrule
Image Size                                                                                     &                           & 128        & 128        & 128                \\
Image Tokens                                                                                   &                           & 1024       & 1024       & 1024   \\ \midrule
\multirow{3}{*}{Transformer Decoder} & Layers                    & 8          & 8          & 8                  \\
                                                                                               & Heads                     & 4          & 4          & 4                   \\
                                                                                               & Hidden Dim.               & 192        & 192        & 192              \\ \midrule
\multirow{3}{*}{Slot Attention}      & Slots                     & 11         & 24         & 6 \& 7      \\
                                                                                               & Iterations                & 3          & 3          & 3                 \\
                                                                                               & Slot Dim.                 & 192        & 192        & 192              \\ \midrule
\end{tabular}
\end{table}

\paragraph{STEVE}
We use the official STEVE implementation\footnote{\url{https://github.com/singhgautam/steve}}.
We use the proposed configuration for the MOVi datasets and only change the number of slots to 11 (for MOVi-C) and 24 (for MOVi-E).
We train the model for \numprint{20000} steps.

\paragraph{SAVi++}
We use the official SAVi++ implementation\footnote{\url{https://github.com/google-research/slot-attention-video/}}.
We use the proposed configuration for MOVi-E and only change it to use fixed, learned slots instead of first-frame conditioning, and train the model with batch size 32 for \numprint{30000} steps.
We train with both image reconstruction and optical flow reconstruction.

\paragraph{STEGO}
To evaluate STEGO on the COCO-Stuff 171 dataset (\sec{subsec:unsup-semantic-segmentation}), we modified the official STEGO implementation\footnote{\url{https://github.com/mhamilton723/STEGO}} to use batch k-means clustering with 172 clusters.
Other parameters were not changed from the STEGO configuration on the COCO-Stuff 27 dataset as the training data is the same for those two datasets.

For the experiment evaluating STEGO on the MOVi datasets (\sec{app:subsec:movi-stego}), we run a DINO ViT-B with patch size 8, using 22 and 27 clusters, and train for \numprint{20000} steps.
We manually tested different hyperparameters and settled on $\lambda_\text{self}=0.6$, $\lambda_\text{knn}=0.3$, $\lambda_\text{rand}=0.6$, $b_\text{self}=0.26$, $b_\text{knn}=0.21$, $b_\text{rand}=0.4$ for MOVi-C, and $\lambda_\text{self}=0.6$, $\lambda_\text{knn}=0.3$, $\lambda_\text{rand}=0.6$, $b_\text{self}=0.22$, $b_\text{knn}=0.17$, $b_\text{rand}=0.4$ for MOVi-E.
We refer to \cite{Hamilton2022Stego} for the meaning of those parameters.

\section{Dataset and Evaluation Settings}
\label{app:sec:dataset-and-eval-settings}

In this section, we provide details about the datasets, evaluation settings and metrics used in this work.
An overview over which datasets are used for which task can be found in \tab{app:tab:eval-overview}.

\begin{table*}[tbh]
    \centering
    \caption{Overview over tasks and datasets used in this work.
    For training, we only use images of the respective datasets, and no labels.
    For central crops, we first resize the mask such that the short side is 320 pixels long, then take the most centered crop of size $320 \times 320$.
    }
    \label{app:tab:eval-overview}
    {\footnotesize
    \setlength{\tabcolsep}{0.3em}
    \begin{tabular}{@{}lllp{0.25\textwidth}@{}}
    \toprule
    Dataset & Images & Description & Citation \\
    \midrule
    MOVi-C & \numprint{87633} & Train split w.~videos & \citet{Greff2021Kubric} \\
    MOVi-C validation & \numprint{6000} & Val.~split w.~instance segm.~labels & \citet{Greff2021Kubric} \\
    MOVi-E & \numprint{87741} & Train split w.~ images & \citet{Greff2021Kubric} \\
    MOVi-E validation & \numprint{6000} & Val.~split w.~instance segm.~labels & \citet{Greff2021Kubric} \\
    VOC 2012 validation & \numprint{1449} & Val.~split w.~instance segm.~labels & \citet{Everingham2012PASCALVOC} \\
    VOC 2012 trainaug & \numprint{10582} & Train split w.~instance segm.~labels & \citet{Everingham2012PASCALVOC} and \citet{Bharath2011SBD} \\
    VOC 2012 trainval & \numprint{11540} & Detection train and val.~splits & \citet{Everingham2012PASCALVOC} \\
    COCO 2017 & \numprint{118287} & Train split w.~instance segm.~labels & \citet{Lin2014COCO} \\
    COCO 2017 validation & \numprint{5000} & Val split w.~instance segm.~labels & \citet{Lin2014COCO} \\
    COCO-20k & \numprint{20000} & COCO 2014 train\& val subset w.~det.~labels & \eg \citet{Simeoni2021LOST} \\
    COCO-27 & \numprint{5000} & Segm.~labels from COCO-Stuff & \citet{Caesar2018COCOStuff}, \eg \citet{Ji2019IIC}\\
    COCO-171 & \numprint{5000} & Segm.~labels from COCO-Stuff & \citet{Caesar2018COCOStuff}, new \\
    KITTI & & Raw videos & \citet{Geiger2013KITTI} \\
    KITTI segmentation & \numprint{200} & Images w.~instance segm.~labels & \citet{Alhaija2018Kitti} \\
    \bottomrule
    \end{tabular}}

    \vspace{1em}
    {
    \small
    \begin{tabular}{@{}lllcc@{}}
    \toprule
    Task & Training Dataset & Eval Dataset & Eval Resolution & Crop Type  \\
    \midrule
    \multirow{5}{*}{Object Discovery} & MOVi-C & MOVi-C validation & $128 \times 128$ & Full \\
    & MOVi-E & MOVi-E validation & $128 \times 128$ & Full \\
    & VOC 2012 trainaug & VOC 2012 validation & $320 \times 320$ & Central \\
    & COCO 2017 & COCO 2017 validation & $320 \times 320$ & Central \\
    & KITTI & KITTI segmentation & $1242 \times 375$ & Full \\
    \midrule
    \multirow{2}{*}{Object Localization} & VOC 2012 trainaug & VOC 2012 trainval & Original & Full \\
    & COCO 2017 & COCO-20k & Original & Full \\
    \midrule
    \multirow{2}{*}{Object Segmentation} & VOC 2012 trainaug & VOC 2012 validation & Original & Full \\
    & COCO 2017 & COCO 2017 validation & Original & Full \\
    \midrule
    \multirow{2}{*}{Scene Segmentation} & COCO 2017 & COCO-27 & $320 \times 320$ & Central \\
    & COCO 2017 & COCO-171  & $320 \times 320$ & Central \\
    \bottomrule
    \end{tabular}}
\end{table*}

\textbf{MOVi\quad}As the MOVi datasets are video datasets, we convert them into an image dataset by sampling 9 random frames per clip.
This yields \numprint{87633} training images for MOVi-C and \numprint{87741} images on MOVi-E.
For evaluation, we use all frames from the 250 clips, yielding \numprint{6000} images.
Note that we use the validation split provided by the MOVi dataset for evaluation, and not the test split containing out-of-distribution objects and backgrounds.
This is in line with prior work~\citep{kipf2022conditional,elsayed2022savi++}.

For the ``semantic'' MOVi-C variants used in \sec{app:subsec:movi-semantic}, we generated 250 videos using the Kubric simulator~\citep{Greff2021Kubric}, yielding \numprint{2250} images.
For the ``semantic objects'' variant, after sampling the first object in each video, only objects of the same category were allowed to be sampled.
The categories are: ``Action Figures'', ``Bag'', ``Board Games'', ``Bottles and Cans and Cups'', ``Camera'', ``Car Seat'', ``Consumer Goods'', ``Hat'', ``Headphones'', ``Keyboard'', ``Legos'', ``Media Cases'', ``Mouse'', ``Shoe'', ``Stuffed Toys'', ``Toys''.
Objects that do not have one of those categories assigned were not considered.
For the ``duplicate objects'' variant, the exact same object was resampled multiple times.
Otherwise, the same settings as for MOVi-C were used.

\textbf{PASCAL VOC 2012\quad}
We train on the ``trainaug'' variant with \numprint{10582} images.
The trainaug variant is an unofficial split that has been used by prior work~\citep{VanGansbeke2021MaskContrast,Melaskyriazi2022DeepSpectral}.
It consists of \numprint{1464} images from the segmentation train set, and \numprint{9118} images from the SBD dataset~\citep{Bharath2011SBD}.
For object discovery and segmentation, we evaluate on the official segmentation validation set, with \numprint{1449} images.
For object localization, we evaluate on the detection ``trainval'' split (joint official training and validation splits), with \numprint{11540} images, as standard for this task~\citep{Simeoni2021LOST,Melaskyriazi2022DeepSpectral}.
For unsupervised object segmentation, we use the 20 object classes plus one background class.
For object discovery and segmentation, unlabeled pixels are ignored during evaluation.

\textbf{COCO\quad} We train on the COCO 2017 dataset with \numprint{118287} images, and evaluate on the validation set with \numprint{5000} images.
For object discovery, we use both instance and segmentation masks, converting instance masks into segmentation masks using a per-pixel $\argmax$ over classes.
Overlaps between instances are ignored during metric evaluation, and crowd instance annotations are not used.
For unsupervised semantic segmentation, we use the COCO-Stuff annotations~\citep{Caesar2018COCOStuff}, which contain the original 80 ``things'' classes from COCO, plus 91 ``stuff'' class covering background.
In this labeling, overlaps have been manually resolved.
We ignore pixels belonging to crowd instances during evaluation.
For unsupervised object segmentation, we use the 80 ``things'' classes, and treat the ``stuff'' classes as background.
For unsupervised scene decomposition, we use the COCO-Stuff 27 labeling~\citep{Ji2019IIC,Cho2021Picie}, resulting from merging the 80 ``things'' classes into 12, and the 91 ``stuff'' classes into 15 super-classes, and the COCO-Stuff 171 labeling, using all ``things'' and ``stuff'' classes.
For object localization, we evaluate on COCO-20k, with \numprint{19817} images.
COCO-20k is a subset of COCO 2014 training and validation images, filtered to remove images with only crowd instances and removing crowd annotations.

\textbf{Evaluation Settings\quad}\looseness-1
For scene decomposition on COCO-Stuff 27 and COCO-Stuff 171, we take a center crop of the image after resizing the minor axis to 320 pixels and evaluate the masks at $320 \times 320$ resolution to be consistent with prior work~\citep{Ji2019IIC,Hamilton2022Stego}.
For object discovery on PASCAL VOC 2012 and COCO, we use the same settings.
For object discovery on MOVi, we use the full image (at $128 \times 128$ resolution).
Again to be consistent with prior work, for object localization on PASCAL VOC 2012 and COCO-20k as well as object segmentation on PASCAL VOC 2012 and COCO, we evaluate the \emph{full} image after resizing to $224 \times 224$, ignoring aspect ratio.
Our model proved to be robust to the aspect ratio distortion to some degree; however, we believe that results can be further improved, \eg by taking the distortion into account during training, or by using multi-crop evaluation.
Predicted soft probability masks are converted to hard masks using the $\argmax$.

\textbf{Metrics\quad}
For \emph{foreground adjusted rand index (FG-ARI)}, we use object/instance masks and only evaluate pixels in the foreground.
We treat unlabeled pixels (\eg on PASCAL) and pixels from overlapping instance masks (\eg on COCO) as background (\ie they are not evaluated).

For \emph{mean best overlap (mBO)}, we use the object masks on MOVi, and both instance and segmentation masks on PASCAL/COCO.
Each ground truth mask (excluding the background mask) is assigned the predicted mask with the largest overlap in terms of IoU.
mBO is computed as the average IoU over the assigned mask pairs.
We do not count unlabeled pixels (\eg on PASCAL) and pixels from overlapping instance masks (\eg on COCO) when computing IoU.

\section{Additional Examples}
\label{app:sec:additional-examples}

\begin{figure}[tb]
    \centering
    {
\centering
\newcommand{\myig}[1]{\includegraphics[width=0.15\textwidth,valign=c]{#1}}
\renewcommand{\arraystretch}{4.5}
\setlength{\tabcolsep}{1.5pt}
\vspace{-2em}
\begin{tabular}{@{}ccccccc@{}}
    Image & \makecell{MLP\\Decoder} & \makecell{Transformer\\Decoder} & \phantom{\,} & Image & \makecell{MLP\\Decoder} & \makecell{Transformer\\Decoder} \\[-0.5em]
    \myig{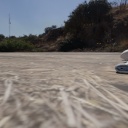} & \myig{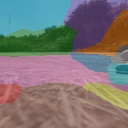} & \myig{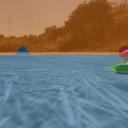} && \myig{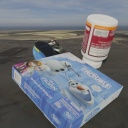} & \myig{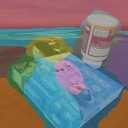} & \myig{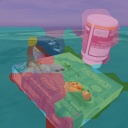} \\
    \myig{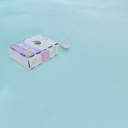} & \myig{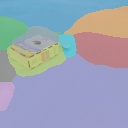} & \myig{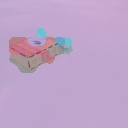} && \myig{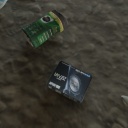} & \myig{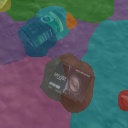} & \myig{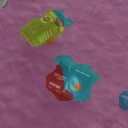} \\
    \myig{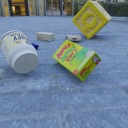} & \myig{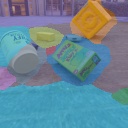} & \myig{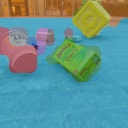} && \myig{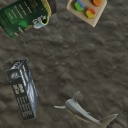} & \myig{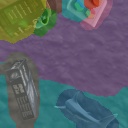} & \myig{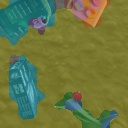} \\
    \myig{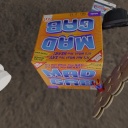} & \myig{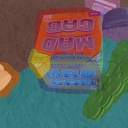} & \myig{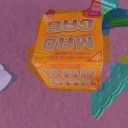} && \myig{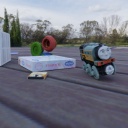} & \myig{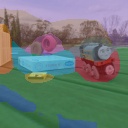} & \myig{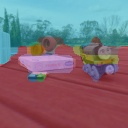} \\
\end{tabular}
}
    \caption{Masks on MOVi-C produced by our method, using 11 slots and a ViT-B/16 encoder. We show predictions from the MLP and the Transformer decoder.}
    \label{app:fig:examples-extended-movi-c}
\end{figure}

\begin{figure}[tb]
    \centering
    {
\centering
\newcommand{\myig}[1]{\includegraphics[width=0.13\textwidth,valign=c]{#1}}
\newcommand{\mycaption}[1]{{\rotatebox[origin=c]{90}{\footnotesize#1}}}
\renewcommand{\arraystretch}{3.9}
\setlength{\tabcolsep}{1pt}
\begin{tabular}{@{}rcccccc@{}}
    \mycaption{Image} & \myig{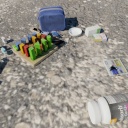} & \myig{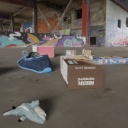} & \myig{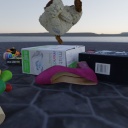} & \myig{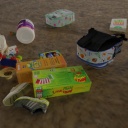} & \myig{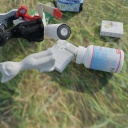} & \myig{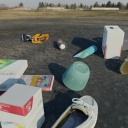} \\
    \mycaption{\dinosaur} & \myig{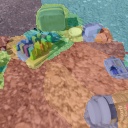} & \myig{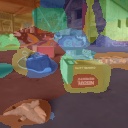} & \myig{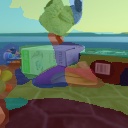} & \myig{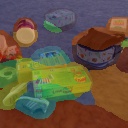} & \myig{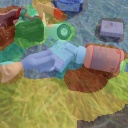} & \myig{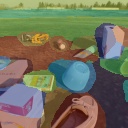} \\
    \mycaption{Slot Attention} & \myig{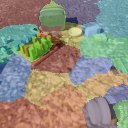} & \myig{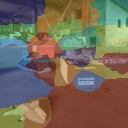} & \myig{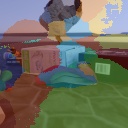} & \myig{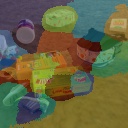} & \myig{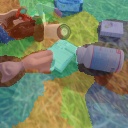} & \myig{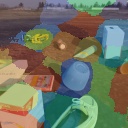} \\
    \mycaption{SLATE} & \myig{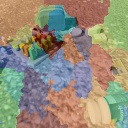} & \myig{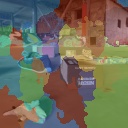} & \myig{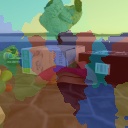} & \myig{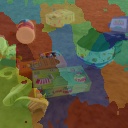} & \myig{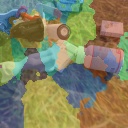} & \myig{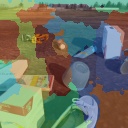} \\
\end{tabular}
}
    \caption{Masks on MOVi-E using 24 slots. Our method uses a ViT-B/8 encoder and the MLP decoder.}
    \label{app:fig:examples-extended-movi-e-24slots}
\end{figure}

\begin{figure}[tb]
    \centering
    {
\centering
\newcommand{\myig}[1]{\includegraphics[width=0.13\textwidth,valign=c]{#1}}
\newcommand{\mycaption}[1]{{\rotatebox[origin=c]{90}{\footnotesize#1}}}
\renewcommand{\arraystretch}{3.9}
\setlength{\tabcolsep}{1pt}
\begin{tabular}{@{}rcccccc@{}}
    \mycaption{Image} & \myig{images/movi_e/images/009.jpg} & \myig{images/movi_e/images/019.jpg} & \myig{images/movi_e/images/020.jpg} & \myig{images/movi_e/images/039.jpg} & \myig{images/movi_e/images/078.jpg} & \myig{images/movi_e/images/094.jpg} \\
    \mycaption{\dinosaur} & \myig{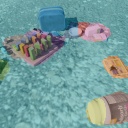} & \myig{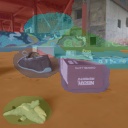} & \myig{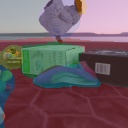} & \myig{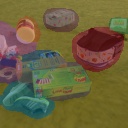} & \myig{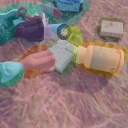} & \myig{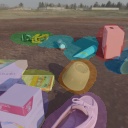} \\
    \mycaption{Slot Attention} & \myig{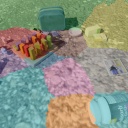} & \myig{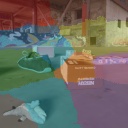} & \myig{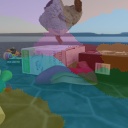} & \myig{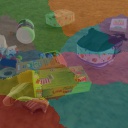} & \myig{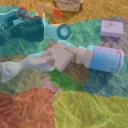} & \myig{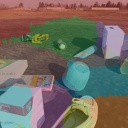} \\
    \mycaption{SLATE} & \myig{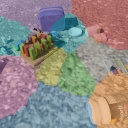} & \myig{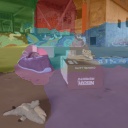} & \myig{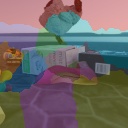} & \myig{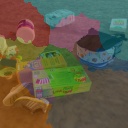} & \myig{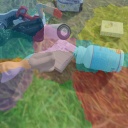} & \myig{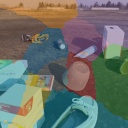} \\
\end{tabular}
}
    \caption{Masks on MOVi-E using 11 instead of 24 slots (quantitative results in \tab{app:tab:results-movi-e-num-slots}). Our method uses a ViT-B/8 encoder and the MLP decoder.}
    \label{app:fig:examples-extended-movi-e-11slots}
\end{figure}

\begin{figure}[tb]
    \centering
    {
\centering
\newcommand{\myig}[1]{\includegraphics[width=0.15\textwidth,valign=c]{#1}}
\renewcommand{\arraystretch}{4.5}
\setlength{\tabcolsep}{1.5pt}
\vspace{-2em}
\begin{tabular}{@{}ccccccc@{}}
    Ground Truth & \makecell{MLP\\Decoder} & \makecell{Transformer\\Decoder} & \phantom{\,} & Ground Truth & \makecell{MLP\\Decoder} & \makecell{Transformer\\Decoder} \\[-0.5em]
    \myig{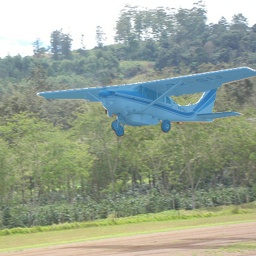} & \myig{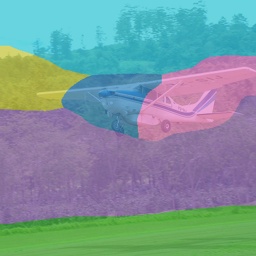} & \myig{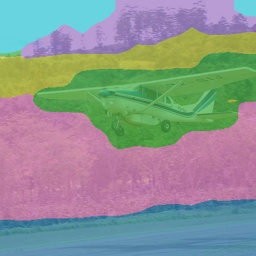} && \myig{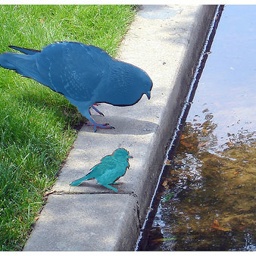} & \myig{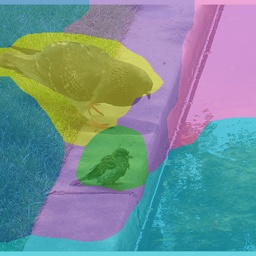} & \myig{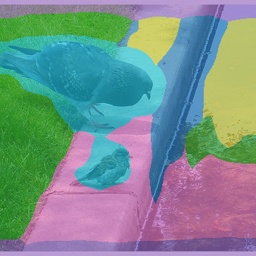} \\
    \myig{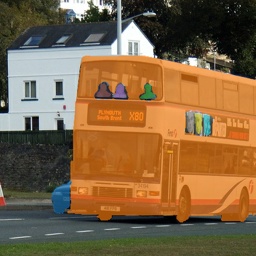} & \myig{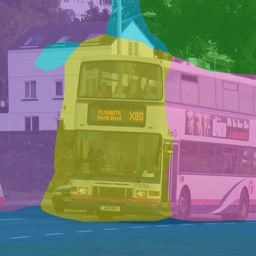} & \myig{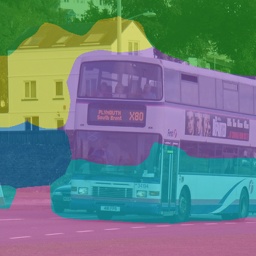} && \myig{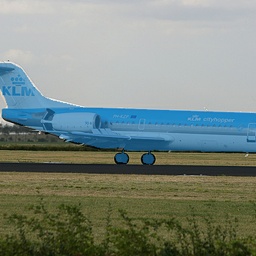} & \myig{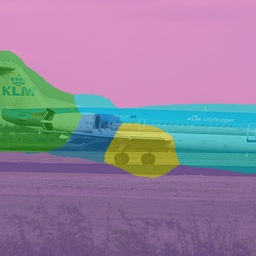} & \myig{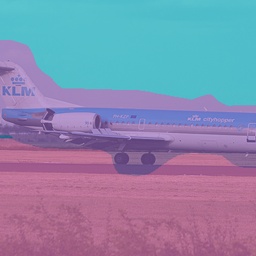} \\
    \myig{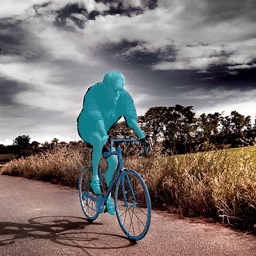} & \myig{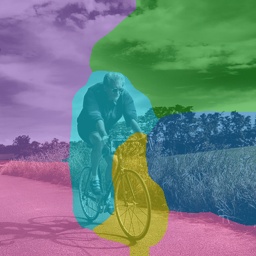} & \myig{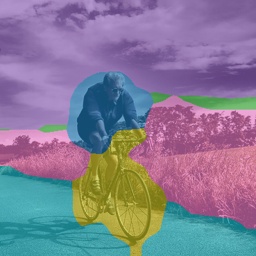} && \myig{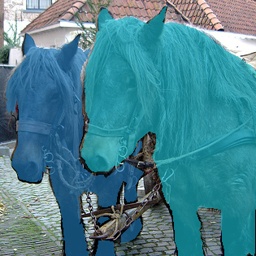} & \myig{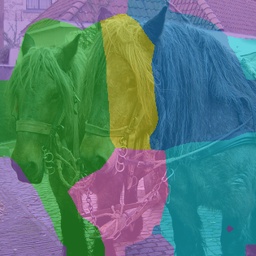} & \myig{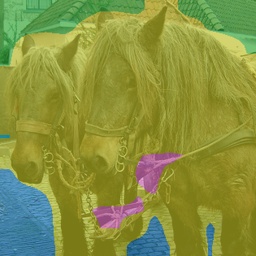} \\
    \myig{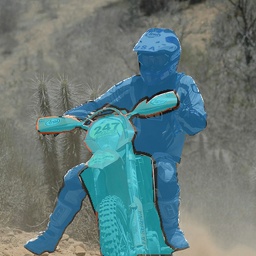} & \myig{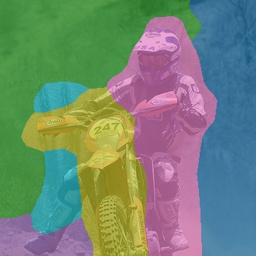} & \myig{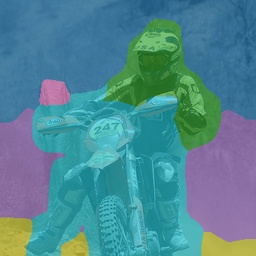} && \myig{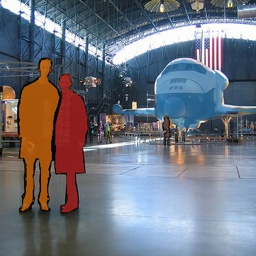} & \myig{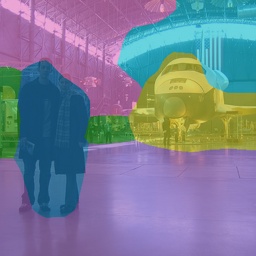} & \myig{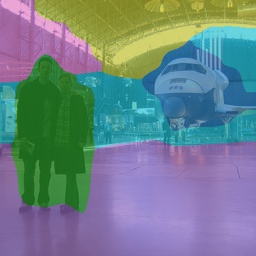} \\
    \myig{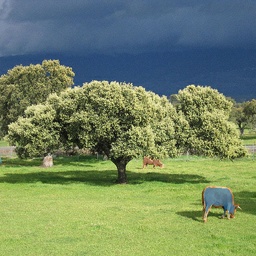} & \myig{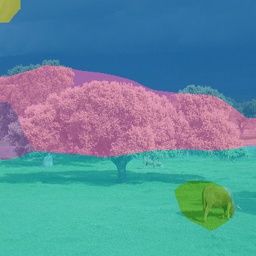} & \myig{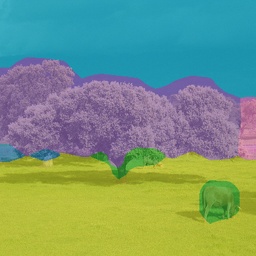} && \myig{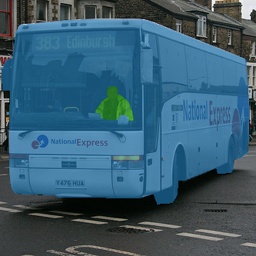} & \myig{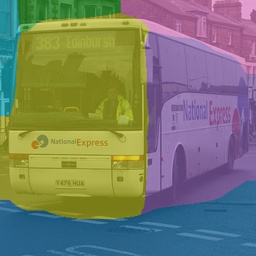} & \myig{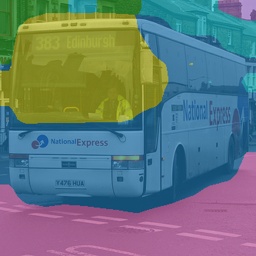} \\
    \myig{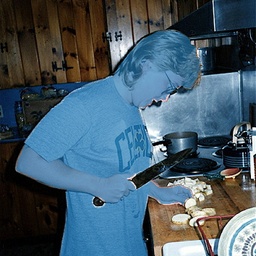} & \myig{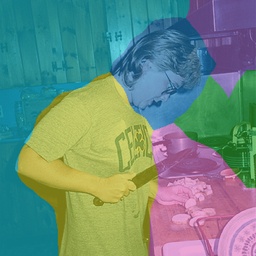} & \myig{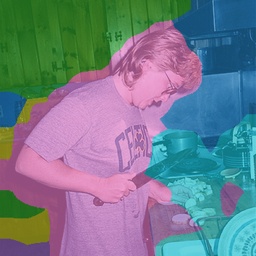} && \myig{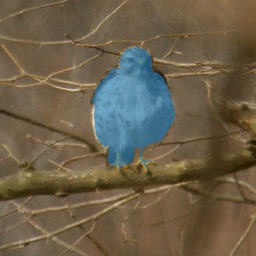} & \myig{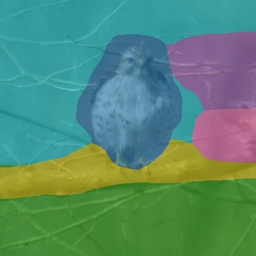} & \myig{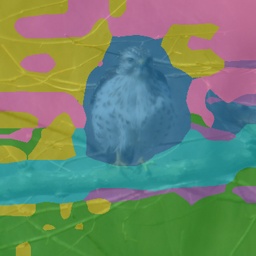} \\
    \myig{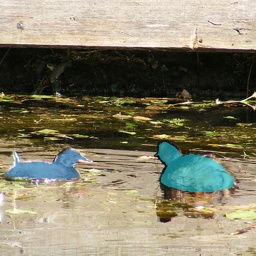} & \myig{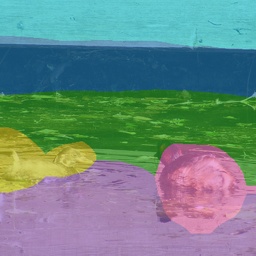} & \myig{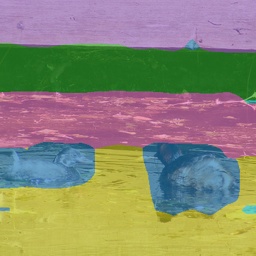} && \myig{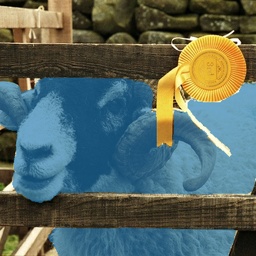} & \myig{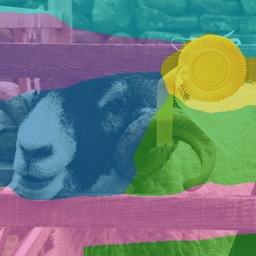} & \myig{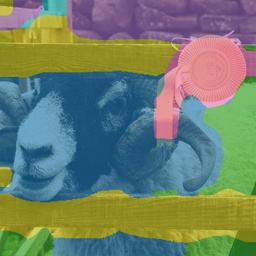} \\
    \myig{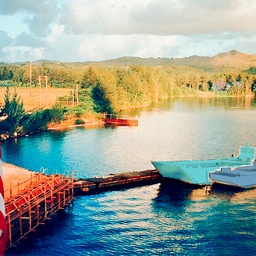} & \myig{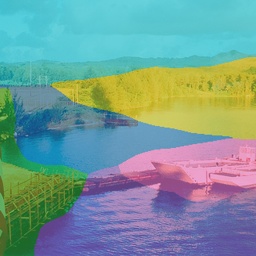} & \myig{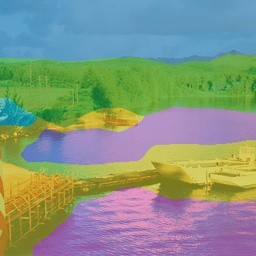} && \myig{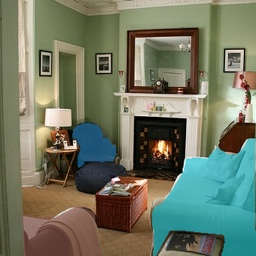} & \myig{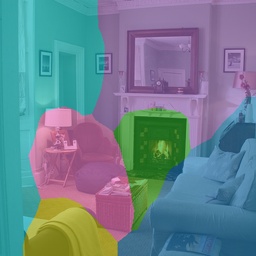} & \myig{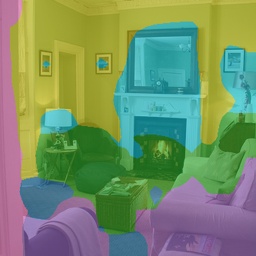} \\
    \myig{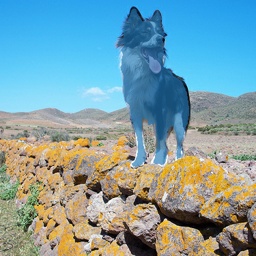} & \myig{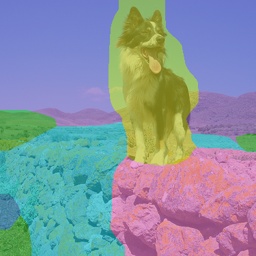} & \myig{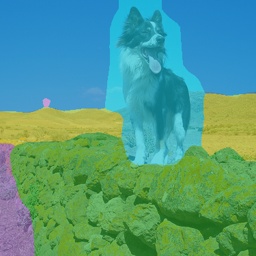} && \myig{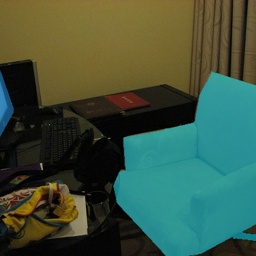} & \myig{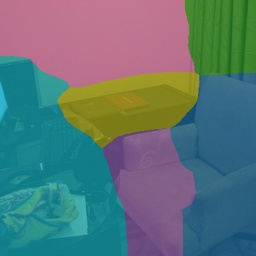} & \myig{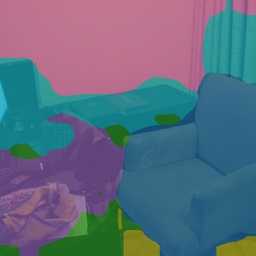} \\
\end{tabular}
}
    \caption{Masks on PASCAL VOC 2012 produced by our method, using 6 slots and a ViT-B/16 encoder. We show predictions from the MLP and the Transformer decoder.}
    \label{app:fig:examples-extended-pascal}
\end{figure}

\begin{figure}[tb]
    \centering
    {
\centering
\newcommand{\myig}[1]{\includegraphics[width=0.15\textwidth,valign=c]{#1}}
\renewcommand{\arraystretch}{4.5}
\setlength{\tabcolsep}{1.5pt}
\vspace{-2em}
\begin{tabular}{@{}ccccccc@{}}
    Ground Truth & \makecell{MLP\\Decoder} & \makecell{Transformer\\Decoder} & \phantom{\,} & Ground Truth & \makecell{MLP\\Decoder} & \makecell{Transformer\\Decoder} \\[-0.5em]
    \myig{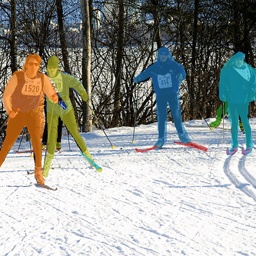} & \myig{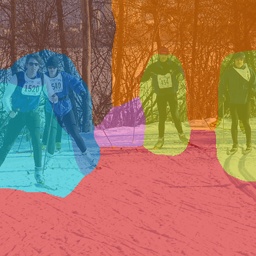} & \myig{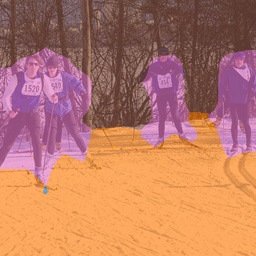} && \myig{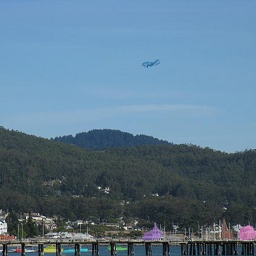} & \myig{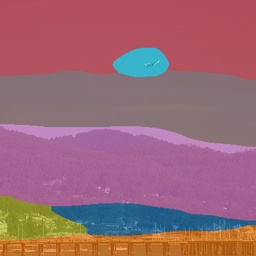} & \myig{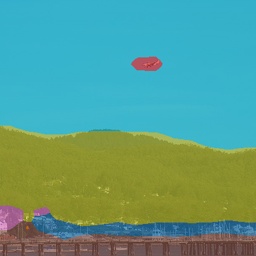} \\
    \myig{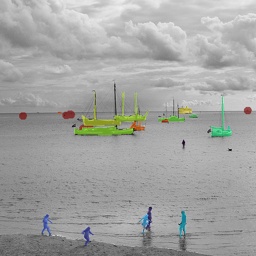} & \myig{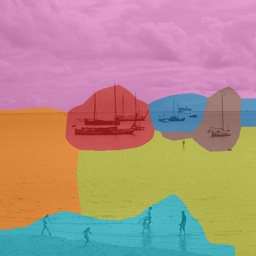} & \myig{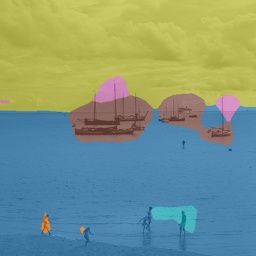} && \myig{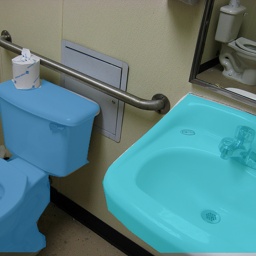} & \myig{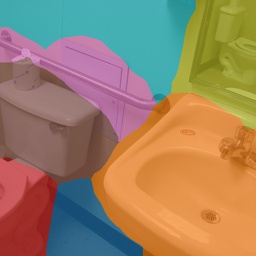} & \myig{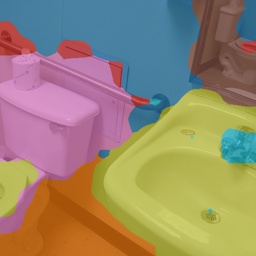} \\
    \myig{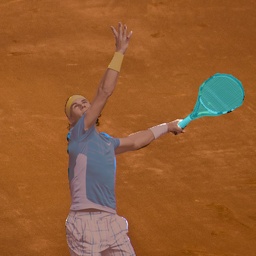} & \myig{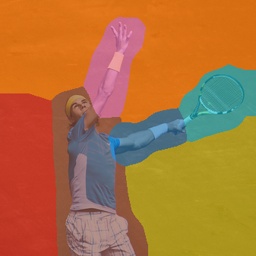} & \myig{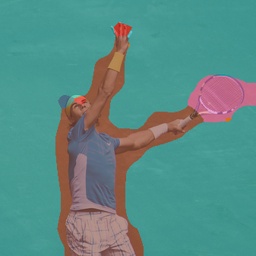} && \myig{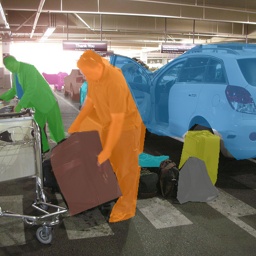} & \myig{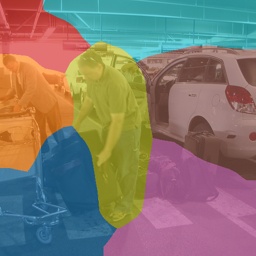} & \myig{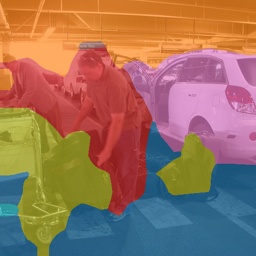} \\
    \myig{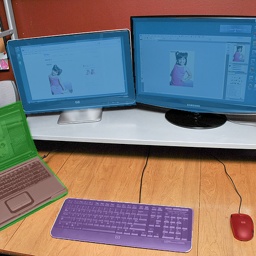} & \myig{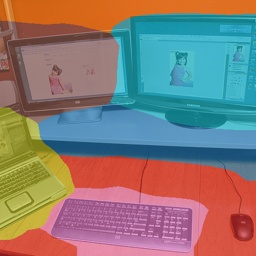} & \myig{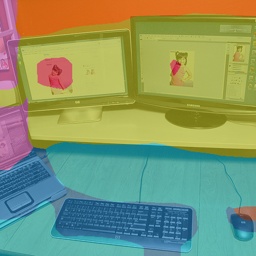} && \myig{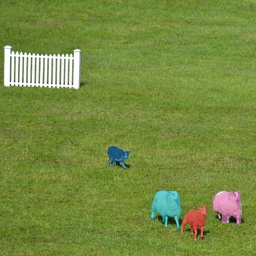} & \myig{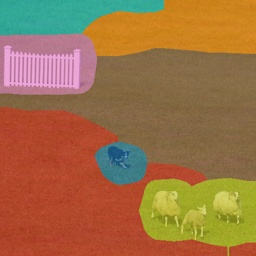} & \myig{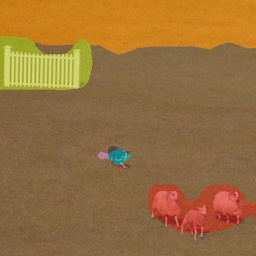} \\
    \myig{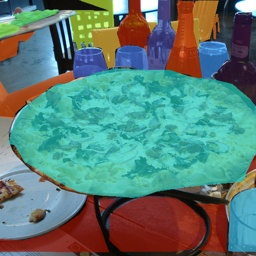} & \myig{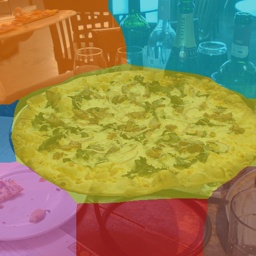} & \myig{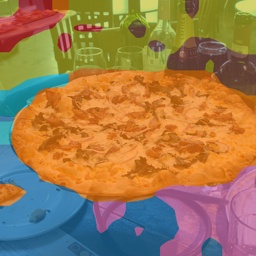} && \myig{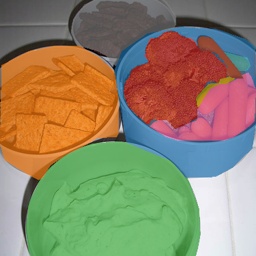} & \myig{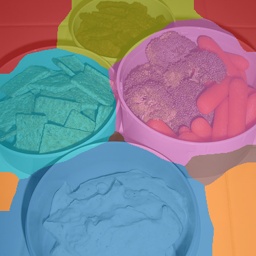} & \myig{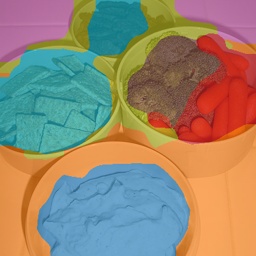} \\
    \myig{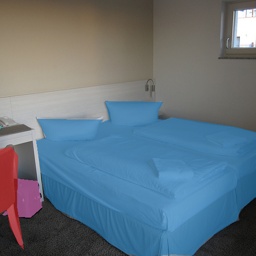} & \myig{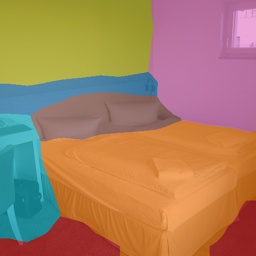} & \myig{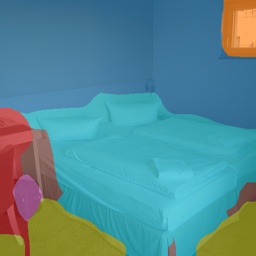} && \myig{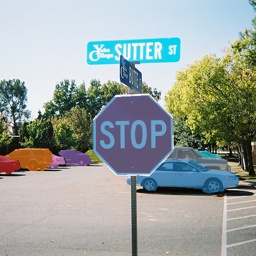} & \myig{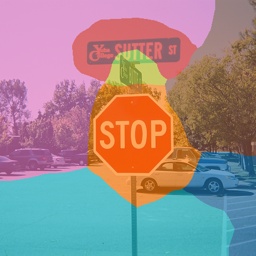} & \myig{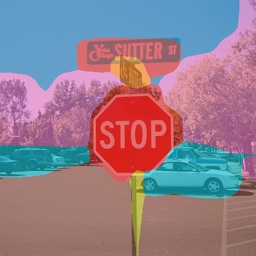} \\
    \myig{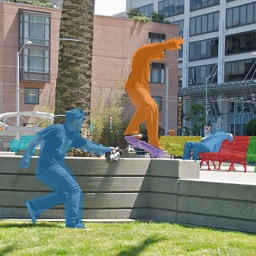} & \myig{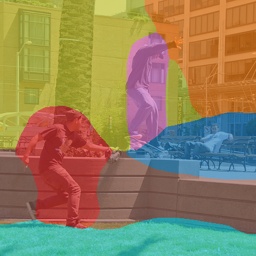} & \myig{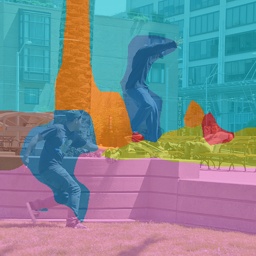} && \myig{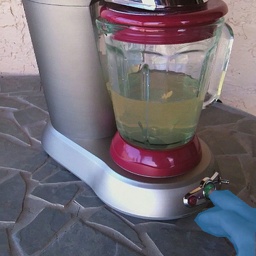} & \myig{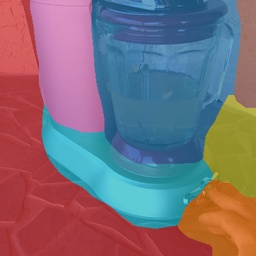} & \myig{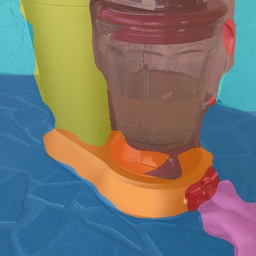} \\
    \myig{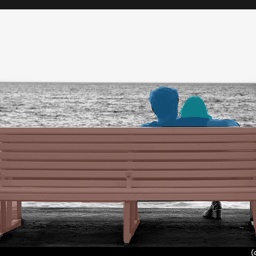} & \myig{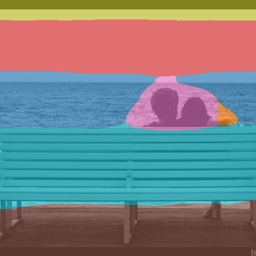} & \myig{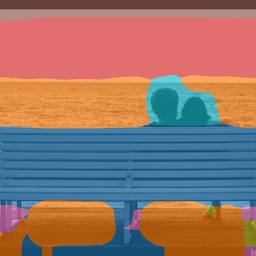} && \myig{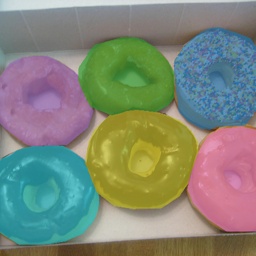} & \myig{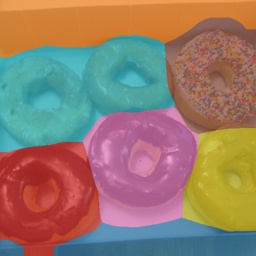} & \myig{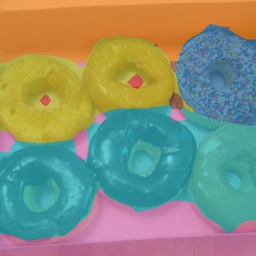} \\
    \myig{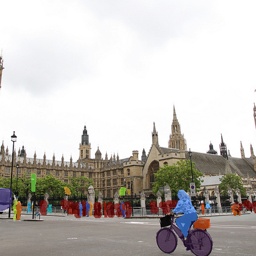} & \myig{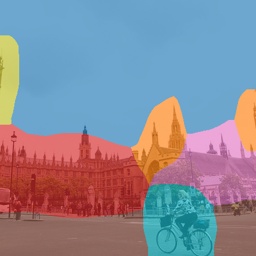} & \myig{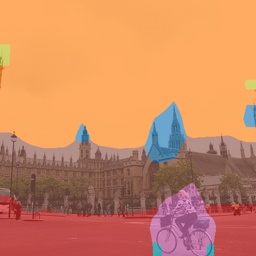} && \myig{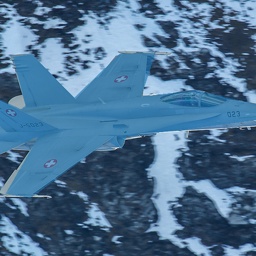} & \myig{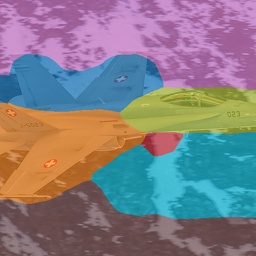} & \myig{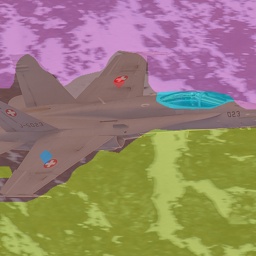} \\
\end{tabular}
}
    \caption{Masks on COCO 2017 produced by our method, using 7 slots and a ViT-B/16 encoder. We show predictions from the MLP and the Transformer decoder.}
    \label{app:fig:examples-extended-coco}
\end{figure}

\begin{figure}[thb]
    \centering
    {
\centering
\newcommand{\myig}[1]{\includegraphics[width=0.135\textwidth,valign=c]{#1}}
\renewcommand{\arraystretch}{4.1}
\setlength{\tabcolsep}{1.3pt}
\begin{tabular}{@{}cccccccc@{}}
    \makecell{ViT-B/16\\Supervised} & \makecell{ViT-S/16\\DINO} & \makecell{ViT-B/16\\DINO} & \makecell{ViT-S/8\\DINO} & \makecell{ViT-B/16\\MoCo-v3} & \makecell{ViT-B/16\\MSN} & \makecell{ViT-B/16\\MAE} \\[-0.5em]
    \myig{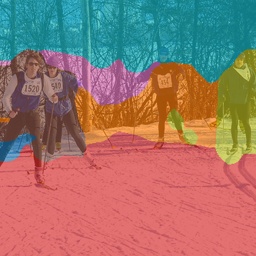} & \myig{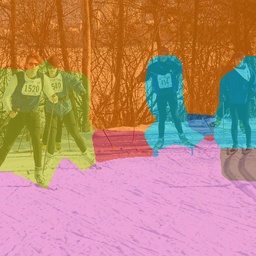} & \myig{images/coco/dinosaur_transformer_s7/144.jpg} & \myig{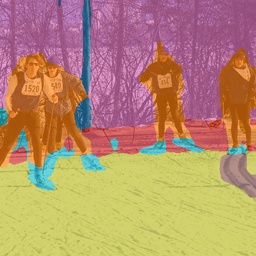} & \myig{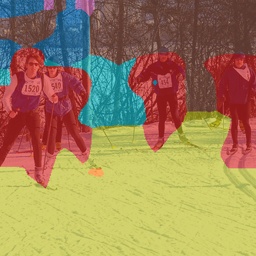} & \myig{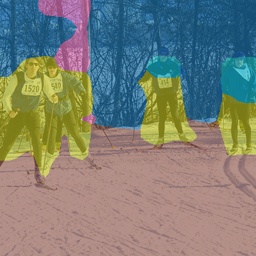} & \myig{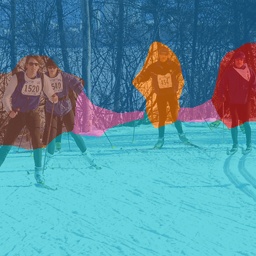} \\
    \myig{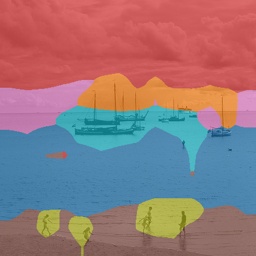} & \myig{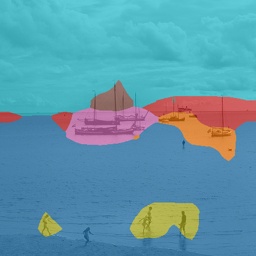} & \myig{images/coco/dinosaur_transformer_s7/147.jpg} & \myig{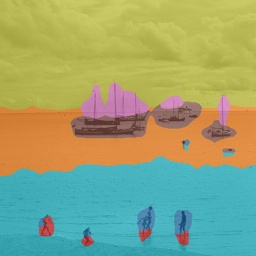} & \myig{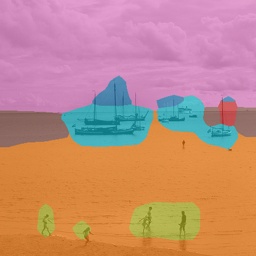} & \myig{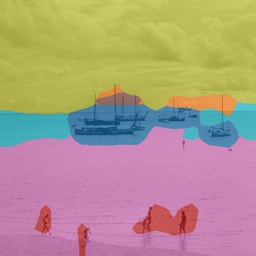} & \myig{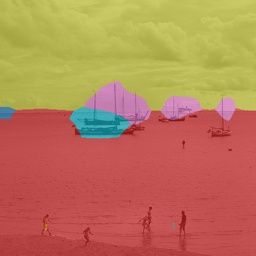} \\
    \myig{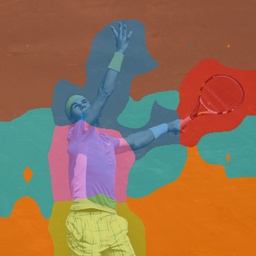} & \myig{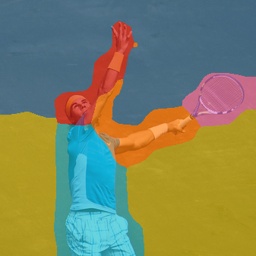} & \myig{images/coco/dinosaur_transformer_s7/161.jpg} & \myig{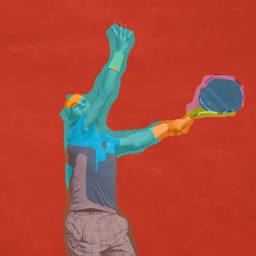} & \myig{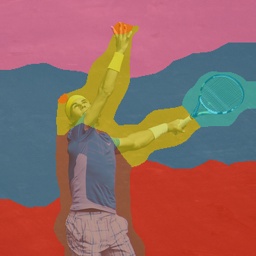} & \myig{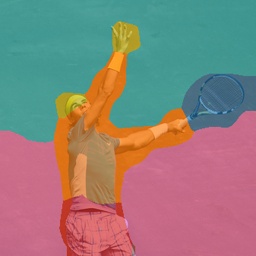} & \myig{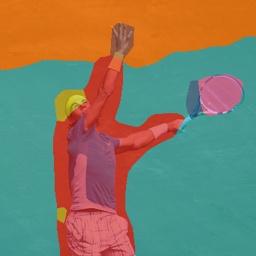} \\
    \myig{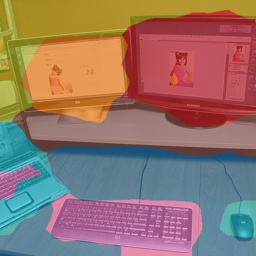} & \myig{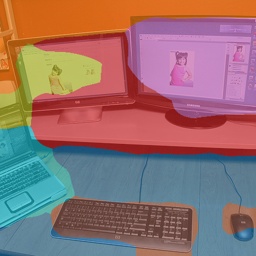} & \myig{images/coco/dinosaur_transformer_s7/169.jpg} & \myig{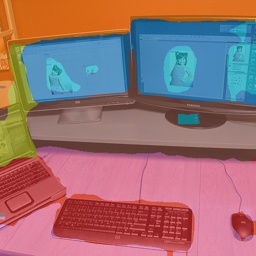} & \myig{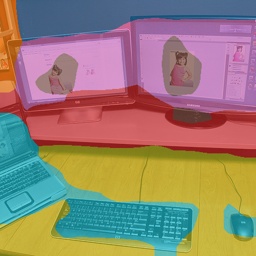} & \myig{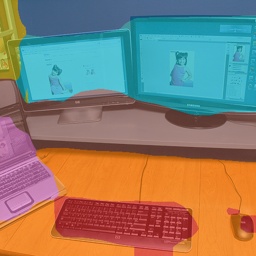} & \myig{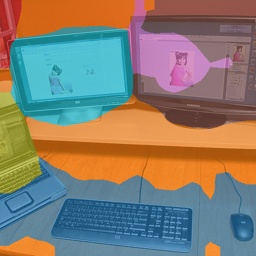} \\
    \myig{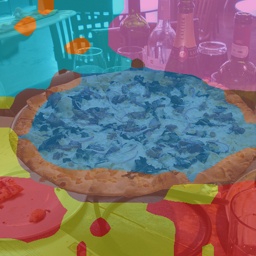} & \myig{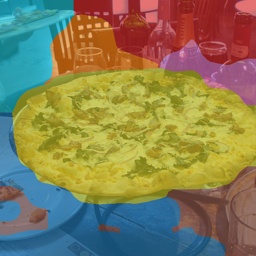} & \myig{images/coco/dinosaur_transformer_s7/199.jpg} & \myig{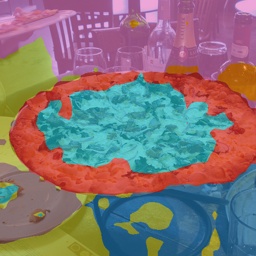} & \myig{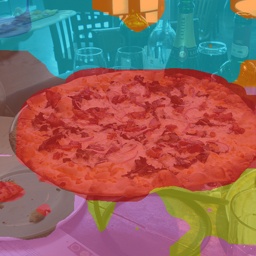} & \myig{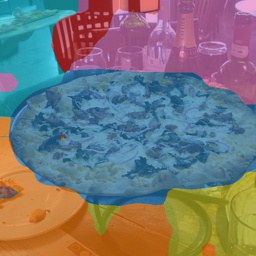} & \myig{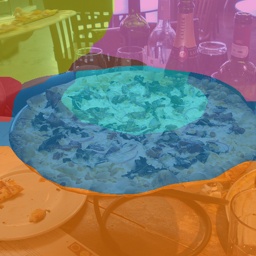} \\
    \myig{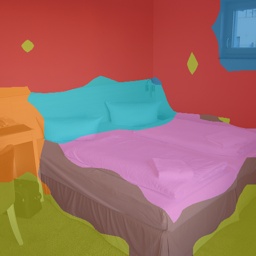} & \myig{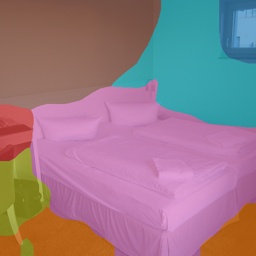} & \myig{images/coco/dinosaur_transformer_s7/207.jpg} & \myig{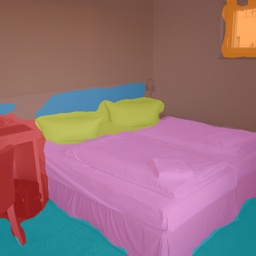} & \myig{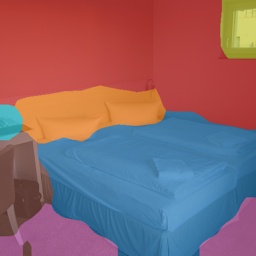} & \myig{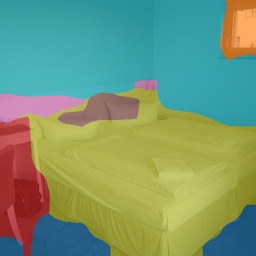} & \myig{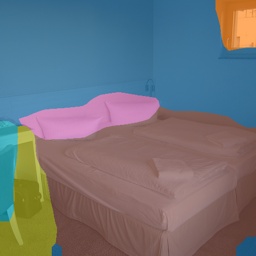} \\
    \myig{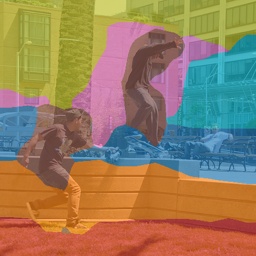} & \myig{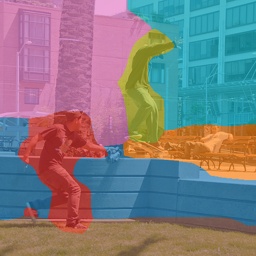} & \myig{images/coco/dinosaur_transformer_s7/314.jpg} & \myig{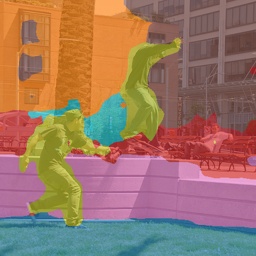} & \myig{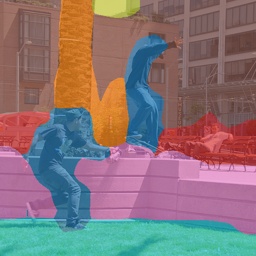} & \myig{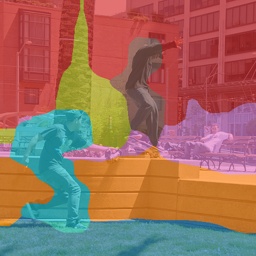} & \myig{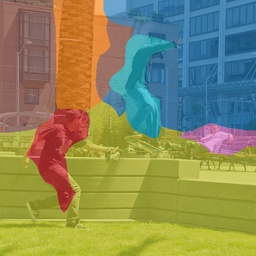} \\
    \myig{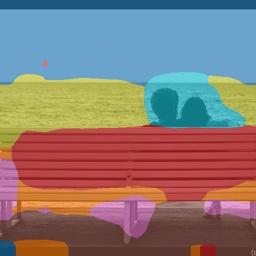} & \myig{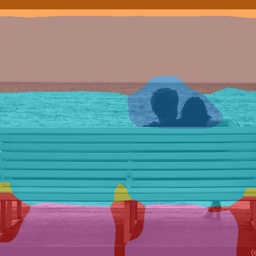} & \myig{images/coco/dinosaur_transformer_s7/269.jpg} & \myig{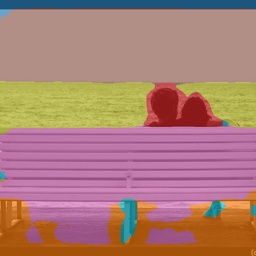} & \myig{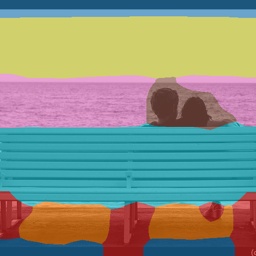} & \myig{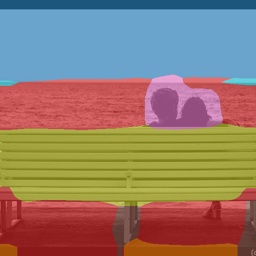} & \myig{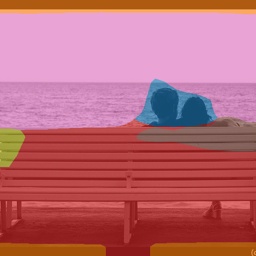} \\
    \myig{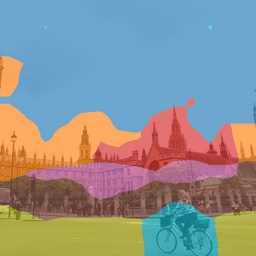} & \myig{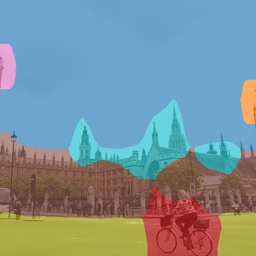} & \myig{images/coco/dinosaur_transformer_s7/290.jpg} & \myig{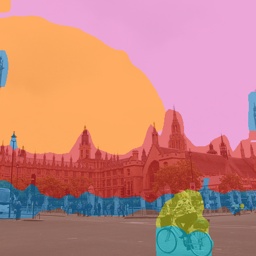} & \myig{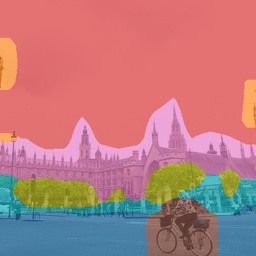} & \myig{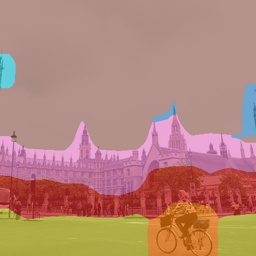} & \myig{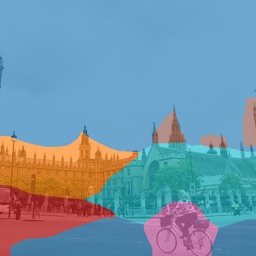} \\
\end{tabular}
}
    \caption{Masks on COCO 2017 produced by our method, using different encoders and pre-training schemes on COCO 2017 (\cf \tab{app:tab:encoder-ablation-additional-results}).
    All models use 7 slots and the Transformer decoder.
    }
    \label{app:fig:examples-coco-encoders}
\end{figure}

\begin{figure}[thb]
    \centering
    {
\centering
\newcommand{\myig}[1]{\includegraphics[width=0.135\textwidth,valign=c]{#1}}
\renewcommand{\arraystretch}{4.1}
\setlength{\tabcolsep}{1.3pt}
\begin{tabular}{@{}cccccccc@{}}
    \makecell{RN50\\Supervised} & \makecell{RN50\\DINO} & \makecell{RN34 (S) \\DINO RN50} & \makecell{RN50\\MoCo-v3} & \makecell{RN34 (S)\\MoCo-v3\\RN50} & \makecell{ViT-B/16\\DINO} & \makecell{RN34 (S)\\DINO ViT} \\
    \myig{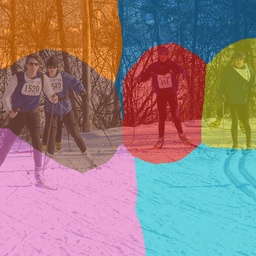} & \myig{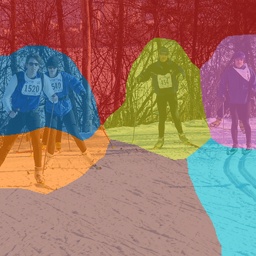} & \myig{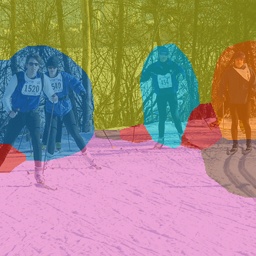} & \myig{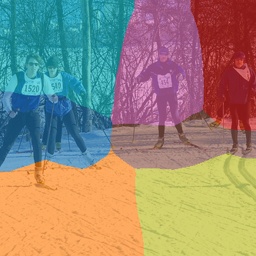} & \myig{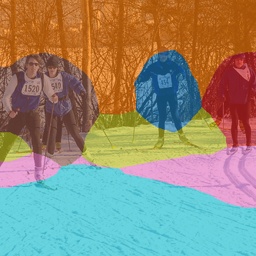} & \myig{images/coco/dinosaur_mlp_s7/144.jpg} & \myig{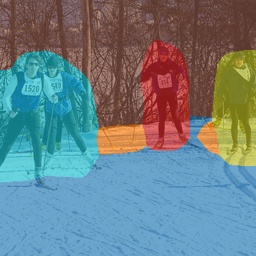} \\
    \myig{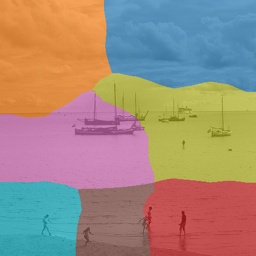} & \myig{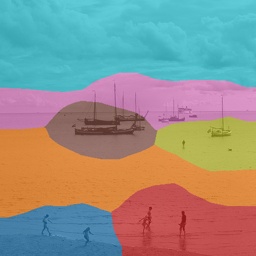} & \myig{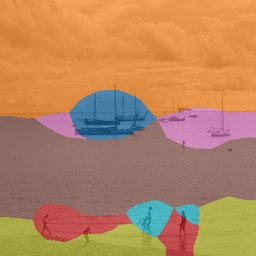} & \myig{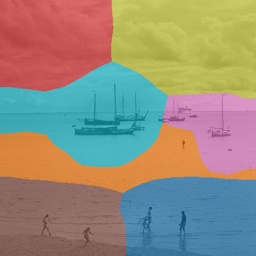} & \myig{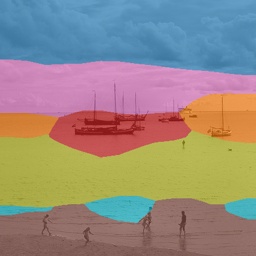} & \myig{images/coco/dinosaur_mlp_s7/147.jpg} & \myig{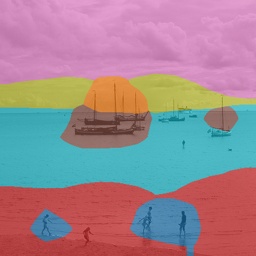} \\
    \myig{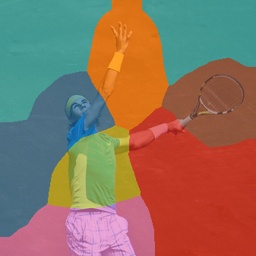} & \myig{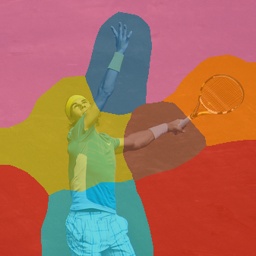} & \myig{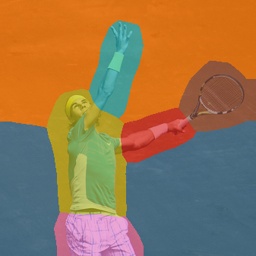} & \myig{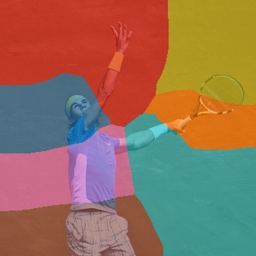} & \myig{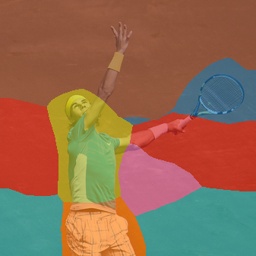} & \myig{images/coco/dinosaur_mlp_s7/161.jpg} & \myig{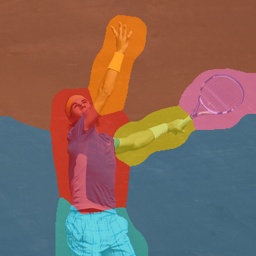} \\
    \myig{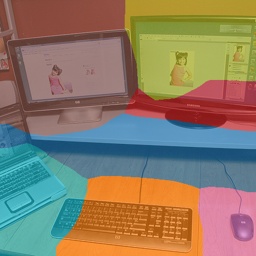} & \myig{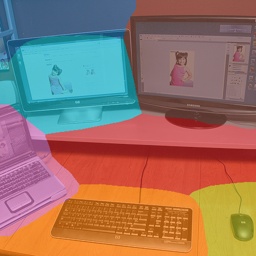} & \myig{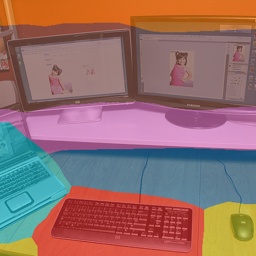} & \myig{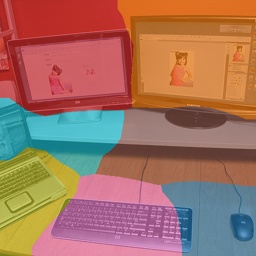} & \myig{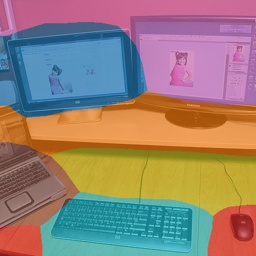} & \myig{images/coco/dinosaur_mlp_s7/169.jpg} & \myig{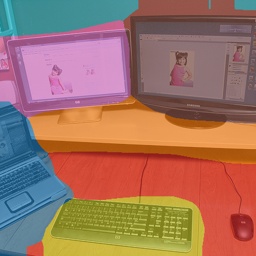} \\
    \myig{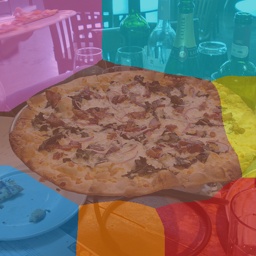} & \myig{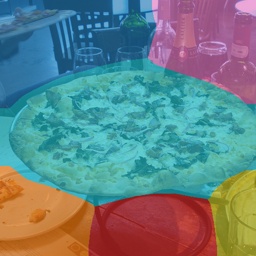} & \myig{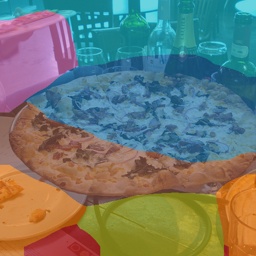} & \myig{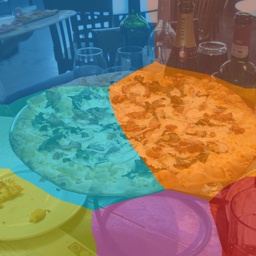} & \myig{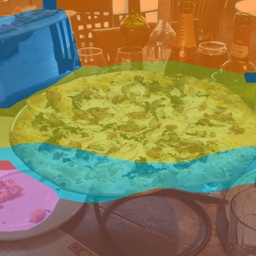} & \myig{images/coco/dinosaur_mlp_s7/199.jpg} & \myig{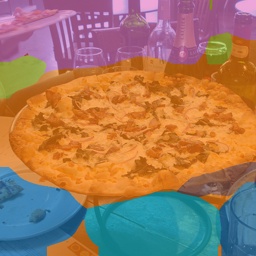} \\
    \myig{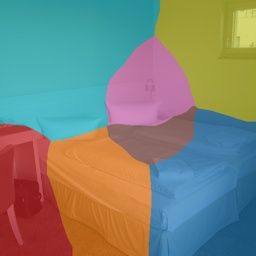} & \myig{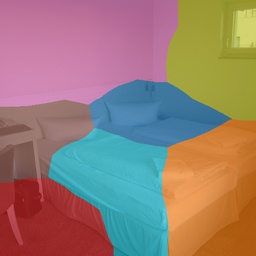} & \myig{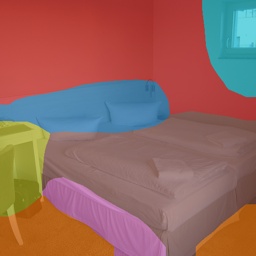} & \myig{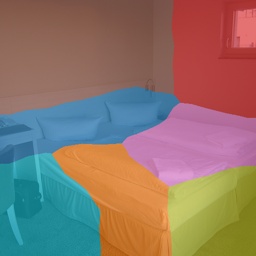} & \myig{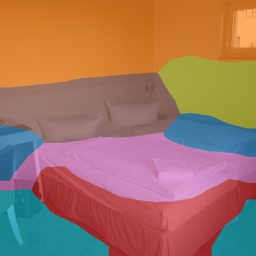} & \myig{images/coco/dinosaur_mlp_s7/207.jpg} & \myig{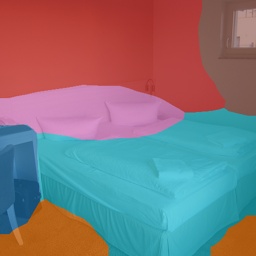} \\
    \myig{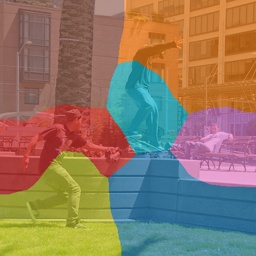} & \myig{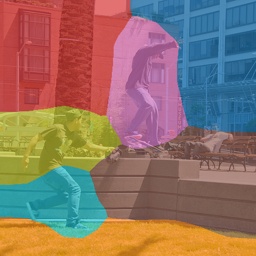} & \myig{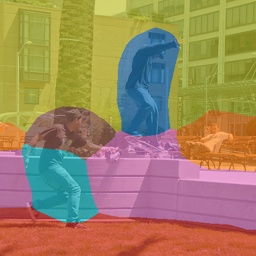} & \myig{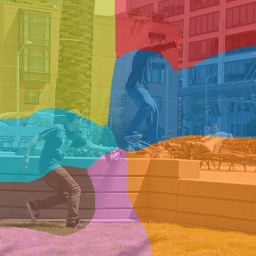} & \myig{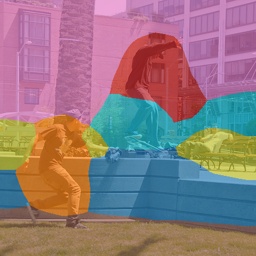} & \myig{images/coco/dinosaur_mlp_s7/314.jpg} & \myig{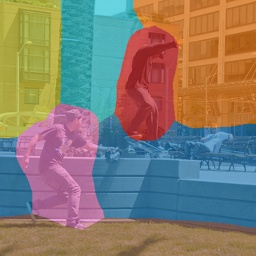} \\
    \myig{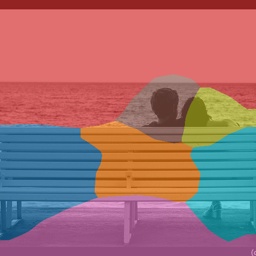} & \myig{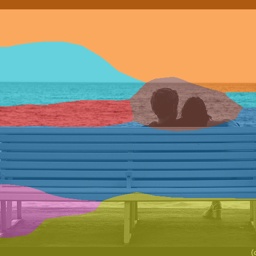} & \myig{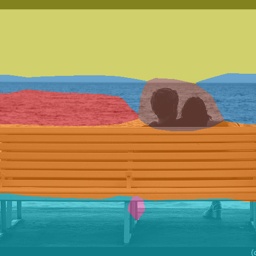} & \myig{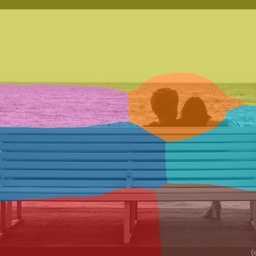} & \myig{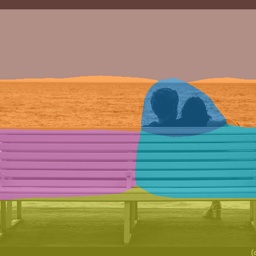} & \myig{images/coco/dinosaur_mlp_s7/269.jpg} & \myig{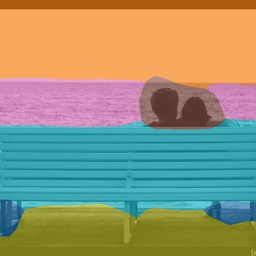} \\
    \myig{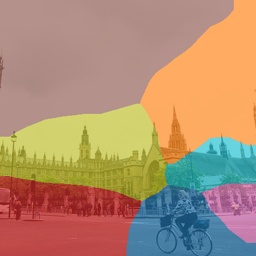} & \myig{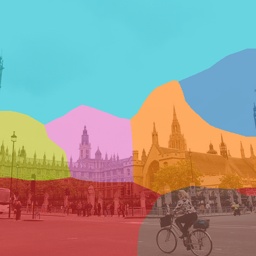} & \myig{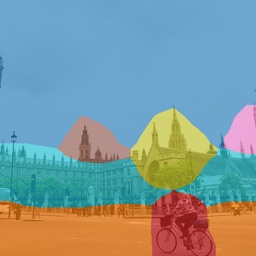} & \myig{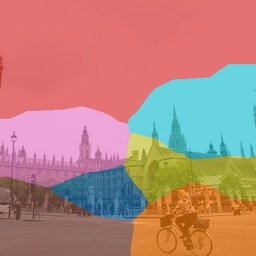} & \myig{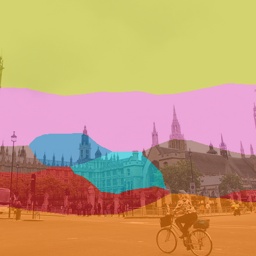} & \myig{images/coco/dinosaur_mlp_s7/290.jpg} & \myig{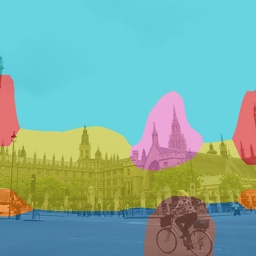} \\
\end{tabular}
}
    \caption{Masks on COCO 2017 produced by our method, using different encoders and pre-training schemes on COCO 2017.
    ResNet34s (RN34) were trained from scratch (``S''), by reconstruction of ResNet50 (RN50) or ViT-B/16 features (\cf \tab{app:tab:analysis-encoder-scratch}).
    All models use 7 slots and the MLP decoder.
    }
    \label{app:fig:examples-coco-encoders-mlp}
\end{figure}

\begin{figure}[thb]
    \centering
    {
\centering
\newcommand{\myig}[1]{\includegraphics[width=0.145\textwidth,valign=c]{#1}}
\renewcommand{\arraystretch}{4.4}
\setlength{\tabcolsep}{1.5pt}
\vspace{-1em}
\begin{tabular}{@{}cccccccccc@{}}
    \makecell{ViT-B/16\\MoCo-v3} & \makecell{RN34 (S)\\MoCo-v3 ViT} & \makecell{ViT-B/16\\MSN} & \makecell{RN34 (S)\\MSN ViT} & \makecell{ViT-B/16\\MAE} & \makecell{RN34 (S)\\MAE ViT} \\[-0.8em]
    \myig{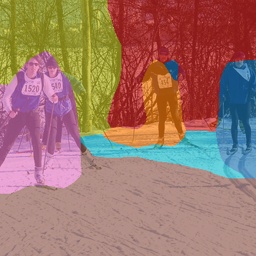} & \myig{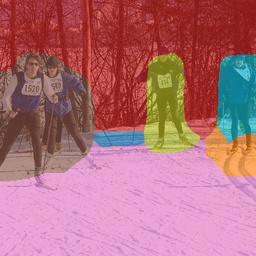} & \myig{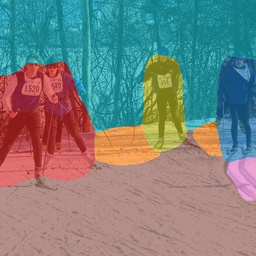} & \myig{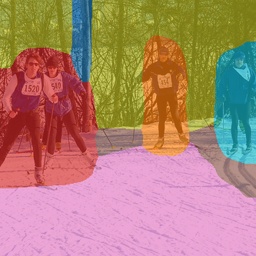} & \myig{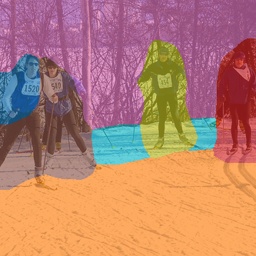} & \myig{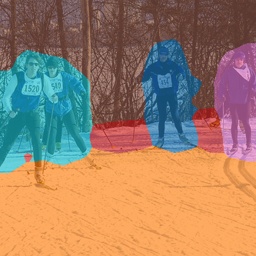} \\
    \myig{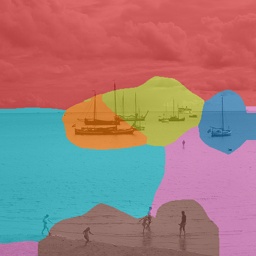} & \myig{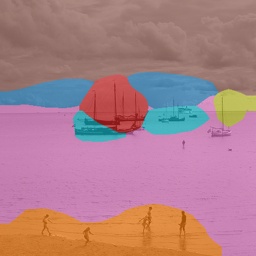} & \myig{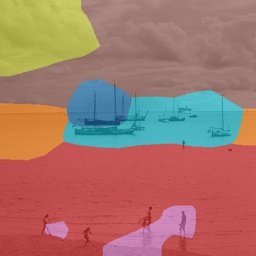} & \myig{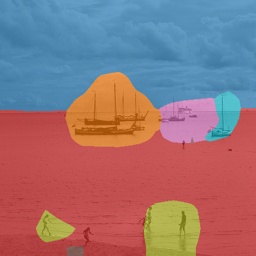} & \myig{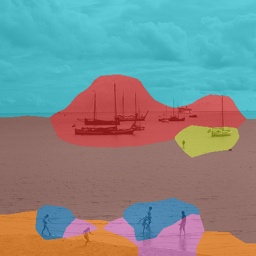} & \myig{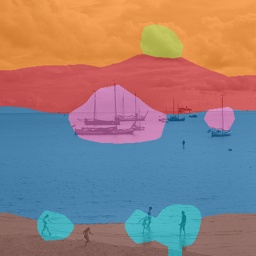} \\
    \myig{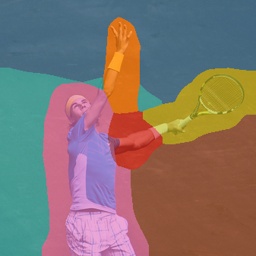} & \myig{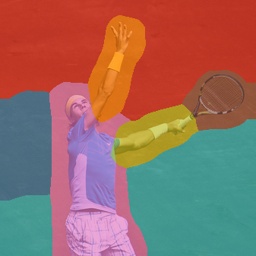} & \myig{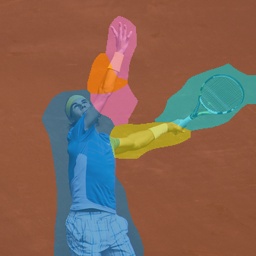} & \myig{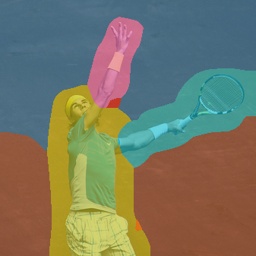} & \myig{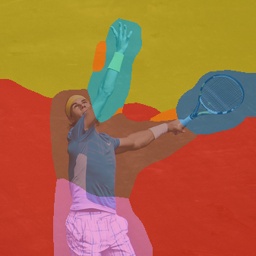} & \myig{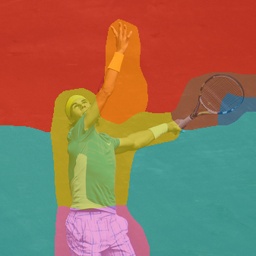} \\
    \myig{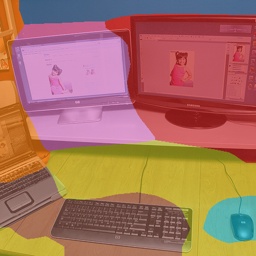} & \myig{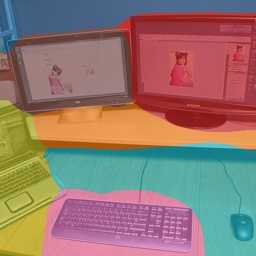} & \myig{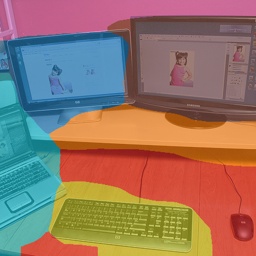} & \myig{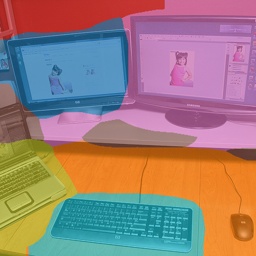} & \myig{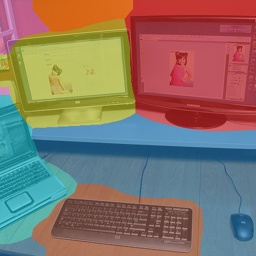} & \myig{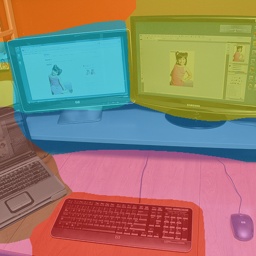} \\
    \myig{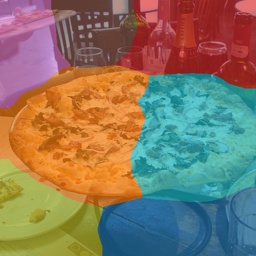} & \myig{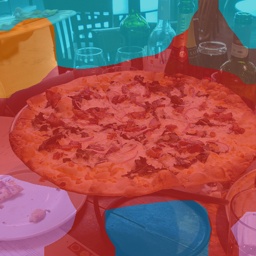} & \myig{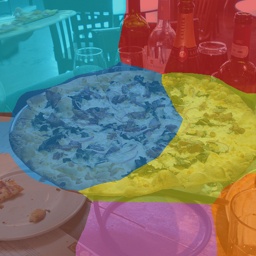} & \myig{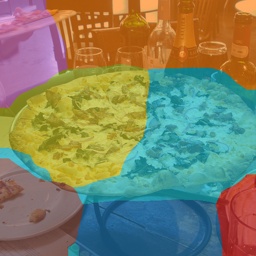} & \myig{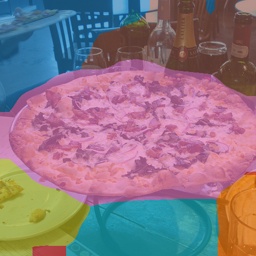} & \myig{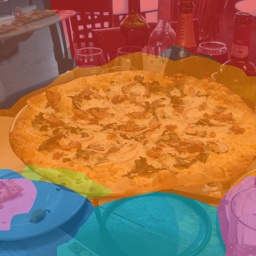} \\
    \myig{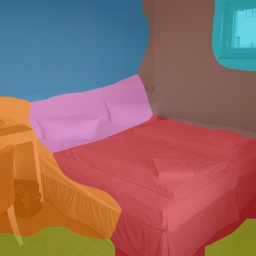} & \myig{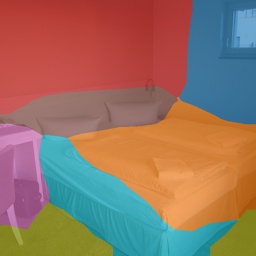} & \myig{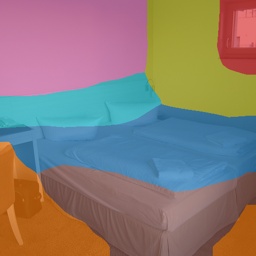} & \myig{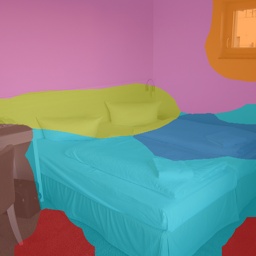} & \myig{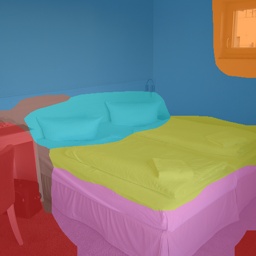} & \myig{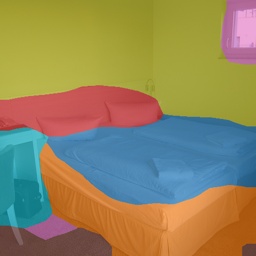} \\
    \myig{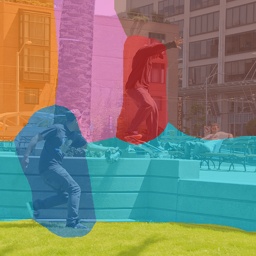} & \myig{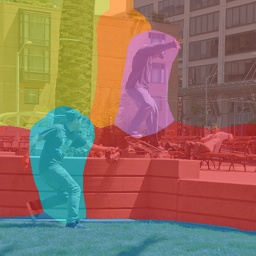} & \myig{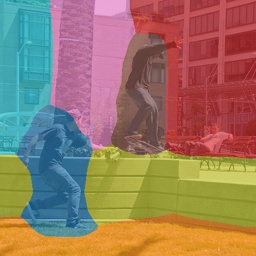} & \myig{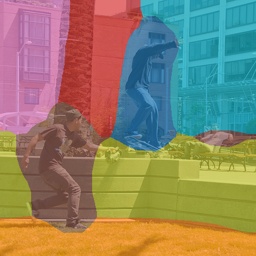} & \myig{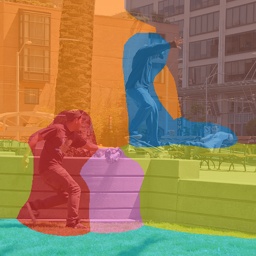} & \myig{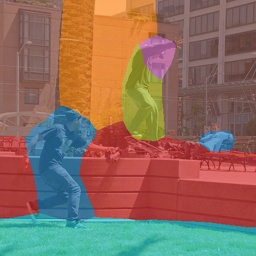} \\
    \myig{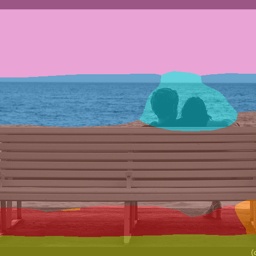} & \myig{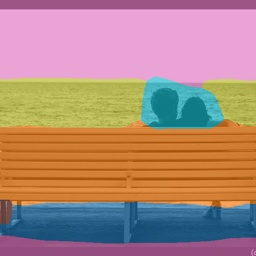} & \myig{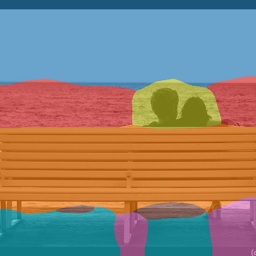} & \myig{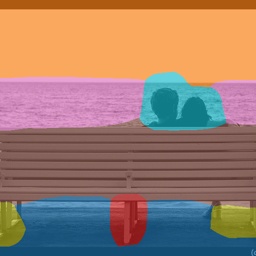} & \myig{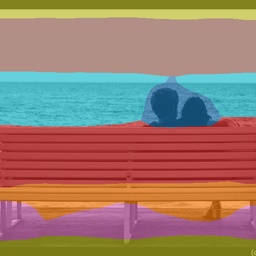} & \myig{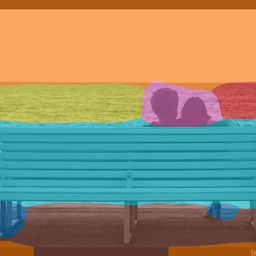} \\
    \myig{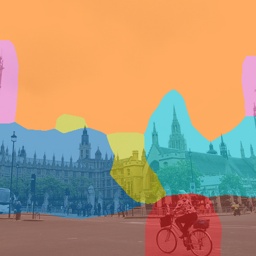} & \myig{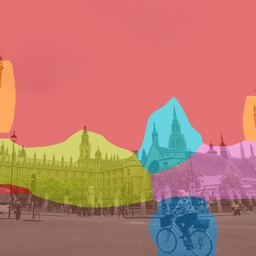} & \myig{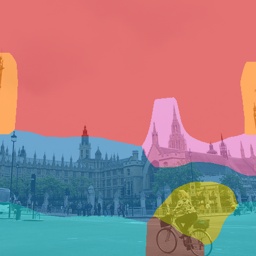} & \myig{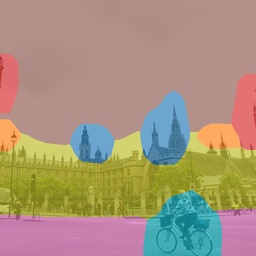} & \myig{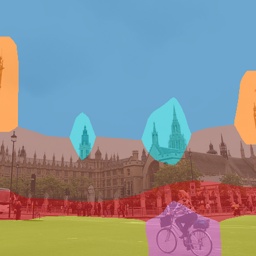} & \myig{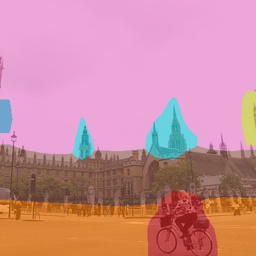} \\
\end{tabular}
}
    \caption{Masks on COCO 2017 produced by our method, using different encoders and pre-training schemes on COCO 2017.
    ResNet34s (RN34) were trained from scratch (``S''), by reconstruction of ViT-B/16 features (\cf \tab{app:tab:analysis-encoder-scratch}).
    All models use 7 slots and the MLP decoder.
    }
    \label{app:fig:examples-coco-encoders-mlp-2}
\end{figure}

We include additional mask predictions of our model for MOVi-C (\fig{app:fig:examples-extended-movi-c}), for MOVi-E with 24 (\fig{app:fig:examples-extended-movi-e-24slots}) and 11 slots (\fig{app:fig:examples-extended-movi-e-11slots}), for PASCAL VOC 2012 (\fig{app:fig:examples-extended-pascal}) and COCO (\fig{app:fig:examples-extended-coco}, \fig{app:fig:examples-coco-encoders}, \fig{app:fig:examples-coco-encoders-mlp}, \fig{app:fig:examples-coco-encoders-mlp-2}).

\end{document}